\pdfoutput=1
\documentclass[a4paper,fleqn]{cas-sc}

\usepackage[numbers,sort&compress]{natbib}
\usepackage{amsmath,amssymb,amsfonts}
\usepackage{mathrsfs}
\usepackage{booktabs}
\usepackage{multirow}
\usepackage{makecell}
\usepackage{subfig}
\usepackage{algorithm}
\usepackage{algorithmic}
\usepackage{url}

\makeatletter
\AtBeginDocument{\csxdef{lastpage}{\noexpand\pageref{LastPage}}}
\makeatother

\graphicspath{{figures/}{./}}

\newif\ifrevision
\revisionfalse
\ifrevision
  \newcommand{\rev}[1]{\textcolor{red}{#1}}
\else
  \newcommand{\rev}[1]{#1}
\fi

\begin{document}
\let\WriteBookmarks\relax
\def\floatpagepagefraction{1}
\def\textpagefraction{.001}

\shorttitle{Zero-shot rib design}
\shortauthors{Y. Kwon and N. Kang}

\title[mode = title]{Zero-shot rib design: merging a training-free generative prior with topology optimization}

\author[1,2]{Yongmin Kwon}
\ead{kymin1002@kaist.ac.kr}
\credit{Conceptualization, Data curation, Formal analysis, Investigation, Methodology, Resources, Software, Validation, Visualization, Writing -- original draft}

\author[1,2]{Namwoo Kang}
\cormark[1]
\ead{nwkang@kaist.ac.kr}
\credit{Project administration, Supervision, Investigation, Funding acquisition, Writing -- review \& editing}

\affiliation[1]{organization={Cho Chun Shik Graduate School of Mobility, Korea Advanced Institute of Science and Technology},
            addressline={193 Munji-ro, Yuseong-gu},
            city={Daejeon},
            postcode={34051},
            country={Republic of Korea}}

\affiliation[2]{organization={AI Research Team, Narnia Labs},
            addressline={193 Munji-ro, Yuseong-gu},
            city={Daejeon},
            postcode={34051},
            country={Republic of Korea}}

\cortext[1]{Corresponding author.}

\begin{abstract}
Natural load-bearing patterns such as leaf venation, trabecular bone, and spider webs achieve high stiffness per unit mass, yet classical topology optimizers rarely reach such geometries, and few let engineers express structural design intent through natural language.
This work treats a frozen text-to-image diffusion model as a training-free source of design knowledge and distills it into the physics loop of density-based topology optimization via score distillation sampling, so that a text prompt becomes an explicit, machine-interpretable representation of engineer intent.
The prompt-induced generative gradient and the finite element sensitivity are combined at every iteration, letting physics decide which prompt-induced features survive.
In 245 primary SDS runs spanning four geometric domains and two physics regimes, 38 of 49 prompt--domain combinations achieved statistically significant compliance reductions (up to $-31.5\%$ mechanical and $-23.0\%$ thermoelastic), outperforming gradient-based baselines.
Cross-domain morphological analysis identifies a recurring structural signature of improvement: in most domains the generative prior suppresses dead-end branches in the rib skeleton, with endpoint--compliance correlation $r = +0.56$ to $+0.99$.
A Heaviside projection with $\beta$-continuation resolves a pronounced intermediate-density tendency in this diffusion--physics coupling ($42.6\%$ to $<3\%$), and an automated skeleton-based pipeline converts optimized density fields into \rev{candidate geometry ready for computer-aided design}.
\rev{By retargeting the generative prior across domains, loading conditions, and physics objectives through a change of text prompt, with each new problem's physics setup specified separately, the framework uses a pretrained generative model as a reusable, training-free prior for engineering design.}
\end{abstract}

\begin{keywords}
Topology optimization \sep Score distillation sampling \sep Text-guided structural design \sep Rib reinforcement design \sep Training-free generative design \sep Design intent representation
\end{keywords}

\maketitle

\section{Introduction}
\label{sec:introduction}

Topology optimization determines the optimal distribution of material within a prescribed design domain to maximize structural performance subject to physical and manufacturing constraints \cite{bendsoe2013topology}.
The Solid Isotropic Material with Penalization (SIMP) method is the most widely adopted framework. A continuous density variable $\rho_e\in[0,1]$ is assigned to each finite element and updated by gradient-based optimization to minimize structural compliance \cite{bendsoe1999material, bendsoe1989optimal, sigmund200199, andreassen2011efficient}. \rev{Recent isogeometric formulations employ multi-patch, T-spline, and B\'ezier-extraction representations \cite{gao2023multi, zhang2024t, zhang2025t} to handle complex plate and shell geometries with smoother boundaries, while particle-flow and material-point formulations \cite{lin2025lagrangian, zhu2026particle} provide alternative Lagrangian descriptions well suited to large geometric changes and hinge-free compliant-mechanism design.}
Despite its maturity, SIMP suffers from a fundamental limitation: the non-convexity introduced by the penalization parameter generates a rugged objective landscape with numerous local minima, so gradient-based optimizers converge to different solutions depending on the initial density distribution, with no guarantee of approaching the global optimum \cite{sigmund1998numerical, rozvany2009critical}.

This limitation is particularly acute in the design of rib reinforcements for thin-walled plate structures.
Rib stiffeners serve as key elements for structural efficiency in virtually all thin-walled artifacts, including automotive tailgate panels, aircraft fuselage and wing skins, electronic device housings, and ship decks \cite{cheng1981investigation, lam2003automated, lee2025bayesian}.
A rib is a thin wall protruding from one side of a plate, which substantially enhances local bending stiffness while minimizing weight increase.
However, determining the optimal rib layout is a combinatorial, knowledge-intensive design task: the number, orientation, height, spacing, and branching topology of ribs all constitute design variables, producing an especially challenging optimization landscape \cite{ding2004stiffener}.
Nature offers many efficient rib-like patterns (for example, leaf venation, trabecular bone, and dragonfly wing venation), many of which may not lie near the local optima identified by SIMP and are therefore unlikely to be reached by its local search.
Multi-start strategies that launch optimization from multiple random initializations can mitigate the problem, but the exploration remains unstructured and the computational cost scales linearly.
A method that escapes local optima while preserving physical optimality is therefore needed.

What conventional methods lack is a way to make design knowledge explicit and machine-interpretable, so that an optimizer can act on it directly.
Expert designers draw on many structural patterns for rib layout, such as radial webs, branching venation, and cellular tessellations from other disciplines and from nature, yet these patterns have lacked a representation the optimizer can consume.
The rib networks sought here are themselves nature-inspired, approximating the load-bearing geometries seen in leaf venation and trabecular bone.
A diffusion model trained on billions of natural images therefore already encodes much of the structural vocabulary that rib design draws on, which makes it a well-matched, and so far unused, source of design knowledge for this task.
Communicating design intent through generated text has long been studied in collaborative engineering design \cite{cheng2019role}, and large language models are rapidly expanding the role of natural language in mechanics, product design, and manufacturing \cite{mustapha2025survey}. What has been missing is a mechanism that turns such linguistic intent into physics-coupled optimization gradients.
Parametric templates capture only a narrow, pre-enumerated slice of this knowledge, and data-driven generative models internalize it implicitly in network weights tied to a single training distribution (Section~\ref{sec:related}).
This work instead represents structural design intent as a natural-language prompt. The visual prior of a frozen text-to-image diffusion model \cite{rombach2022high} is distilled through score distillation sampling (SDS) \cite{poole2022dreamfusion} and coupled to the finite element sensitivity of SIMP at the gradient level.
The prompt thereby becomes a direct input to the optimization: an explicit knowledge representation whose physical consequences are filtered, at every iteration, by finite element analysis (FEA).
No prior study has applied SDS to density-based topology optimization with generative priors coupled to FEA at the gradient level.

The framework proposed here addresses this gap through three contributions.

\textbf{First, this constitutes the first application of score distillation sampling to physics-constrained topology optimization.}
By coupling a frozen Stable Diffusion 2.1 \cite{rombach2022high} to SIMP at the gradient level, the framework enables text-guided structural design without any training data or fine-tuning, a capability absent from existing data-driven generative methods.
\rev{The framework is training-free. It needs no task-specific training, fine-tuning, or labeled topology-optimization data. Instead, it builds on a large-scale, publicly pretrained diffusion model whose weights stay frozen throughout optimization.}
\rev{Retargeting the generative prior to a new domain, loading condition, or physics objective needs no model retraining, though each new engineering problem still requires its own geometry, boundary and loading conditions, physical model, constraints, and numerical parameters.} This extends the training-free paradigm of DreamFusion \cite{poole2022dreamfusion} from vision to structural engineering and turns a pretrained foundation model into a reusable knowledge source for engineering design.
\rev{The prompt is the only change to the generative prior. As in any topology-optimization study, each new problem still requires the design domain, boundary and loading conditions, physical model, volume fraction, and numerical parameters, including the SDS weight $\lambda_\text{sds}^0$ (Section~\ref{sec:sds}). The retargeting cost is therefore much lower than for data-driven models that require retraining.}

\textbf{Second, dead-end suppression is identified as the predominant structural signature of SDS-guided improvement.}
Cross-domain morphological analysis across six physics settings reveals a positive correlation between skeleton endpoint count and compliance (Pearson $r = +0.56$ to $+0.99$, median $r=+0.97$), indicating that the generative prior improves designs through one physically interpretable mechanism.
This cross-domain finding clarifies the mechanism: in most domains the gain comes from removing non-load-bearing dead ends, not from adding branches.

\textbf{Third, the framework diagnoses and resolves a pronounced gray-density tendency in the SDS--SIMP coupling and is validated for generality and scalability.}
\rev{Heaviside projection with $\beta$-continuation is itself a standard tool in density-based optimization. The contribution here is not the projection but the diagnosis that SDS induces a pronounced gray-density tendency (roughly doubling the intermediate-density ratio in the present SDS--SIMP coupling), resolved by synchronizing the $\beta$-schedule with the SDS schedule.} This reduces intermediate densities from $42.6\%$ to below $3\%$, and an automated skeleton-based pipeline converts the optimized density field directly into computer-aided design (CAD) geometry, i.e., a \rev{CAD-convertible candidate}.
Validation spans 245 primary SDS runs across four geometric domains (synthetic benchmarks through an industrial automotive suspension link) and two physics regimes (mechanical and thermoelastic bending). Of 49 primary prompt--domain combinations, 38 achieved statistically significant compliance reductions ($p=0.031$ for the mechanical subset), with gains up to $-31.5\%$ mechanical and $-23.0\%$ thermoelastic that consistently outperform all alternative baselines.
A negative result on heat conduction (Section~\ref{sec:discussion}) delineates the boundary of applicability.

The remainder of this paper is organized as follows.
Section~\ref{sec:related} reviews related work on data-driven generative topology optimization and score-based priors.
Section~\ref{sec:methodology} details the proposed framework, and Section~\ref{sec:setup} the experimental setup.
Section~\ref{sec:results} reports results across the four domains and two physics regimes, and Section~\ref{sec:mechanisms} analyzes the mechanisms underlying the observed improvements.
Section~\ref{sec:discussion} discusses implications and limitations, and Section~\ref{sec:conclusion} concludes.

\section{Related work}
\label{sec:related}

\subsection{Data-driven generative models for topology optimization}
\label{sec:related_datadriven}

Recent work has pursued data-driven generative models to overcome the local-optima limitation of classical topology optimization.
TopoDiff \cite{maze2023diffusion} trained a diffusion model on approximately 33,000 topology-optimization solutions, using von Mises stress, strain energy density, and volume fraction as conditioning channels, achieving an eightfold reduction in compliance error compared to earlier generative adversarial network-based approaches.
Diffusion Optimization Models (DOM) \cite{giannone2023aligning} learn entire SIMP optimization trajectories rather than single optimal designs, exploiting the structural similarity between iterative denoising and iterative convergence.
The Physics-Informed Diffusion Model (PIDM) \cite{bastek2024physics} integrates partial differential equation (PDE) residuals into training to reduce physical constraint violations by two orders of magnitude.
Extensions to three-dimensional (3D) problems \cite{kwon2025three, bekbolat2025diffusion}, to large-scale structural datasets \cite{yoo2026deepwheel, hong2025deepjeb}, and to manufacturability-aware deep generative design \cite{kim2024deep} have also been explored.
\rev{Learning-based components have also been integrated into topology optimization, including differentiable multiscale design \cite{wu2026machine} and network-based robust or concurrent formulations \cite{li2026level, li2025model, zhao2026novel}. These still rely on task-specific training or surrogate data tied to a particular design distribution, unlike the frozen, publicly pretrained prior coupled to FEA at the gradient level in the present work.}
Recent latent-diffusion pipelines \cite{zhang2025multistage, zhang2026mamba} still learn their priors from precomputed solutions and thus inherit the dataset dependence discussed below, whereas training-dataset-free reparameterization \cite{liu2026topology} avoids the dataset but injects no external design knowledge or language interface.

\rev{The representative data-driven methods considered here retain, to varying degrees, three limitations.}
First, they require massive domain-specific training datasets (tens of thousands of samples) tied to specific load and boundary conditions. Data-efficiency remedies such as geometric augmentation guided by engineering uncertainty \cite{kwon5380284three} alleviate but do not remove this dependence.
Second, they lose generalization when presented with unfamiliar conditions outside the training distribution.
Third, the generative model produces outputs independently of physical analysis, offering no intrinsic guarantee of compliance optimality.
\rev{These limitations apply unevenly, and several recent methods partially address them. PIDM \cite{bastek2024physics} embeds PDE residuals during training to improve physical consistency, and DOM \cite{giannone2023aligning} learns full optimization trajectories rather than static optima. Prior methods, however, incorporate physics primarily at training time. Among the methods reviewed in Tables~\ref{tab:comparison} and \ref{tab:approach_comparison}, none appears to couple a frozen text-to-image score gradient directly with an FEA sensitivity at every density-update iteration, so the evolving design is not tested against physical optimality during generation. The present framework closes this gap.}
Table~\ref{tab:comparison} contrasts the training and retargeting costs of these methods with the present framework.

\begin{table}[pos=htbp]
\centering
\caption{Comparison of data-driven diffusion-based topology optimization (TO) methods.}
\label{tab:comparison}
\setlength{\tabcolsep}{4pt}
\renewcommand{\arraystretch}{1.25}
\footnotesize
\begin{tabular}{@{}p{2.6cm}p{2.6cm}p{2.3cm}p{2.4cm}p{2.6cm}@{}}
\toprule
 & TopoDiff \cite{maze2023diffusion} & DOM \cite{giannone2023aligning} & PIDM \cite{bastek2024physics} & This work \\
\midrule
Training data & $\sim$33,000 TO solutions & $\sim$10,000 trajectories & PDE-specific & 0 (zero-shot) \\
Est.\ training cost & $\sim$2,750 GPU-hrs & $\sim$830 GPU-hrs & Physics training & 0 \\
New domain & Full rebuild & Full rebuild & Retraining required & \rev{No retraining} \\
New load condition & Full rebuild & Full rebuild & Retraining required & \rev{No retraining} \\
\bottomrule
\end{tabular}
\end{table}

\subsection{Score distillation and text-driven structural design}
\label{sec:related_sds}

In a parallel development in computer vision, score distillation sampling, introduced by DreamFusion \cite{poole2022dreamfusion}, demonstrated that the visual prior of a frozen text-to-image model (e.g., Stable Diffusion \cite{rombach2022high}) can be distilled into a 3D optimization loop, a Neural Radiance Field (NeRF) \cite{mildenhall2021nerf}, without any training data, achieving high-quality text-to-3D generation.
The core principle is that the score function of the pretrained diffusion model yields pixel-level gradients that indicate how the current output should be modified to better align with the text prompt.
SDS-type score priors have also been applied to inverse problems such as medical imaging \cite{feng2023score, chung2023solving, song2023pseudoinverse}, similarly leveraging diffusion knowledge to structure an optimization problem.
Separately, Contrastive Language-Image Pre-training (CLIP)-based topology optimization \cite{zhong2023topology} used contrastive embeddings to impose text-based styles on optimized structures, but CLIP provides only a global image--text similarity measure and cannot generate spatially detailed generative gradients at the pixel level.
In product design more broadly, knowledge-graph-guided semantic diffusion has been used for bio-inspired form generation \cite{wang2024bioinspired}, though without coupling to physical analysis.
Closest in spirit to the present work, the recently proposed LMTO framework \cite{liang2025integrating} couples a large visual--language model (LVLM) to conceptual topology optimization, semantically decomposing candidate structures within the large model's knowledge space and steering the optimization toward human preference.
LMTO and the present framework are complementary: LMTO uses the large model interpretively at the semantic level, whereas the present framework distills the generative prior into pixel-level score gradients summed with the FEA sensitivity at every density update. The two could in principle be combined. Table~\ref{tab:approach_comparison} contrasts all text- and diffusion-based approaches.

\begin{table}[pos=htbp]
\centering
\caption{Comparison of diffusion- and text-based approaches for structural design.}
\label{tab:approach_comparison}
\setlength{\tabcolsep}{4pt}
\renewcommand{\arraystretch}{1.2}
\footnotesize
\begin{tabular}{@{}p{2.1cm}p{2.4cm}p{1.9cm}p{2.1cm}p{2.3cm}p{2.3cm}@{}}
\toprule
& Data-driven diffusion TO \cite{maze2023diffusion, giannone2023aligning} & SDS for vision \cite{poole2022dreamfusion} & CLIP-based TO \cite{zhong2023topology} & LVLM-based TO \cite{liang2025integrating} & This work \\
\midrule
Training data & Tens of thousands of TO solutions & Not required & Not required & Not required & Not required \\
Physics integration & Condition channel (passive) & None & Compliance (objective) & Preference-weighted TO loop & FEA sensitivity (active grad.) \\
Design guidance & Conditional generation & Text $\rightarrow$ 3D & Global similarity (CLIP) & Semantic decomposition \& preference weighting & Pixel-level score gradient \\
Output / domain & Pixel density (TO) & NeRF (3D vision) & Density field (TO style) & Density field (conceptual TO) & Element density (rib design) \\
New-condition adaptability & Full retraining & High (prompt) & High (prompt) & High (concept) & \rev{High (no retraining)} \\
\bottomrule
\end{tabular}
\end{table}

\subsection{Research gap}
\label{sec:related_gap}

Domain-specific generative models remain valuable when abundant labeled data exist, but their applicability is tied to the training distribution: a model trained under specific boundary conditions cannot transfer to other physics problems or unfamiliar domain geometries without retraining.
Worse, because generated designs are produced \rev{at generation time} without active coupling to an FEA solver, nothing guarantees they satisfy physical equilibrium, meet volume constraints, or approach optimality by any quantitative criterion. \rev{Even methods that embed physics during training} may yield outputs that are visually plausible yet physically suboptimal.
From a knowledge-representation standpoint, the design knowledge such models encode is implicit, non-transferable, and inaccessible to the engineer at design time.
In summary, existing approaches fall into three groups: those that embed design knowledge implicitly in retraining-bound network weights, those that impose text guidance without spatially detailed physics-coupled gradients, and those that use a large model interpretively to weight the optimization toward semantic preference. Gradient-level coupling between a frozen text-to-image prior and an FEA sensitivity remains unexplored for density-based topology optimization. Closing this gap is the focus of the present work.

\section{Methodology}
\label{sec:methodology}

\begin{figure}[pos=htbp]
\centering
\includegraphics[width=\textwidth]{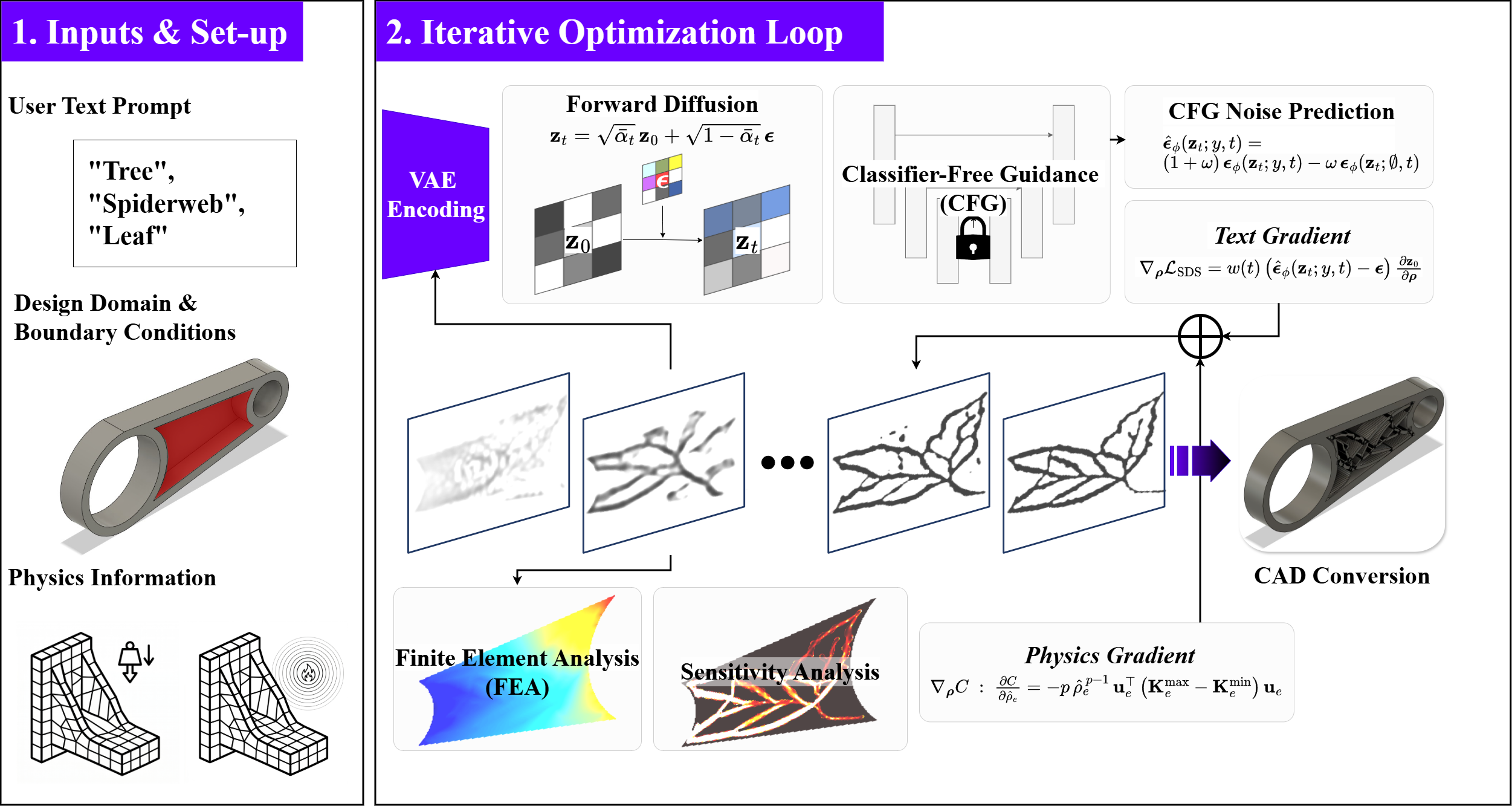}
\caption{Computational framework for text-guided, physics-constrained topology optimization. Three stages: \textbf{(i) Inputs \& Set-up} (text prompt, design domain, boundary conditions, volume fraction; frozen Stable Diffusion 2.1 as generative prior); \textbf{(ii) Iterative Optimization Loop} coupling a Mindlin--Reissner plate FEA physics gradient with an SDS text gradient under four-stage scheduling; \textbf{(iii) CAD Conversion} of the converged density field into \rev{CAD-convertible candidate} geometry.}
\label{fig:framework}
\end{figure}

The framework integrates two gradient sources into a single optimization loop that updates the density field $\hat{\rho}$ (Fig.~\ref{fig:framework}).
The \textbf{Inputs \& Set-up} phase specifies a user text prompt, the design domain with boundary conditions, and the physics information.
At each iteration of the \textbf{Iterative Optimization Loop}, a Mindlin--Reissner plate FEA solver produces the physics gradient $\nabla_\rho C$.
In parallel, the current density field is encoded into the latent space of a frozen Stable Diffusion 2.1, forward diffusion adds noise $\boldsymbol{\epsilon}$, and the U-Net predicts this noise under classifier-free guidance (CFG). The discrepancy between predicted and actual noise yields the SDS text gradient $\nabla_\rho \mathcal{L}_\text{SDS}$ via backpropagation.
The two gradients combine as a weighted sum, Eq.~\eqref{eq:combined}, under a four-stage schedule governing their interaction:
\begin{itemize}
\setlength\itemsep{0pt}
\item \textbf{Warmup} ($k_\text{warm}=30$): only the physics gradient acts, stabilizing initial load paths.
\item \textbf{Cosine cooldown} ($k_\text{cool}=500$): the SDS weight $\lambda_\text{sds}(k)$ decays gradually to zero.
\item \textbf{Timestep annealing}: the diffusion timestep $t$ linearly decreases from $t_\text{max}=0.98$ to $t_\text{min}=0.50$ for coarse-to-fine structural guidance.
\item \textbf{Exponential moving average (EMA) smoothing}: an EMA of the SDS gradient damps single-sample stochastic noise.
\end{itemize}
The \textit{Adam} optimizer updates the density field under a bisection-based volume constraint and Heaviside projection with $\beta$-continuation for binarization.
In the \textbf{CAD Conversion} stage, the converged density field is automatically processed by binarization, medial-axis thinning, B\'ezier curve fitting, and extrusion to yield \rev{CAD-convertible candidate} geometries.

\subsection{Problem formulation}
\label{sec:formulation}

The design problem is formulated as a two-dimensional (2D) density-based topology optimization.
The design domain $\Omega$ is discretized into an $n_\text{elx}\times n_\text{ely}$ finite element mesh, with a continuous density variable $\rho_e\in[0,1]$ assigned to each element.
The optimization problem minimizes structural compliance subject to a volume fraction constraint:
\begin{equation}
\begin{aligned}
\min_{\boldsymbol{\rho}} \quad & C(\boldsymbol{\rho}) = \mathbf{F}^\top \mathbf{U} = \sum_{e=1}^{N} (\hat{\rho}_e)^p \, \mathbf{u}_e^\top \mathbf{k}_e \, \mathbf{u}_e \\
\text{s.t.} \quad & \mathbf{K}(\hat{\boldsymbol{\rho}})\,\mathbf{U} = \mathbf{F}, \\
& \frac{V(\hat{\boldsymbol{\rho}})}{|\Omega|} = V^*, \\
& 0 \leq \rho_e \leq 1 \quad \forall\, e
\end{aligned}
\label{eq:optimization}
\end{equation}
where $C$ is the compliance, $\mathbf{K}$, $\mathbf{U}$, and $\mathbf{F}$ are the global stiffness matrix, nodal displacement vector, and external force vector, respectively, and $V^*$ is the target volume fraction.
The density field passes through three stages. The \textit{raw design variables} $\boldsymbol{\rho}$ are updated directly by the optimizer. The \textit{filtered density} $\tilde{\boldsymbol{\rho}} = \mathcal{F}(\boldsymbol{\rho})$ is obtained by a spatial averaging density filter \cite{bourdin2001filters, bruns2001topology} with radius $r_\text{min}=5$ that removes checkerboard patterns\rev{. Because this radius is fixed in physical units, the filter also limits mesh-dependent features (Appendix~\ref{app:mesh})}. The \textit{physical density} $\hat{\boldsymbol{\rho}} = \mathcal{H}(\tilde{\boldsymbol{\rho}})$ is produced by a smoothed Heaviside projection \cite{guest2004achieving} that drives elements toward near-binary $\{0,1\}$ material states.
The sequence $\boldsymbol{\rho} \xrightarrow{\mathcal{F}} \tilde{\boldsymbol{\rho}} \xrightarrow{\mathcal{H}} \hat{\boldsymbol{\rho}}$ is fully differentiable. Compliance, stiffness, and all sensitivities are evaluated on the physical density $\hat{\boldsymbol{\rho}}$, and chain-rule backpropagation carries the gradient back to $\boldsymbol{\rho}$.
Under SIMP interpolation \cite{bendsoe1999material}, the element stiffness is computed as $\mathbf{K}_e(\hat{\rho}_e) = \hat{\rho}_e^p\,\mathbf{K}_e^\text{max} + (1-\hat{\rho}_e^p)\,\mathbf{K}_e^\text{min}$ with penalty $p=3$, where $\mathbf{K}_e^\text{max}$ and $\mathbf{K}_e^\text{min}$ correspond to elements with and without rib reinforcement, respectively.
Since base-plate regions retain physical substance ($\mathbf{K}_e^\text{min}>0$), the formulation avoids the unphysical assumption that zero-density regions bear no load.
Detailed derivations of the SIMP interpolation, density filter, and Heaviside projection are provided in Appendices~\ref{app:simp}, \ref{app:filter}, and \ref{app:heaviside}.

\subsection{Finite element analysis}
\label{sec:fea}

The three plate-bending domains ($D_{\text{Rect}}$, $D_{\text{Circle}}$, $D_{\text{Hole}}$) are analyzed using Mindlin--Reissner plate theory \cite{mindlin1951influence, reissner1945effect} with the Mixed Interpolation of Tensorial Components (MITC4) element \cite{bathe1985four}.
The Mindlin--Reissner theory is adopted because rib-reinforced plates behave locally as thick plates in the rib regions, where transverse shear deformation ($\boldsymbol{\gamma} \neq \mathbf{0}$) cannot be neglected as in the Kirchhoff thin-plate theory.
The MITC4 element is a 4-node bilinear quadrilateral with three degrees of freedom (DOF) per node ($w, \theta_x, \theta_y$), whose element stiffness matrix comprises bending and shear contributions:
\begin{equation}
\mathbf{K}_e = \int_{\Omega_e} \mathbf{B}_b^\top \mathbf{D}_b \mathbf{B}_b \,d\Omega + \int_{\Omega_e} \mathbf{B}_s^{\text{MITC}\,\top} \mathbf{D}_s \mathbf{B}_s^{\text{MITC}} \,d\Omega
\label{eq:element_stiffness}
\end{equation}
where $\mathbf{D}_b$ is the bending constitutive matrix and $\mathbf{D}_s$ the shear constitutive matrix (with Reissner correction factor $\kappa_s=5/6$).
The MITC4 formulation overcomes shear locking by evaluating shear strains at four tying points on the element edges and interpolating them into the interior, yielding accurate solutions even in the thin-plate limit.

In the rib-reinforced plate model, zero-density regions retain the physical substance of the base plate rather than being void.
A modified SIMP interpolation directly interpolates bending and shear stiffnesses: regions with ribs have effective thickness $t_\text{max}$ (base plate + rib height), while regions without ribs have only $t_\text{min}$ (base plate).
Since flexural rigidity scales as $t^3$ and shear stiffness as $t$, the element stiffness is interpolated as $\mathbf{K}_e(\hat{\rho}_e) = \hat{\rho}_e^p\,\mathbf{K}_e^\text{max} + (1-\hat{\rho}_e^p)\,\mathbf{K}_e^\text{min}$, where $\mathbf{K}_e^\text{max}$ and $\mathbf{K}_e^\text{min}$ can be precomputed.
The industrial $D_{\text{Link}}$ domain is analyzed under plane stress using a standard 4-node bilinear quadrilateral element with two degrees of freedom per node ($u$, $v$).

In both formulations, the compliance sensitivity exploits self-adjointness:
$\partial C/\partial\hat{\rho}_e = -p\hat{\rho}_e^{p-1}\,\mathbf{u}_e^\top(\mathbf{K}_e^\text{max}-\mathbf{K}_e^\text{min})\,\mathbf{u}_e$, where the quantity $\mathbf{u}_e^\top(\mathbf{K}_e^\text{max}-\mathbf{K}_e^\text{min})\,\mathbf{u}_e$ represents the strain energy associated with the stiffness gain from adding a rib to element $e$.
Elements with larger values yield greater compliance reduction upon rib placement.
The sensitivity is extended to the raw design variables via the chain rule through the Heaviside projection and density filter.
Checkerboard patterns and mesh dependency are suppressed by the density filter ($r_\text{min}=5$), which spatially averages raw densities within a neighborhood and provides indirect control of the minimum feature size.
A companion maximum-thickness filter ($r_\text{max}=10$) additionally caps local rib thickness for manufacturability through a quadratic local-volume penalty added to the physics update. Its formulation and sensitivity are given in Appendix~\ref{app:rmax}.

For the thermoelastic experiments, the same MITC4 plate FEA is used with a different load vector:
a prescribed uniform through-thickness temperature gradient $\Delta T$ induces thermal bending moments $\mathbf{M}_\text{th} = \alpha_T \Delta T [1, 1, 0]^\top$, with thermal expansion coefficient $\alpha_T$, on every element, which are converted to equivalent nodal forces $\mathbf{F}_\text{th}$ via Gauss integration.
\rev{In contrast to the design-dependent thermal loads of material-based thermoelastic formulations \cite{rodrigues1995thermoelastic}, a simplified prescribed-thermal-moment weak coupling is adopted. The thermal moment $\mathbf{M}_\text{th}$ is set by the prescribed $\Delta T$ alone and does not scale with the element stiffness $E(\hat{\rho})$. The assembled load $\mathbf{F}_\text{th}$ is therefore density-independent ($\partial \mathbf{F}_\text{th}/\partial \hat{\rho} = \mathbf{0}$), so the compliance sensitivity keeps its standard self-adjoint form, $\partial C/\partial \hat{\rho}_e = -\mathbf{u}_e^\top (\partial \mathbf{K}_e/\partial \hat{\rho}_e)\, \mathbf{u}_e$. A finite-difference check is provided in Appendix~\ref{app:fd_thermo}.}
Full derivations of the element formulations, constitutive matrices, and sensitivity are given in Appendices~\ref{app:fea}, \ref{app:simp}, and \ref{app:sensitivity}.

\subsection{Score distillation sampling for topology optimization}
\label{sec:sds}

\begin{figure}[pos=htbp]
\centering
\includegraphics[width=\textwidth]{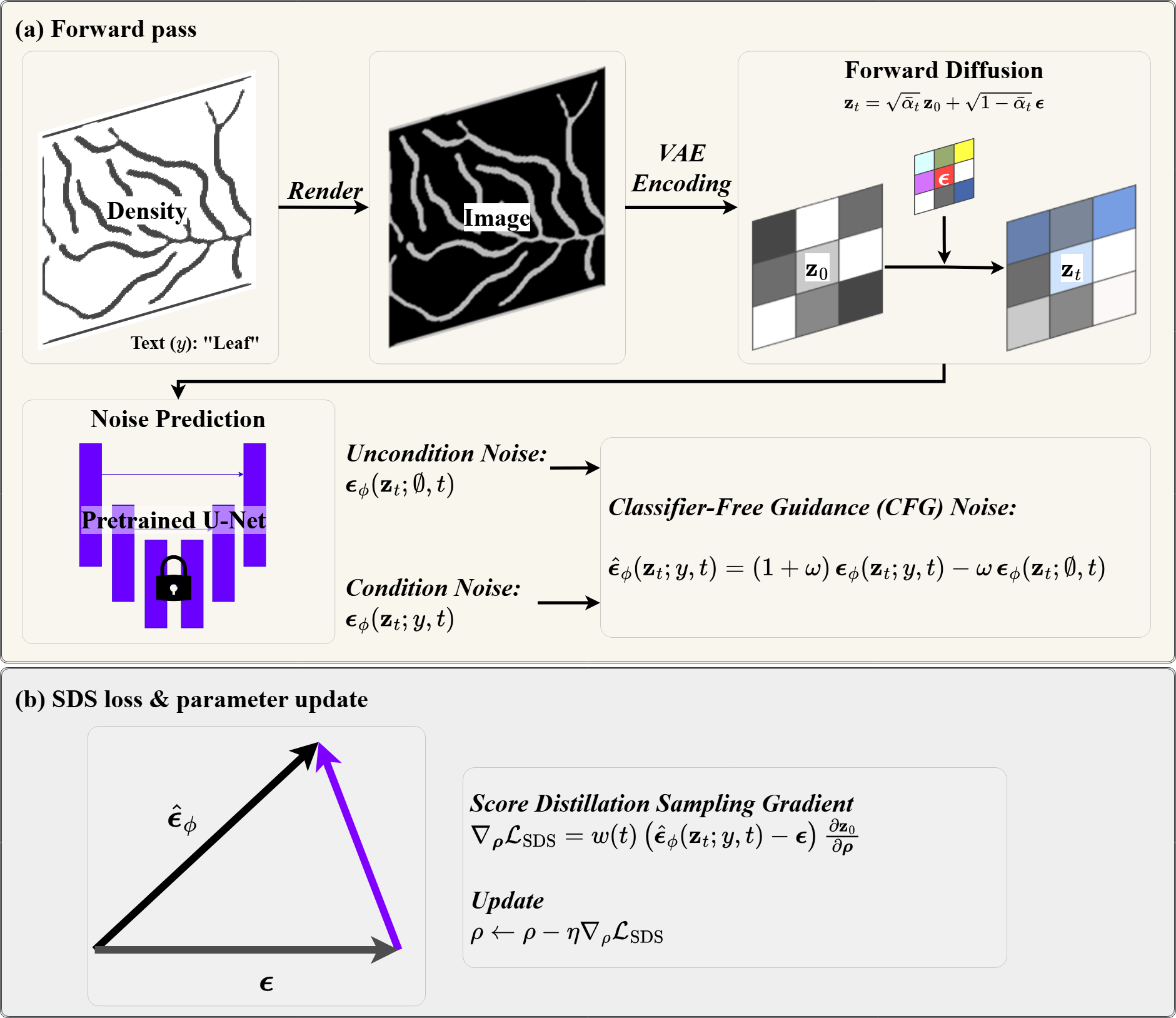}
\caption{SDS pipeline for topology optimization. (a) Forward pass: the density field $\hat{\boldsymbol{\rho}}$ is rendered as a grayscale image and encoded into the latent $\mathbf{z}_0$ by the frozen VAE; Gaussian noise $\boldsymbol{\epsilon}$ is added at a sampled diffusion timestep $t$ to produce $\mathbf{z}_t$; the frozen U-Net predicts the noise under both the text-conditioned input (prompt $y$) and the unconditional (null) input; the two predictions are combined via CFG with weight $\omega$ to form $\hat{\boldsymbol{\epsilon}}_\phi$. (b) SDS loss and parameter update: the difference $\hat{\boldsymbol{\epsilon}}_\phi - \boldsymbol{\epsilon}$ is weighted by $w(t)=1-\bar{\alpha}_t$ and backpropagated through the differentiable encode--render chain $\partial \mathbf{z}_0 / \partial \boldsymbol{\rho}$ to yield the SDS gradient $\nabla_{\boldsymbol{\rho}}\mathcal{L}_\text{SDS}$, which is combined with the physics gradient in the final density update.}
\label{fig:sds_process}
\end{figure}

The physical density field $\hat{\boldsymbol{\rho}}\in\mathbb{R}^{n_\text{elx}\times n_\text{ely}}$ is converted to a three-channel RGB image by replicating the single grayscale channel, resized to $512\times512$ via bilinear interpolation, and mapped to the latent space by the encoder $\mathcal{E}$ of the frozen variational autoencoder (VAE) of Stable Diffusion 2.1, giving the latent code $\mathbf{z}_0 = \mathcal{E}(\mathbf{x})\in\mathbb{R}^{4\times64\times64}$.
This entire pipeline is differentiable, enabling backpropagation to $\hat{\boldsymbol{\rho}}$ (Appendix~\ref{app:density_image}).
Fig.~\ref{fig:sds_process} illustrates the full SDS pipeline.

Forward diffusion adds Gaussian noise to the latent vector at a randomly sampled timestep $t$:
\begin{equation}
\mathbf{z}_t = \sqrt{\bar{\alpha}_t}\,\mathbf{z}_0 + \sqrt{1-\bar{\alpha}_t}\,\boldsymbol{\epsilon}, \quad \boldsymbol{\epsilon}\sim\mathcal{N}(\mathbf{0},\mathbf{I})
\label{eq:forward}
\end{equation}
where $\bar{\alpha}_t = \prod_{s=1}^t(1-\beta_s)$ is the cumulative noise schedule \cite{ho2020denoising}.
The U-Net noise predictions for the text prompt $y$ and the null condition are combined via CFG \cite{ho2022classifier}:
\begin{equation}
\hat{\boldsymbol{\epsilon}}_\phi(\mathbf{z}_t, y, t) = (1+\omega)\,\boldsymbol{\epsilon}_\phi(\mathbf{z}_t, y, t) - \omega\,\boldsymbol{\epsilon}_\phi(\mathbf{z}_t, \varnothing, t)
\label{eq:cfg}
\end{equation}
with guidance scale $\omega=100$.
The SDS gradient is computed with a stop-gradient on the U-Net Jacobian:
\begin{equation}
\nabla_{\boldsymbol{\rho}}\mathcal{L}_\text{SDS} = w(t)\left(\hat{\boldsymbol{\epsilon}}_\phi(\mathbf{z}_t, y, t) - \boldsymbol{\epsilon}\right)\frac{\partial\mathbf{z}_0}{\partial\boldsymbol{\rho}}
\label{eq:sds}
\end{equation}
where $w(t) = 1-\bar{\alpha}_t$ is a time-dependent weight.
The physics and SDS gradients are combined as a weighted sum:
\begin{equation}
\nabla_{\boldsymbol{\rho}}\mathcal{L}_\text{total} = \lambda_\text{phys}\,\nabla_{\boldsymbol{\rho}} C + \lambda_\text{sds}(k)\,\nabla_{\boldsymbol{\rho}} \mathcal{L}_\text{SDS}
\label{eq:combined}
\end{equation}
For clarity, Eq.~\eqref{eq:combined} shows only the two principal gradients. The physics term additionally carries the maximum-thickness penalty gradient $\nabla_{\boldsymbol{\rho}}\mathcal{L}_\text{max}$ used for manufacturability (Appendix~\ref{app:rmax}).
The weights are $\lambda_\text{phys}=1.0$, while $\lambda_\text{sds}^0$ is scaled to the magnitude of the physics loss so that the generative signal stays competitive in the raw weighted sum. This sets $\lambda_\text{sds}^0=10$ for the mechanical point-load domains $D_{\text{Rect}}$, $D_{\text{Circle}}$, and $D_{\text{Link}}$ (baseline compliance $\sim\!3$--$9$), $\lambda_\text{sds}^0=50$ for $D_{\text{Hole}}$ (baseline compliance $\sim\!63$, roughly seven times higher due to 4-sided loading around the central hole), and $\lambda_\text{sds}^0=100$ for thermoelastic compliance (baseline compliance $\sim\!600$, with distributed thermal moments on every element. Empirical sweep in Appendix~\ref{app:thermo_lambda} and Table~\ref{tab:ablation}).
In each case, raising $\lambda_\text{sds}^0$ proportionally to the physics-loss scale preserves a comparable relative influence between the physics and SDS gradients in the raw-sum update.
\rev{Eq.~\eqref{eq:combined} is a heuristic gradient-blending update rather than the exact gradient of a single scalar objective, with $\lambda_\text{sds}(k)$ setting the trade-off between the compliance and generative terms \cite{poole2022dreamfusion} (the SDS term uses a stop-gradient approximation and is not the exact gradient of a scalar loss). Its behaviour is controlled empirically through gradient-scale matching, a decaying SDS weight, exact volume projection, and a final physics-only convergence phase: the cosine cooldown drives $\lambda_\text{sds}(k)\!\to\!0$, so the final iterations follow the pure physics gradient toward a physically consistent optimum.}

\subsection{Scheduling strategy}
\label{sec:scheduling}

Naively applying the SDS gradient at constant intensity throughout optimization causes two problems:
(1) before the physics solver establishes load paths in the early stages, SDS distorts the density field and generates unphysical patterns, and (2) in the final stages, stochastic SDS noise interferes with fine convergence.
A four-stage scheduling strategy is therefore designed.

During the warmup phase (iterations $1$ to $k_\text{warm}=30$), the SDS gradient is deactivated and only the physics gradient is used, forming a basic material distribution between external forces and supports that SDS subsequently refines.
Since the Heaviside projection is weak at $\beta=1$ during the first 50 iterations, the density field evolves smoothly and exploratorily during this period.

After warmup, the SDS weight follows a cosine cooldown:
\begin{equation}
\lambda_\text{sds}(k) = \lambda_\text{sds}^0 \cdot \frac{1}{2}\left(1 + \cos\left(\frac{\pi\cdot\min(k-k_\text{warm},\,k_\text{cool})}{k_\text{cool}}\right)\right)
\label{eq:cooldown}
\end{equation}
with $k_\text{cool}=500$, so that the SDS influence converges to zero after approximately 530 iterations and the final convergence proceeds with the physics gradient alone.
Cosine decay maintains the SDS influence longer in the early stages compared to linear decay while decreasing sharply in the later stages, enabling a natural transition from exploration (SDS-guided) to exploitation (physics-driven).

Simultaneously, the diffusion timestep is linearly annealed:
\begin{equation}
t(k) = t_\text{max} - (t_\text{max}-t_\text{min})\cdot\frac{k-k_\text{warm}}{k_\text{cool}}
\label{eq:anneal}
\end{equation}
from $t_\text{max}=0.98$ to $t_\text{min}=0.50$, implementing a coarse-to-fine strategy \cite{tang2023stable}.
Noise predictions at high $t$ induce global geometry (the coarse arrangement of the overall pattern, symmetry, and connectivity), while predictions at low $t$ refine local details such as branching points, member thickness, and boundary smoothness.

Finally, the SDS gradient is stabilized by EMA smoothing:
\begin{equation}
\mathbf{g}_\text{ema}^{(k)} = \alpha\,\mathbf{g}_\text{ema}^{(k-1)} + (1-\alpha)\,\nabla\mathcal{L}_\text{SDS}^{(k)}
\label{eq:ema}
\end{equation}
with $\alpha=0.9$, averaging gradients over approximately the most recent 10 iterations to damp the stochastic variance inherent in single-sample SDS gradient estimation (Appendix~\ref{app:ema}).

\subsection{Optimization loop and implementation}
\label{sec:optimization_loop}
The \textit{Adam} optimizer \cite{kingma2014adam} is used with momentum coefficients $\beta_1=0.5$ and $\beta_2=0.9$ and learning rate annealed from 0.04 to 0.01.
\textit{Adam} is preferred over the conventional Optimality Criteria (\textit{OC}) update for two reasons: it accommodates arbitrary gradient combinations (physics + SDS), and it converges stably in conjunction with the Heaviside projection, whereas \textit{OC} exhibits period-2 oscillation at high $\beta$ that prevents convergence (Appendix~\ref{app:optimizer}; quantitative baseline comparison in Section~\ref{sec:baselines}).
A bisection projection enforces the volume fraction constraint exactly at each iteration (Appendix~\ref{app:volume}).
Algorithm~\ref{alg:zeroshot} summarizes the complete optimization loop, and Table~\ref{tab:parameters} collects the hyperparameter values used throughout the study.

\begin{algorithm}[htbp]
\caption{Zero-Shot Rib Design}
\label{alg:zeroshot}
\begin{algorithmic}[1]
\REQUIRE Domain $\Omega$, boundary conditions, $V^*$, text prompt $y$
\ENSURE Optimized density field $\hat{\boldsymbol{\rho}}$
\STATE Initialize $\rho_e = V^*$ for all $e \in \Omega_\text{active}$ \COMMENT{Uniform density}
\STATE Load pretrained Stable Diffusion 2.1 (all weights frozen)
\STATE $\beta \leftarrow 1.0$ \COMMENT{Initial Heaviside sharpness}
\STATE $\mathbf{g}_\text{ema} \leftarrow \mathbf{0}$ \COMMENT{EMA buffer}
\FOR{$k = 1$ to $K = 1000$}
  \STATE $\tilde{\boldsymbol{\rho}} \leftarrow \text{DensityFilter}(\boldsymbol{\rho}, r_\text{min})$
  \STATE $\hat{\boldsymbol{\rho}} \leftarrow \text{HeavisideProject}(\tilde{\boldsymbol{\rho}}, \beta)$
  \STATE Solve $\mathbf{K}(\hat{\boldsymbol{\rho}})\mathbf{u} = \mathbf{f}$ via sparse Cholesky \COMMENT{FEA}
  \STATE $\mathbf{g}_\text{phys} \leftarrow \partial C / \partial \boldsymbol{\rho} + \partial \mathcal{L}_\text{max} / \partial \boldsymbol{\rho}$ \COMMENT{compliance + max-thickness penalty (App.~\ref{app:rmax})}
  \IF{$k > k_\text{warm}$ \AND $\lambda_\text{sds}(k) > 0$}
    \STATE $\mathbf{x} \leftarrow \text{resize}(\text{stack\_rgb}(\hat{\boldsymbol{\rho}}),\; 512 \times 512)$
    \STATE $\mathbf{z}_0 \leftarrow \mathcal{E}(\mathbf{x})$ \COMMENT{VAE encoding}
    \STATE $t \leftarrow \text{anneal}(k,\; t_\text{max},\; t_\text{min})$
    \STATE $\boldsymbol{\epsilon} \sim \mathcal{N}(\mathbf{0}, \mathbf{I})$
    \STATE $\mathbf{z}_t \leftarrow \sqrt{\bar{\alpha}_t}\, \mathbf{z}_0 + \sqrt{1 - \bar{\alpha}_t}\, \boldsymbol{\epsilon}$ \COMMENT{Forward diffusion}
    \STATE $\hat{\boldsymbol{\epsilon}} \leftarrow (1 + \omega)\,\boldsymbol{\epsilon}_\phi(\mathbf{z}_t, y, t) - \omega \cdot \boldsymbol{\epsilon}_\phi(\mathbf{z}_t, \varnothing, t)$ \COMMENT{CFG}
    \STATE $\mathbf{g}_\text{sds}^\text{raw} \leftarrow w(t)(\hat{\boldsymbol{\epsilon}} - \boldsymbol{\epsilon})\, \partial \mathbf{z}_0 / \partial \boldsymbol{\rho}$ \COMMENT{SDS gradient}
    \STATE $\mathbf{g}_\text{ema} \leftarrow \alpha \cdot \mathbf{g}_\text{ema} + (1 - \alpha) \cdot \mathbf{g}_\text{sds}^\text{raw}$
    \STATE $\lambda_\text{sds} \leftarrow \text{CosineDecay}(k,\; \lambda_\text{sds}^0,\; k_\text{cool})$
    \STATE $\mathbf{g}_\text{sds} \leftarrow \mathbf{g}_\text{ema}$
  \ELSE
    \STATE $\mathbf{g}_\text{sds} \leftarrow \mathbf{0};\quad \lambda_\text{sds} \leftarrow 0$
  \ENDIF
  \STATE $\mathbf{g} \leftarrow \lambda_\text{phys} \cdot \mathbf{g}_\text{phys} + \lambda_\text{sds} \cdot \mathbf{g}_\text{sds}$ \COMMENT{Weighted sum of raw gradients}
  \STATE $\boldsymbol{\rho} \leftarrow \textit{Adam}(\boldsymbol{\rho}, \mathbf{g}, lr(k))$
  \STATE $\boldsymbol{\rho} \leftarrow \text{BisectionProject}(\boldsymbol{\rho},\; V^*)$
  \STATE $\boldsymbol{\rho} \leftarrow \text{clip}(\boldsymbol{\rho},\; 0,\; 1)$
  \IF{$k \bmod N_\beta = 0$}
    \STATE $\beta \leftarrow \min(2\beta,\; \beta_\text{max})$
  \ENDIF
\ENDFOR
\RETURN $\hat{\boldsymbol{\rho}}$
\end{algorithmic}
\end{algorithm}

\begin{table}[pos=htbp]
\centering
\caption{Summary of experimental parameters.}
\label{tab:parameters}
\renewcommand{\arraystretch}{1.25}
\begin{tabular}{@{}lll@{}}
\toprule
Parameter & Value & Description \\
\midrule
Grid resolution & $150 \times 150$ & FE mesh \\
Volume fraction $V^*$ & 0.20 (0.30 for $D_{\text{Link}}$) & Target volume \\
SIMP penalization $p$ & 3.0 & Material interpolation \\
Filter radius $r_\text{min}$ & 5.0 & Density filter \\
Max-thickness radius $r_\text{max}$ & 10.0 & Local-volume penalty (App.~\ref{app:rmax}) \\
Heaviside $\beta$ & $1.0 \to 64.0$ & Doubling every 50 iter \\
Optimizer & \textit{Adam} ($\beta_1\!=\!0.5$, $\beta_2\!=\!0.9$) & Gradient descent \\
Learning rate & $0.04 \to 0.01$ & Cosine annealing \\
CFG scale $\omega$ & 100 & Guidance strength \\
\multirow{3}{*}{$\lambda_\text{phys}$ / $\lambda_\text{sds}^0$}
 & 1.0 / 10 \quad (mech $D_{\text{Rect}}$, $D_{\text{Circle}}$, $D_{\text{Link}}$) & \multirow{3}{*}{Gradient weighting} \\
 & 1.0 / 50 \quad ($D_{\text{Hole}}$) & \\
 & 1.0 / 100 \quad (thermoelastic) & \\
SDS warmup / cooldown & 30 / 500 iter & Scheduling \\
$t$-annealing & $0.98 \to 0.50$ & Coarse-to-fine \\
EMA decay $\alpha$ & 0.9 & Gradient smoothing \\
Iterations $K$ & 1000 & Total steps \\
\bottomrule
\end{tabular}
\end{table}

\subsection{Post-processing}
\label{sec:postprocessing}

The optimized density field is automatically converted into \rev{CAD-convertible candidate} geometry in the Standard for the Exchange of Product model data (STEP) format through a six-stage pipeline.
The continuous density field is binarized at threshold $\rho_\text{th}=0.5$. Zhang--Suen thinning \cite{zhang1984fast} then extracts the medial-axis skeleton, and a Euclidean distance transform computes local rib half-widths. The skeleton is converted to a topological graph with branch-point and endpoint nodes, each edge is fitted with piecewise cubic B\'{e}zier curves by least squares, and offset polygons of physically meaningful width are generated from the fitted centerlines. Finally, the 2D polygons are extruded into 3D solids and exported as AP214-schema STEP files.
The entire pipeline completes in 5 seconds on a single CPU core and requires no user parameters beyond the extrusion height.
\rev{The exported STEP solids are watertight and importable into standard CAD software as ``CAD-convertible candidates,'' their fabrication feasibility assessed downstream rather than verified here. A 3D solid re-analysis (Appendix~\ref{app:reanalysis}) confirms the 2D design rankings are largely preserved.}
Full pipeline details are provided in Appendix~\ref{app:postprocessing}.

\section{Experimental setup}
\label{sec:setup}

The three plate-bending domains ($D_{\text{Rect}}$, $D_{\text{Circle}}$, $D_{\text{Hole}}$) use Mindlin--Reissner plate theory with the MITC4 element (3 DOF/node) at $V^*=0.20$. The industrial $D_{\text{Link}}$ domain uses plane-stress 4-node bilinear elements (2 DOF/node) at $V^*=0.30$, with fixed supports on the left arc and combined tension--shear loading ($f_x=f_y=0.01$) on the right.
Domain geometries, load/support configurations, and representative baselines appear in Sections~\ref{sec:main_results} and \ref{sec:link}.

Ten text prompts span three semantic categories (Table~\ref{tab:prompts}): biomimetic (\textit{Tree}, \textit{Leaf}, \textit{Bone}), engineering (\textit{Truss}, \textit{Grid}, \textit{Honeycomb}), and geometric (\textit{Voronoi}, \textit{Spiderweb}, \textit{Ornamental}, \textit{Diamond}).
Every prompt carries the prefix ``bold black lines on white background,'' which bridges the density field's visual style to the diffusion model's training distribution and mitigates source-distribution mismatch \cite{mcallister2024rethinking}.
A negative prompt suppresses blurriness, color, 3D cues, and fine details below the filter radius.

\begin{table}[pos=htbp]
\centering
\caption{Classification of text prompts by structural characteristics.}
\label{tab:prompts}
\begin{tabular*}{\textwidth}{@{\extracolsep{\fill}} l c c c }
\toprule
& \textbf{Biomimetic} & \textbf{Engineering} & \textbf{Geometric} \\
\midrule
Prompts &
\makecell{\textit{Tree}, \textit{Leaf}, \\ \textit{Bone}} &
\makecell{\textit{Truss}, \textit{Grid}, \\ \textit{Honeycomb}} &
\makecell{\textit{Voronoi}, \textit{Spiderweb}, \\ \textit{Ornamental}, \textit{Diamond}} \\ [2.5ex]
Induced pattern &
\makecell{Hierarchical, \\ Fractal} &
\makecell{Regular cells, \\ Tessellated} &
\makecell{Irregular, \\ Radial} \\
\bottomrule
\end{tabular*}
\end{table}

Each run uses $N=1000$ iterations on an NVIDIA RTX 4090 (24 GB VRAM), taking $\sim$44 minutes per SDS run, a $\sim$3-minute ($\sim$7\%) overhead over the pure-SIMP baseline.
Full parameter values appear in Table~\ref{tab:parameters}.

Statistical significance for SDS-guided improvements uses a one-sided binomial test on the five random seeds: the minimum achievable $p$-value for 5/5 seeds beating the baseline is $p=1/2^5=0.031$.
The analysis is complemented by a one-sample effect size, Cohen's $d=(\bar{C}_\text{SDS}-C_\text{base})/s_\text{SDS}$ measured relative to the baseline, with a median $|d|=1.47$ across significant combinations (all corresponding to compliance reductions) indicating large practical effects.

\section{Results}
\label{sec:results}

\subsection{SDS-guided optimization discovers superior topologies}
\label{sec:main_results}

\begin{figure}[pos=htbp]
\centering
\includegraphics[width=\textwidth]{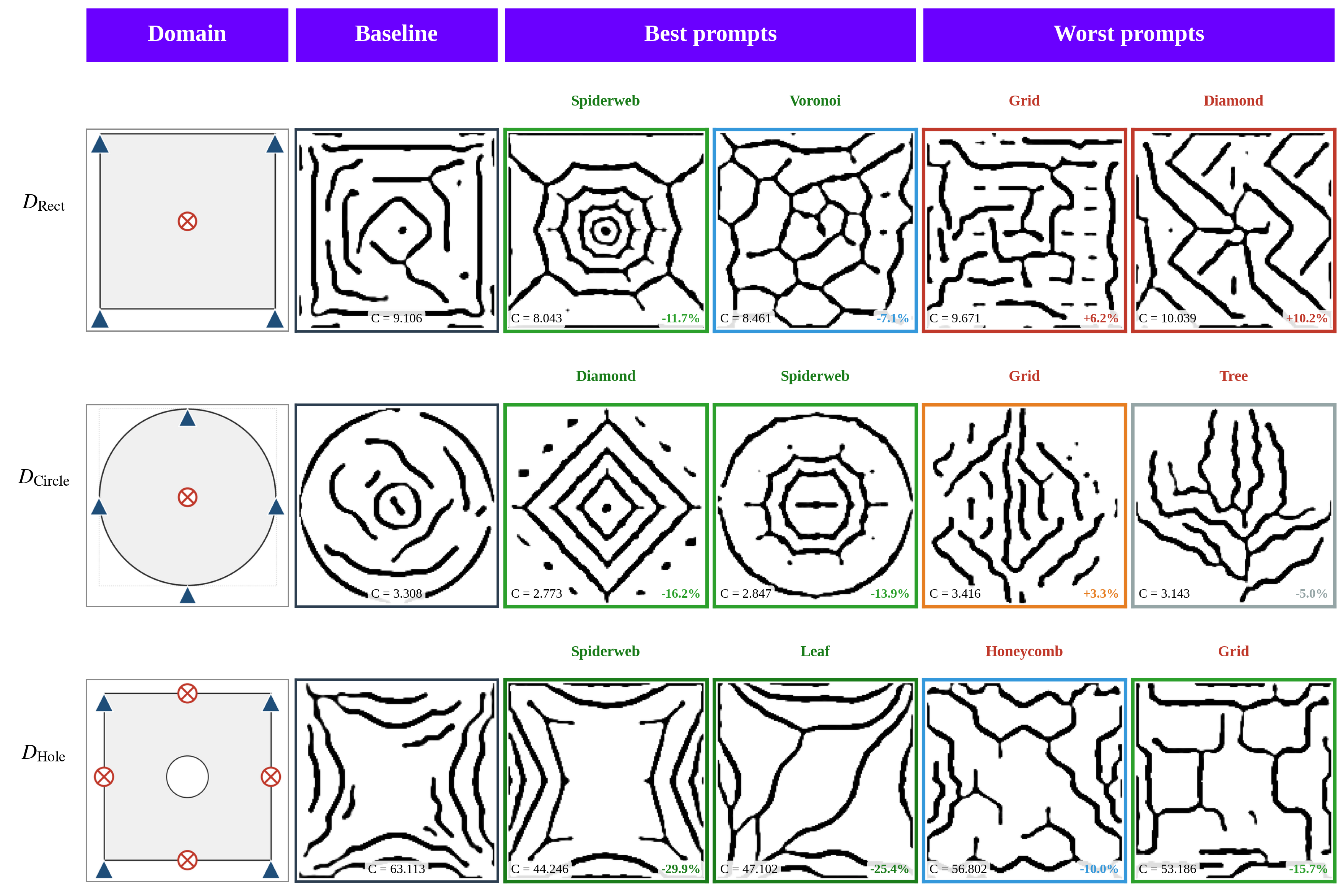}
\caption{Representative optimized density fields across the three plate-bending domains ($D_{\text{Rect}}$, $D_{\text{Circle}}$, $D_{\text{Hole}}$). Each row: boundary-condition schematic (blue triangles = pin supports; red $\otimes$ = transverse point load) $\mid$ baseline SIMP $\mid$ two best-performing and two worst-performing prompts (ranked by 5-seed mean compliance). Each SDS cell displays the seed whose final compliance is closest to the 5-seed mean. Cell-border color encodes $\Delta C/C_\text{base}$ (green tones = improvement, red tones = degradation).}
\label{fig:main_results}
\end{figure}

A total of 150 experiments (10 prompts $\times$ 3 domains $\times$ 5 random seeds) ran across three plate-bending domains with volume fraction $V^*=0.20$.
Statistical methodology and effect sizes are detailed in Section~\ref{sec:setup}. In brief, the binomial-null $p=0.031$ threshold for 5/5 wins defines significance, and median Cohen's $|d|=1.47$ confirms large practical effect.
Table~\ref{tab:summary} consolidates the results, and Fig.~\ref{fig:main_results} shows representative designs.

\textbf{$D_{\text{Rect}}$ (150$\times$150, 4-corner pinned, center point load)} offers the framework's smallest headroom.
SIMP already converges to an efficient solution, so only 5 of 10 prompts achieve significance, led by \textit{Spiderweb} ($-11.9\%$), \textit{Voronoi}, \textit{Bone}, \textit{Leaf}, and \textit{Tree}.
\textit{Grid} and \textit{Diamond} ($\approx\!+6\%$) actively degrade performance: their rigid lattice conflicts with the diagonal load paths and wastes material in low-stress regions.

\textbf{$D_{\text{Circle}}$ (circular domain, 4-directional pins, center load)} exposes the value of the generative prior.
The combination of non-rectangular geometry and multi-directional supports produces a symmetric landscape with many near-optimal topologies, and SIMP converges to just one.
SDS guides the density field to deeper optima that SIMP alone cannot reach, so 9 of 10 prompts achieve significance. \textit{Diamond} leads at $-15.2\%$ (best single run $-23.8\%$, $C=2.522$ vs.\ baseline $3.308$), with \textit{Spiderweb} and \textit{Voronoi} close behind.

\textbf{$D_{\text{Hole}}$ (central hole, 4-side center loads)} delivers the framework's largest gains.
The load-path detours around the hole and 4-sided symmetric loading trap the baseline SIMP at $C=63.113$, roughly seven times the $D_{\text{Rect}}$ baseline, so the raw weighted sum calls for a stronger SDS weight ($\lambda_\text{sds}^0=50$; Section~\ref{sec:sds}).
Under this setting, all ten prompts achieve 5/5 wins, with \textit{Spiderweb} leading at $-28.4\%$, followed by distributed-branching priors \textit{Leaf} ($-25.1\%$), \textit{Diamond} ($-24.6\%$), and \textit{Bone} ($-24.5\%$). The best single run reaches $-31.5\%$ (\textit{Spiderweb}, $C=43.20$).
Tiling priors whose geometry conflicts with the radial load flow lag behind (\textit{Honeycomb} $-9.7\%$, \textit{Grid} $-16.7\%$), foreshadowing the load-path alignment hypothesis developed in Section~\ref{sec:prompt_domain}.
Full mean $\pm$ std per-prompt values appear in Table~\ref{tab:summary}.

\begin{table}[pos=htbp]
\centering
\caption{Per-prompt compliance $C$ across the four design domains (mean $\pm$ std over $n=5$ seeds). Bold entries denote statistically significant improvements over the baseline (5/5 seeds win, $p=0.031$ binomial). The first two rows give the SIMP baseline and the SDS weight $\lambda_\text{sds}^0$ per domain; the last two rows report the single best run and the number of prompts (out of 10) achieving significance. $V^*=0.20$ for the three plate-bending domains and $V^*=0.30$ for the plane-stress $D_{\text{Link}}$. \rev{$95\%$ confidence intervals follow as mean $\pm\,2.78\,\text{std}/\sqrt{5}$; the per-seed distributions for all combinations are shown in Figs.~\ref{app:fig:seed_mech}--\ref{app:fig:seed_thermo}.}}
\label{tab:summary}
\setlength{\tabcolsep}{5pt}
\small
\begin{tabular}{@{}lcccc@{}}
\toprule
                      & $D_{\text{Rect}}$ & $D_{\text{Circle}}$ & $D_{\text{Hole}}$ & $D_{\text{Link}}$ \\
\midrule
Baseline $C_\text{base}$        & $9.106$ & $3.308$ & $63.113$ & $5.258$ \\
$\lambda_\text{sds}^0$          & $10$    & $10$    & $50$     & $10$    \\
\midrule
\textit{Bone}       & $\mathbf{8.48 \pm 0.07}$  & $\mathbf{3.001 \pm 0.14}$  & $\mathbf{47.68 \pm 2.42}$  & $\mathbf{4.73 \pm 0.18}$  \\
\textit{Diamond}    & $9.65 \pm 0.68$           & $\mathbf{2.804 \pm 0.22}$  & $\mathbf{47.59 \pm 2.31}$  & $\mathbf{4.80 \pm 0.20}$  \\
\textit{Grid}       & $9.64 \pm 0.06$           & $3.359 \pm 0.14$           & $\mathbf{52.58 \pm 2.76}$  & $\mathbf{4.78 \pm 0.21}$  \\
\textit{Honeycomb}  & $9.06 \pm 0.15$           & $\mathbf{3.131 \pm 0.12}$  & $\mathbf{56.97 \pm 1.01}$  & $\mathbf{4.79 \pm 0.25}$  \\
\textit{Leaf}       & $\mathbf{8.50 \pm 0.09}$  & $\mathbf{3.020 \pm 0.07}$  & $\mathbf{47.30 \pm 1.14}$  & $\mathbf{4.73 \pm 0.19}$  \\
\textit{Ornamental} & $9.17 \pm 0.35$           & $\mathbf{3.052 \pm 0.06}$  & $\mathbf{48.27 \pm 2.82}$  & $5.37 \pm 0.75$           \\
\textit{Spiderweb}  & $\mathbf{8.02 \pm 0.10}$  & $\mathbf{2.846 \pm 0.06}$  & $\mathbf{45.22 \pm 2.04}$  & $5.12 \pm 0.32$           \\
\textit{Tree}       & $\mathbf{8.51 \pm 0.19}$  & $\mathbf{3.143 \pm 0.04}$  & $\mathbf{47.78 \pm 0.51}$  & $\mathbf{4.81 \pm 0.10}$  \\
\textit{Truss}      & $8.97 \pm 0.18$           & $\mathbf{3.083 \pm 0.05}$  & $\mathbf{50.22 \pm 3.55}$  & $5.06 \pm 0.18$           \\
\textit{Voronoi}    & $\mathbf{8.43 \pm 0.12}$  & $\mathbf{2.906 \pm 0.05}$  & $\mathbf{48.39 \pm 2.67}$  & $\mathbf{4.70 \pm 0.10}$  \\
\midrule
Best single run $C$       & $7.886$ & $2.522$ & $43.20$ & $4.459$ \\
Sig.\ prompts (5/5 wins)  & $5/10$  & $9/10$  & $10/10$ & $7/10$  \\
\bottomrule
\end{tabular}
\end{table}

\subsection{Comparison with alternative baselines}
\label{sec:baselines}

\begin{figure}[pos=htbp]
\centering
\subfloat[$D_{\text{Rect}}$]{\raisebox{-.5\height}{\includegraphics[width=0.32\textwidth]{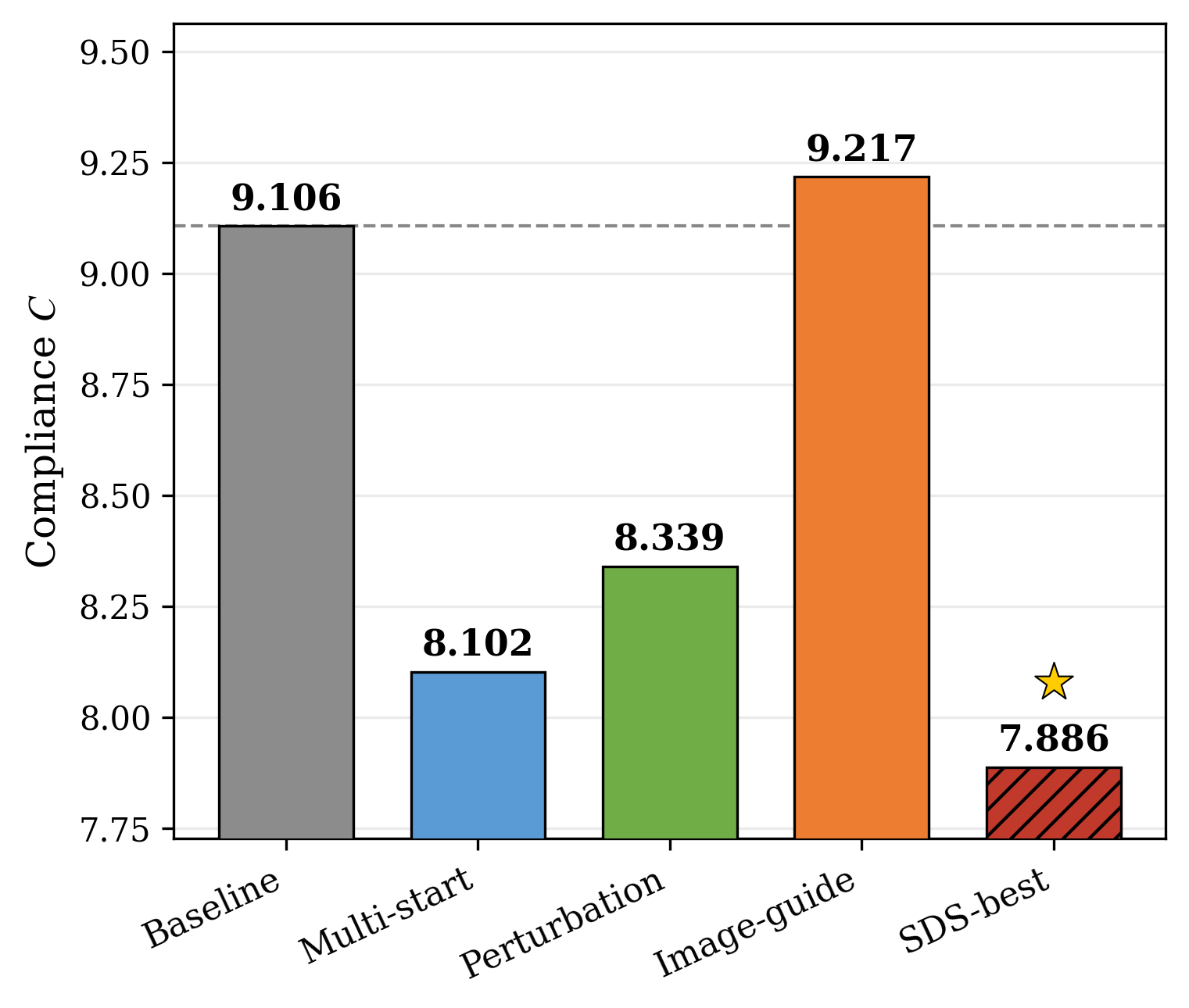}}}\hfill
\subfloat[$D_{\text{Circle}}$]{\raisebox{-.5\height}{\includegraphics[width=0.32\textwidth]{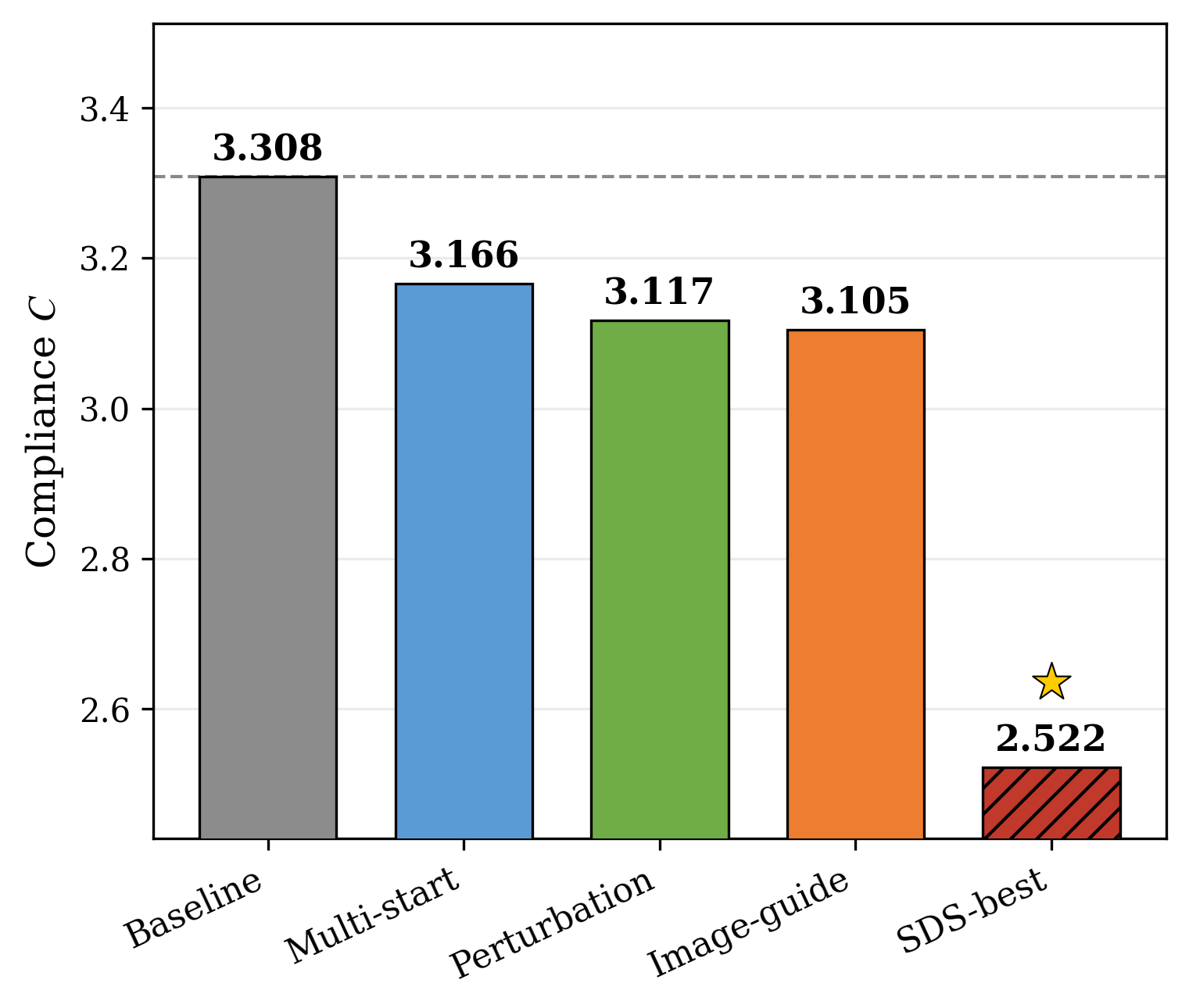}}}\hfill
\subfloat[$D_{\text{Hole}}$]{\raisebox{-.5\height}{\includegraphics[width=0.32\textwidth]{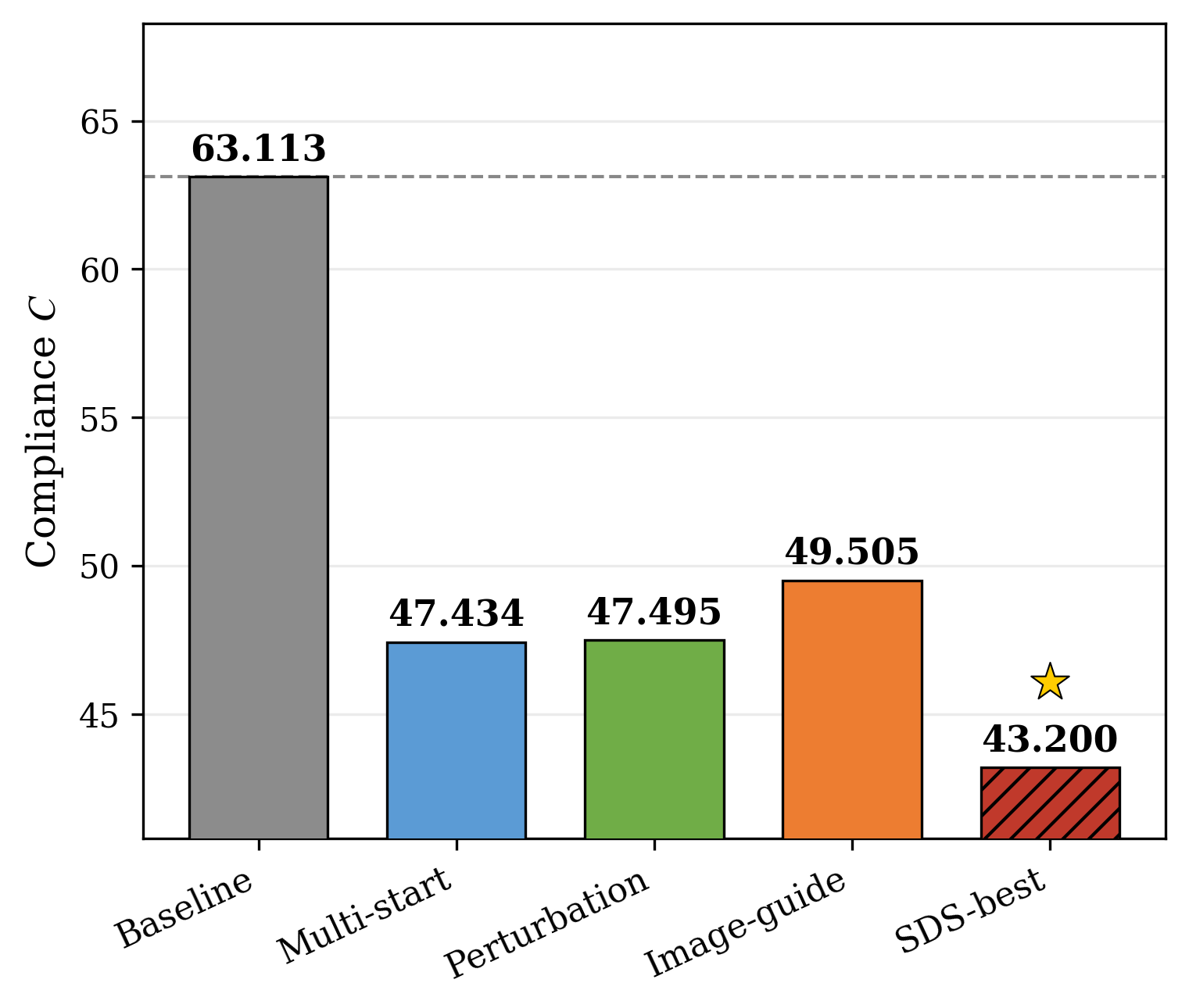}}}
\caption{Compliance comparison of SDS-guided optimization against three SIMP-only alternatives (Multi-start, Perturbation, Image-guided) across the three plate-bending domains\rev{; the multi-start and perturbation bars are the best of 50 runs per domain, with the matched-number-of-runs distributional analysis given in Appendix~\ref{app:baselines}}.}
\label{fig:baselines}
\end{figure}

Three alternative SIMP-only baselines test whether simpler methods can replicate the benefits of SDS (Fig.~\ref{fig:baselines}).
Multi-start SIMP launches optimization from \rev{50} random initial density fields (Gaussian noise, $\sigma=0.15$) and selects the best result.
Although multi-start surpasses the single-seed baseline in all domains, the best SDS result consistently outperforms the multi-start best.
Specifically, SDS achieves a best-run compliance of 7.886 versus \rev{8.102} for multi-start in $D_{\text{Rect}}$, 2.522 versus \rev{3.166} in $D_{\text{Circle}}$, and 43.20 versus \rev{47.434} in $D_{\text{Hole}}$ (the latter at $\lambda_\text{sds}^0=50$; Section~\ref{sec:main_results}; numerical summary in Appendix~\ref{app:baselines}).

Random perturbation SIMP periodically injects Gaussian noise into the density field. Such unstructured noise, however, cannot replace the text-conditioned structural guidance of SDS.
Consequently, performance remains comparable to or worse than the baseline on average.
Similarly, image-guided topology optimization tests whether the benefit originates from the final image or the dynamic optimization process by using the mean squared error loss with a fixed SDS-generated target.
While this method partially recovers some benefits in $D_{\text{Circle}}$ (3.105) and $D_{\text{Hole}}$ (49.505), it remains substantially inferior to the adaptive SDS approach, which achieves 2.522 and 43.20, respectively.
In $D_{\text{Rect}}$, image-guided optimization even degrades performance relative to the baseline (9.217 vs.\ 9.106).

\begin{figure}[pos=htbp]
\centering
\subfloat[$D_{\text{Rect}}$ ($C=7.721$)]{\raisebox{-.5\height}{\includegraphics[width=0.32\textwidth]{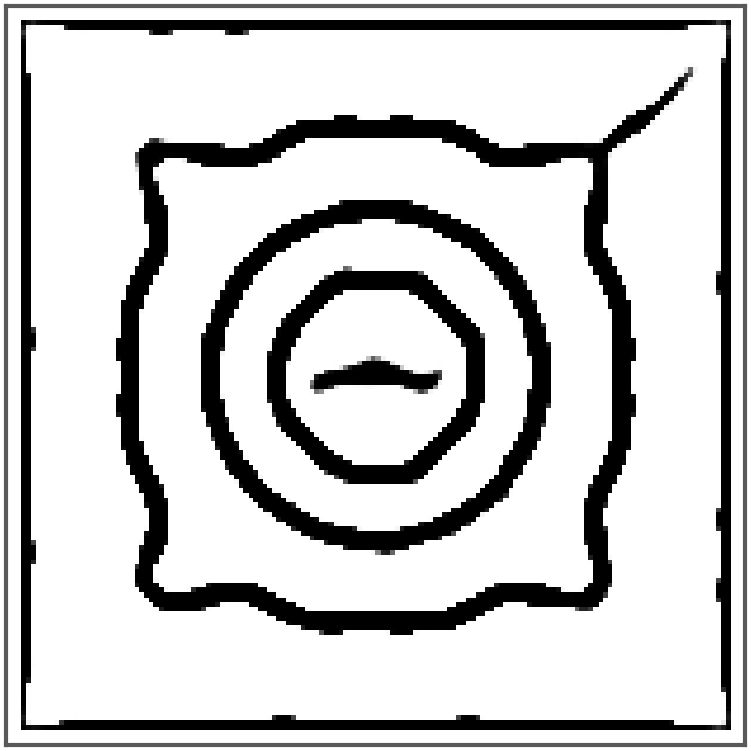}}}\hfill
\subfloat[$D_{\text{Circle}}$ ($C=2.598$)]{\raisebox{-.5\height}{\includegraphics[width=0.32\textwidth]{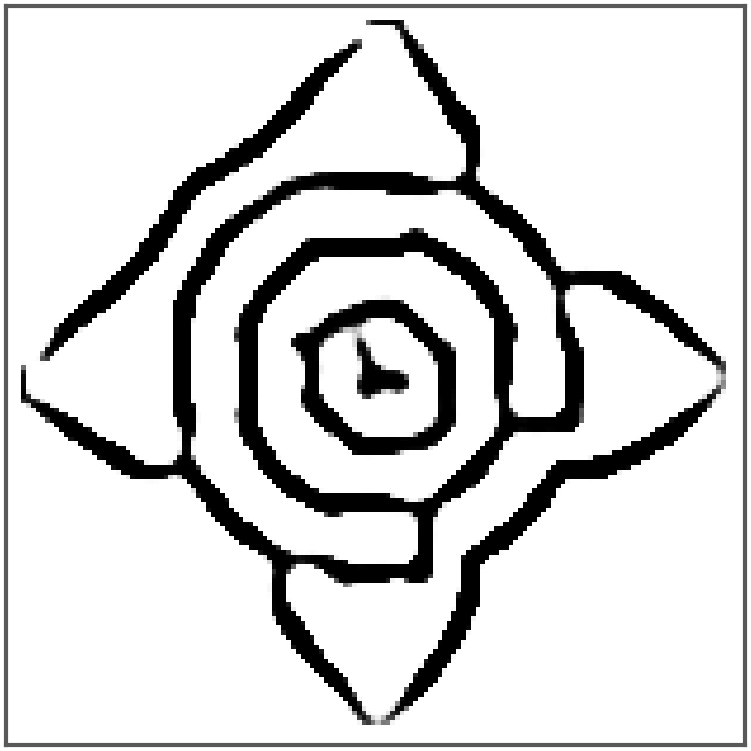}}}\hfill
\subfloat[$D_{\text{Hole}}$ ($C=48.848$)]{\raisebox{-.5\height}{\includegraphics[width=0.32\textwidth]{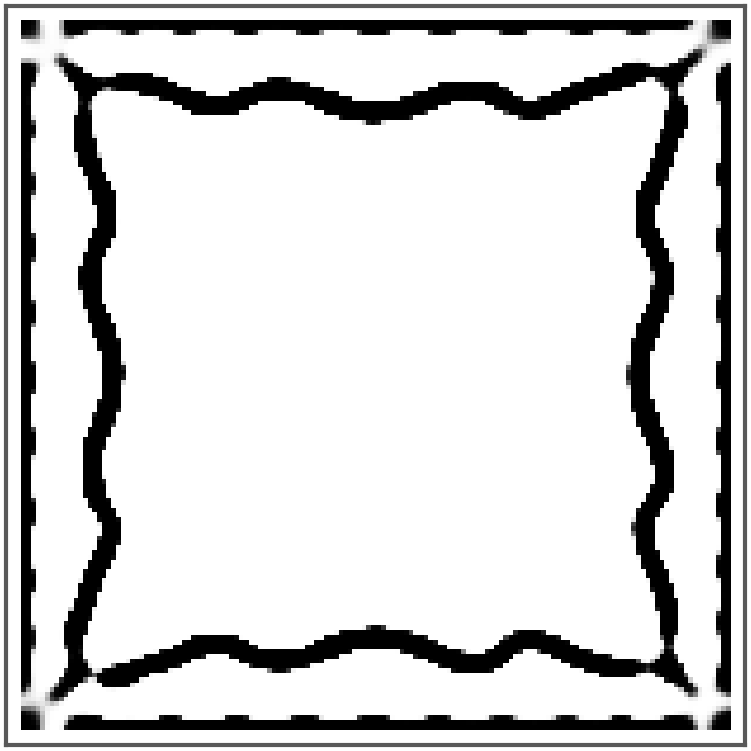}}}
\caption{\rev{Converged \textit{MMA} baseline designs for the three plate-bending domains (the \textit{MMA} column of Table~\ref{tab:mma_baseline}; $V^*=0.20$, 1000 iterations, Heaviside $\beta$-continuation, filter radii $r_{min}=5$ and $r_{max}=10$). The per-panel compliance $C$ is the best value attained over the run.}}
\label{fig:mma_baseline}
\end{figure}

The advantage of SDS therefore lies in its iterative, state-dependent gradient: even when given the exact SDS-generated design as a fixed target, image-guided optimization cannot recover the improvement, because it lacks the adaptive topology reshaping that SDS supplies during intermediate iterations.
Multi-start SIMP, for its part, samples multiple initial conditions but remains limited to the basins of attraction reachable from unstructured Gaussian-noise initializations.
In contrast, SDS accesses qualitatively different topological regions by imposing structured, physics-compatible exploration guided by the generative prior.
This gap is largest in $D_{\text{Hole}}$, where the complex load paths most severely trap conventional methods.
\rev{Beyond the conventional \textit{OC} optimizer, which performs substantially worse (Appendix~\ref{app:optimizer}), a tuned Method of Moving Asymptotes (\textit{MMA}) optimizer \cite{svanberg1987method}, a standard strong baseline in gradient-based topology optimization, is also benchmarked. Run for the same 1000 iterations under identical filtering and $\beta$-continuation, \textit{MMA} improves substantially on the \textit{Adam} baseline in every domain (Table~\ref{tab:mma_baseline}; converged designs in Fig.~\ref{fig:mma_baseline}) and thus serves as a strong single-start gradient baseline. Even against this baseline, the best SDS run attains lower compliance in the two multi-directional domains, $D_{\text{Circle}}$ ($2.522$ vs.\ $2.598$) and $D_{\text{Hole}}$ ($43.20$ vs.\ $48.85$). Because \textit{MMA} is a single deterministic run, this best-run comparison is the matched one; on the five-seed mean the stochastic SDS prompts straddle \textit{MMA}, as expected for a method with run-to-run variation. It trails only in the single-load radial domain $D_{\text{Rect}}$ ($7.886$ vs.\ $7.721$), where the radial load path makes the compliance landscape simple enough that gradient descent already reaches a near-global optimum. The comparison is conservative in favour of \textit{MMA}. It is $\approx$2.1$\times$ more expensive per iteration than \textit{Adam} ($4.90$ vs.\ $2.34$~s/iter), so an equal-iteration comparison already grants it roughly twice the compute budget of the gradient baselines. \textit{MMA} is also a deterministic single-start method, whereas the matched-number-of-runs multi-start study (Appendix~\ref{app:baselines}) shows that even 50 random restarts do not reach the SDS result. That SDS still prevails wherever the load paths are non-trivial indicates that its contribution is not to win a compliance race against a tuned optimizer. Steered by a text prompt and without task-specific training, it reaches topological basins that gradient descent does not access, precisely in the domains where the design problem is hard.}

\begin{table}[pos=htbp]
\centering
\caption{\rev{Baseline compliance: tuned \textit{MMA} vs.\ \textit{Adam} and the best SDS prompt (no SDS for the optimizer baselines; $V^*=0.20$, 1000 iterations). \textit{MMA} improves on \textit{Adam} everywhere but is surpassed by SDS in the two multi-directional domains.}}
\label{tab:mma_baseline}
\rev{\begin{tabular}{lcccc}
\toprule
Domain & \textit{Adam} $C$ & \textit{MMA} $C$ & SDS best $C$ & Lowest \\
\midrule
$D_{\text{Rect}}$   & 9.106  & \textbf{7.721}  & 7.886  & \textit{MMA} \\
$D_{\text{Circle}}$ & 3.308  & 2.598  & \textbf{2.522}  & SDS \\
$D_{\text{Hole}}$   & 63.113 & 48.848 & \textbf{43.20}  & SDS \\
\bottomrule
\end{tabular}}
\end{table}

The cost contrast with data-driven generative alternatives is also large.
Data-driven alternatives such as TopoDiff \cite{maze2023diffusion} and DOM \cite{giannone2023aligning} require thousands of GPU-hours and large pre-computed solution datasets (Table~\ref{tab:comparison}).
The present framework eliminates both: retargeting to a new boundary condition, physics objective, or domain geometry reduces to a single optimization run (Section~\ref{sec:setup}) with zero label cost.
The saving recurs for each new condition explored, since retargeting needs no retraining.

\subsection{Generalization to industrial geometry}
\label{sec:link}

\begin{figure}[pos=htbp]
\centering
\includegraphics[width=\textwidth]{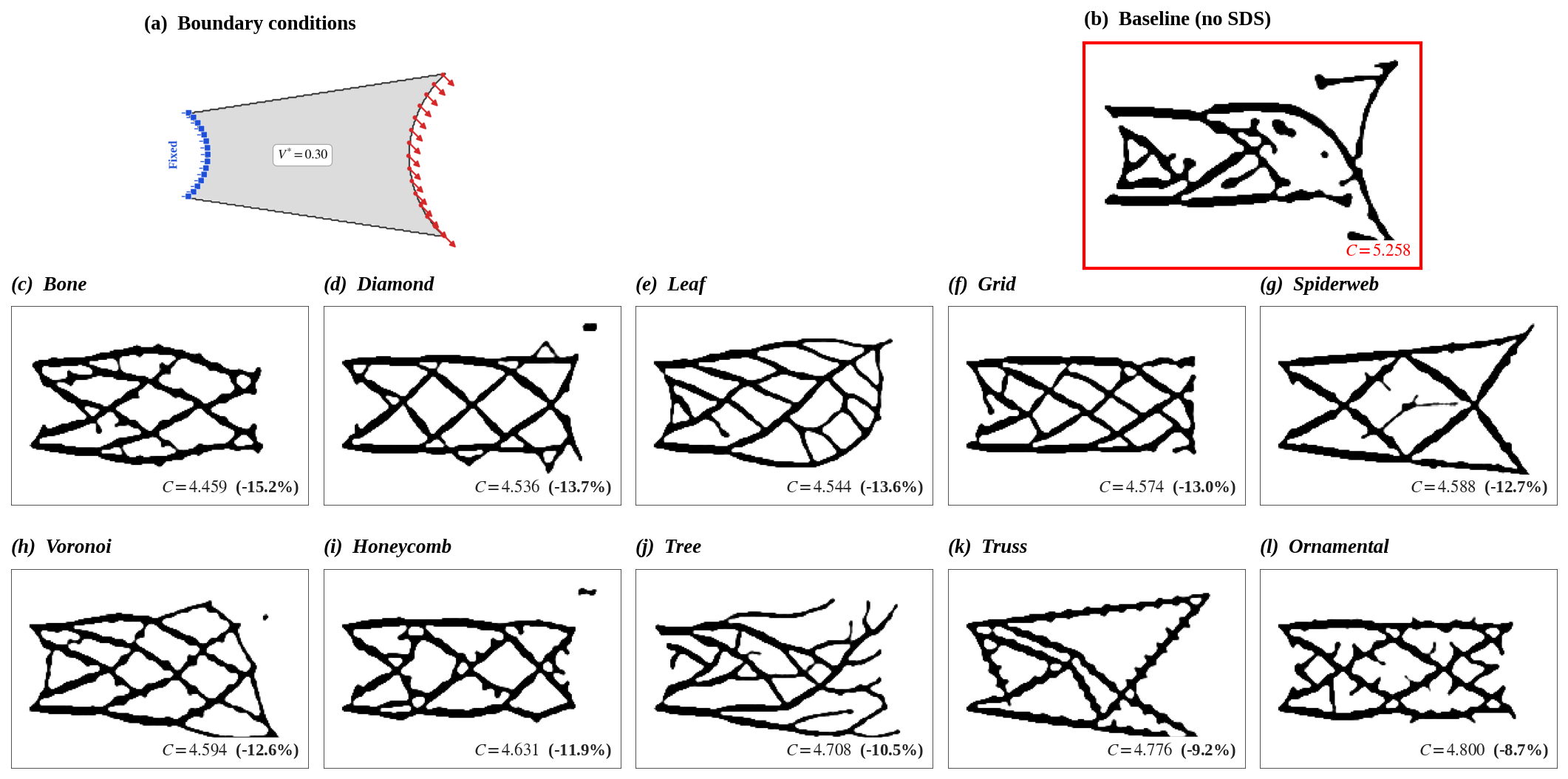}
\caption{Optimized density fields for the $D_{\text{Link}}$ domain (automotive suspension link, plane stress, $V^*=0.30$). \rev{(a) Boundary conditions: the left arc is fixed and the right arc carries combined tension--shear loading; (b) baseline SIMP (no SDS); (c--l) designs under ten text prompts, ordered by ascending compliance.} The irregular domain geometry and distributed boundary conditions produce complex, non-radial load paths that are qualitatively distinct from the synthetic domains.}
\label{fig:link}
\end{figure}

To test generalization beyond synthetic benchmarks, 50 additional experiments (10 prompts $\times$ 5 seeds) ran on $D_{\text{Link}}$, which uses the 2D cross-section of an automotive suspension link as the design domain (Fig.~\ref{fig:link}).
Unlike the three plate-bending domains analyzed under Mindlin--Reissner plate theory, $D_{\text{Link}}$ employs plane stress analysis (4-node bilinear element, 2 DOF per node) with $V^*=0.30$, and features an asymmetric geometry with distributed supports on the left arc and combined tension--shear loading ($f_x = f_y = 0.01$) on the right arc.
Seven out of ten prompts achieved statistically significant improvements (5/5 wins, $p=0.031$. See $D_{\text{Link}}$ column of Table~\ref{tab:summary}).
The top performer \textit{Voronoi} reduced mean compliance by $-10.6\%$, closely followed by \textit{Leaf} ($-10.0\%$) and \textit{Bone} ($-10.0\%$).
These leading prompts share a common characteristic: they induce distributed, irregularly branching networks that can adapt freely to asymmetric left-to-right load transmission paths.
Conversely, the only prompt that degraded performance (\textit{Ornamental}, $+2.1\%$) exhibited the highest seed-level variability ($\text{std}=0.75$), consistent with the observation that misaligned prompts produce unreliable results.
Full mean $\pm$ std values per prompt appear in Table~\ref{tab:summary}.

\subsection{Prompt--domain interaction governs performance}
\label{sec:prompt_domain}

\begin{figure}[pos=htbp]
\centering
\includegraphics[width=\textwidth]{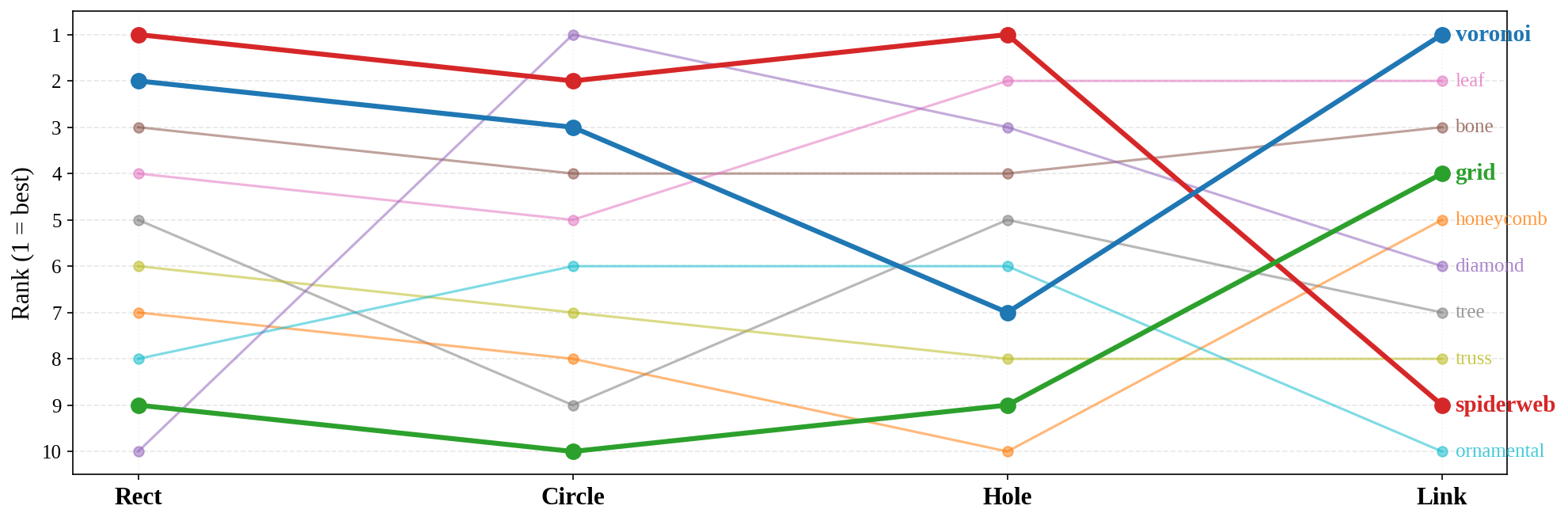}
\caption{Parallel-coordinates plot of prompt rankings across the four design domains (Rank 1 = lowest mean compliance). Three prompts with the most informative trajectories (\textit{Spiderweb}, \textit{Voronoi}, \textit{Grid}) are highlighted; the remaining seven appear in muted colors.}
\label{fig:ranking}
\end{figure}

Prompt rankings across the four domains reveal pronounced reversals that define a \textit{load-path alignment hypothesis}: compliance improves when the SDS-induced pattern geometrically aligns with the domain's actual load paths (Fig.~\ref{fig:ranking}).

\textit{Spiderweb}, which induces a radial pattern, leads in $D_{\text{Rect}}$ ($-11.9\%$) and $D_{\text{Hole}}$ ($-28.4\%$) and ranks second in $D_{\text{Circle}}$ ($-14.0\%$), where center-to-support load paths naturally align with radial geometry.
The same prompt drops to 9th in the industrial $D_{\text{Link}}$ domain ($-2.6\%$), because its single-focal-point bias does not match the distributed arc loading.
\textit{Voronoi} shows the opposite pattern: its irregular cell partitioning imposes minimal geometric constraint, letting the physics solver adapt to distributed loads, which lifts it from 2nd in $D_{\text{Rect}}$ and 3rd in $D_{\text{Circle}}$ to 1st in $D_{\text{Link}}$ ($-10.6\%$).
Biomimetic branching priors (\textit{Leaf}, \textit{Bone}) remain broadly effective, placing 3rd--4th in $D_{\text{Rect}}$ and among the top four in $D_{\text{Hole}}$ and $D_{\text{Link}}$.

\textit{Grid} underperforms in every radially loaded domain: $+5.9\%$ degradation in $D_{\text{Rect}}$, 10th in $D_{\text{Circle}}$ ($+1.5\%$), and 9th in $D_{\text{Hole}}$ ($-16.7\%$).
Its orthogonal geometry conflicts with radial load flow.
The exception is $D_{\text{Link}}$, where \textit{Grid} rises to 4th ($-9.1\%$) because the orthogonal lattice aligns productively with the tension--shear decomposition of the distributed arc loading.
The same alignment principle holds even in the most deeply trapped domain $D_{\text{Hole}}$, where radial/branching priors cluster near $-25\%$ while conflicting tiling priors (\textit{Honeycomb} $-9.7\%$) trail the group by more than 15 percentage points.

These rank reversals are a direct consequence of making prompt--domain compatibility explicit through physics feedback at every iteration, rather than learning it implicitly from labeled data as a trained generative model would. The design implications are developed in Section~\ref{sec:discussion}.

\subsection{Generalization to thermoelastic loading}
\label{sec:thermoelastic}

\begin{figure}[pos=htbp]
\centering
\includegraphics[width=\textwidth]{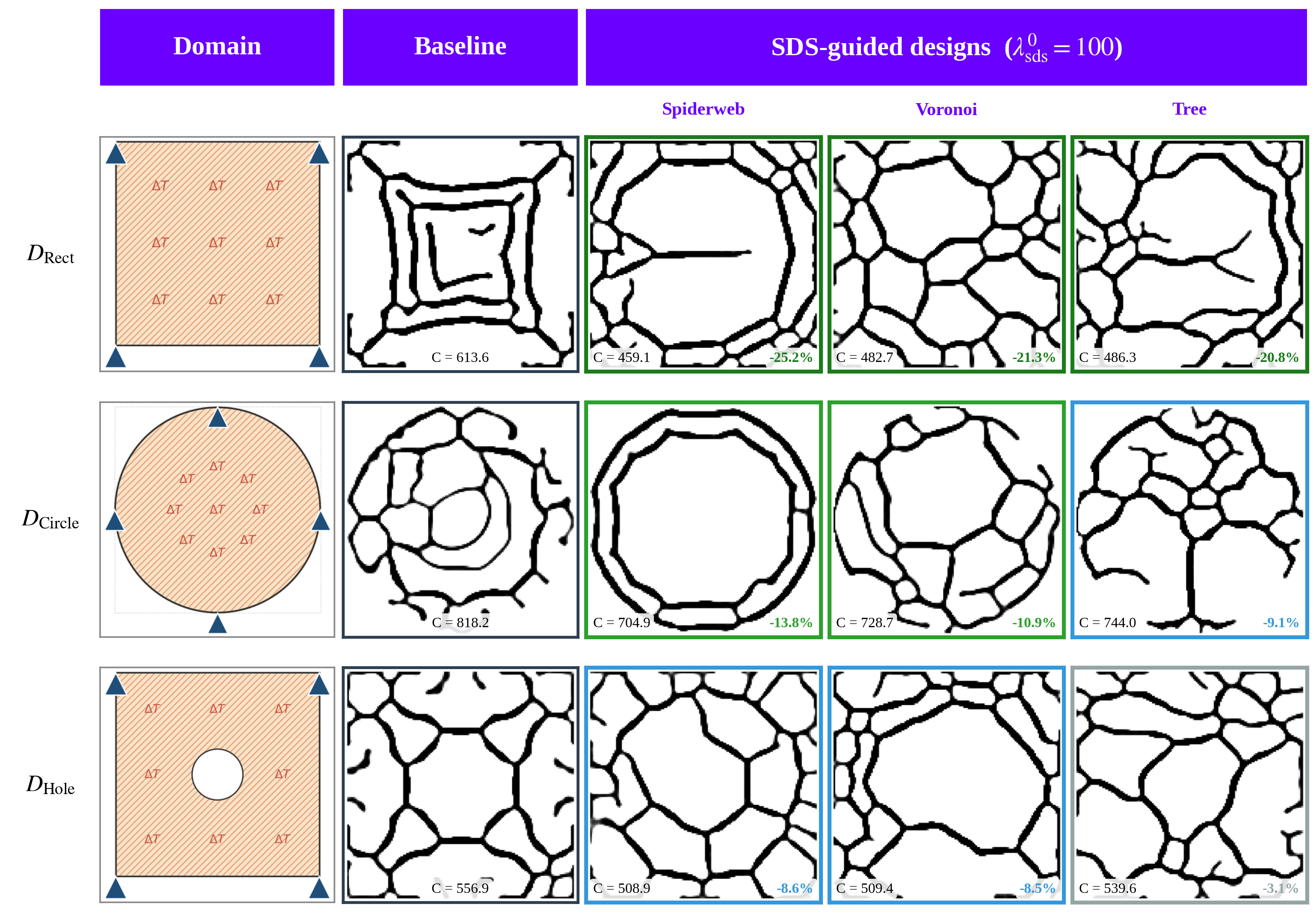}
\caption{Thermoelastic topology optimization under a uniform through-thickness temperature gradient $\Delta T$. Each row corresponds to one domain ($D_{\text{Rect}}$, $D_{\text{Circle}}$, $D_{\text{Hole}}$); columns show the boundary-condition schematic, the baseline SIMP solution, and SDS-guided designs at $\lambda_\text{sds}^0 = 100$ for \textit{Spiderweb}, \textit{Voronoi}, and \textit{Tree}. Cell-border color encodes the improvement using the same scale as Fig.~\ref{fig:main_results}; per-cell compliance $C$ and $\Delta\%$ are overlaid.}
\label{fig:thermoelastic}
\end{figure}

A further test applies the framework to thermoelastic compliance minimization, where a uniform through-thickness temperature gradient $\Delta T$ induces distributed thermal bending moments rather than concentrated external forces \cite{rodrigues1995thermoelastic}.
The objective becomes $C = \mathbf{F}_\text{th}^\top \mathbf{u}$, with the thermal load prescribed (design-independent) so the sensitivity retains its standard self-adjoint form.
The MITC4 plate FEA, SIMP interpolation, Heaviside projection, and SDS scheduling carry over unchanged. Only the load vector changes.

The SDS weight scales to $\lambda_\text{sds}^0=100$ to match the larger and more diffuse thermal physics gradient (Section~\ref{sec:sds}; empirical sweep in Appendix~\ref{app:thermo_lambda}).
Three prompts (\textit{Spiderweb}, \textit{Voronoi}, \textit{Tree}) ran with five seeds each across $D_{\text{Rect}}$, $D_{\text{Circle}}$, and $D_{\text{Hole}}$. Fig.~\ref{fig:thermoelastic} shows representative designs and Table~\ref{tab:thermoelastic} the full results.
In $D_{\text{Rect}}$ (baseline $C=613.6$), all three prompts reached 5/5 wins with mean improvements up to $-23.0\%$. In $D_{\text{Circle}}$ (baseline $C=818.2$), all three reached 5/5 wins ($-9.6\%$ to $-13.9\%$). In $D_{\text{Hole}}$ (baseline $C=556.9$), \textit{Voronoi} achieved 5/5 significance ($-8.5\%$) while \textit{Spiderweb} and \textit{Tree} reached 4/5 with $-7.2\%$ and $-2.6\%$ mean improvement, and no prompt degraded the baseline.
Across all thermoelastic settings, 7 of 9 combinations reached 5/5 wins and the remaining two reached 4/5.

\begin{table}[pos=htbp]
\centering
\caption{Thermoelastic compliance results with $\lambda_\text{sds}^0=100$ (mean $\pm$ std over 5 seeds, $V^*=0.20$). Values are thermoelastic compliance $C$; lower is better. Bold with ``5/5'' indicates statistically significant improvement (5/5 wins vs.\ baseline); ``4/5'' denotes four out of five seeds improving over the baseline.}
\label{tab:thermoelastic}
\setlength{\tabcolsep}{4pt}
\footnotesize
\begin{tabular}{lccc}
\toprule
& $D_{\text{Rect}}$ & $D_{\text{Circle}}$ & $D_{\text{Hole}}$ \\
\midrule
Baseline (\textit{Adam}) & 613.6 & 818.2 & 556.9 \\
\textit{Spiderweb} & $\mathbf{472.5 \pm 37}$ ($-23.0\%$, 5/5) & $\mathbf{704.6 \pm 4}$ ($-13.9\%$, 5/5) & $516.5 \pm 36$ ($-7.2\%$, 4/5) \\
\textit{Voronoi} & $\mathbf{480.5 \pm 6}$ ($-21.7\%$, 5/5) & $\mathbf{727.7 \pm 19}$ ($-11.1\%$, 5/5) & $\mathbf{509.6 \pm 10}$ ($-8.5\%$, 5/5) \\
\textit{Tree} & $\mathbf{486.9 \pm 2}$ ($-20.7\%$, 5/5) & $\mathbf{739.9 \pm 15}$ ($-9.6\%$, 5/5) & $542.4 \pm 12$ ($-2.6\%$, 4/5) \\
\midrule
Best $\Delta\%$ & $-23.0\%$ & $-13.9\%$ & $-8.5\%$ \\
\bottomrule
\end{tabular}
\end{table}

The domain difficulty ranking reverses relative to mechanical loading: $D_{\text{Rect}}$, the easiest mechanical domain, becomes the most responsive under uniform thermal loading because its isotropic thermal moment admits many competing near-optimal topologies that the generative prior helps discriminate among.
The central hole in $D_{\text{Hole}}$, conversely, geometrically constrains feasible topologies and drives the baseline close to optimality ($C=556.9$ vs.\ $613.6$ for $D_{\text{Rect}}$), leaving little headroom.
This reversal is consistent with the load-path alignment hypothesis: radial priors still succeed at $D_{\text{Circle}}$ and branching priors at $D_{\text{Rect}}$, but prompts whose spatial bias is less directly mirrored by the loading benefit from stronger generative weight to compete against the larger physics gradient.

\section{Mechanisms of improvement}
\label{sec:mechanisms}

Across the six experiments (mechanical point loads, distributed arc loading, and thermoelastic bending), the same pattern of improvement recurs.
This section examines the structural mechanism behind SDS-guided gains, the conditions under which it operates, and the roles of the scheduling and gray-density components.

\subsection{Structural signature: dead-end suppression}
\label{sec:dead_end}

\begin{figure}[pos=htbp]
\centering
\rev{\includegraphics[width=0.82\textwidth]{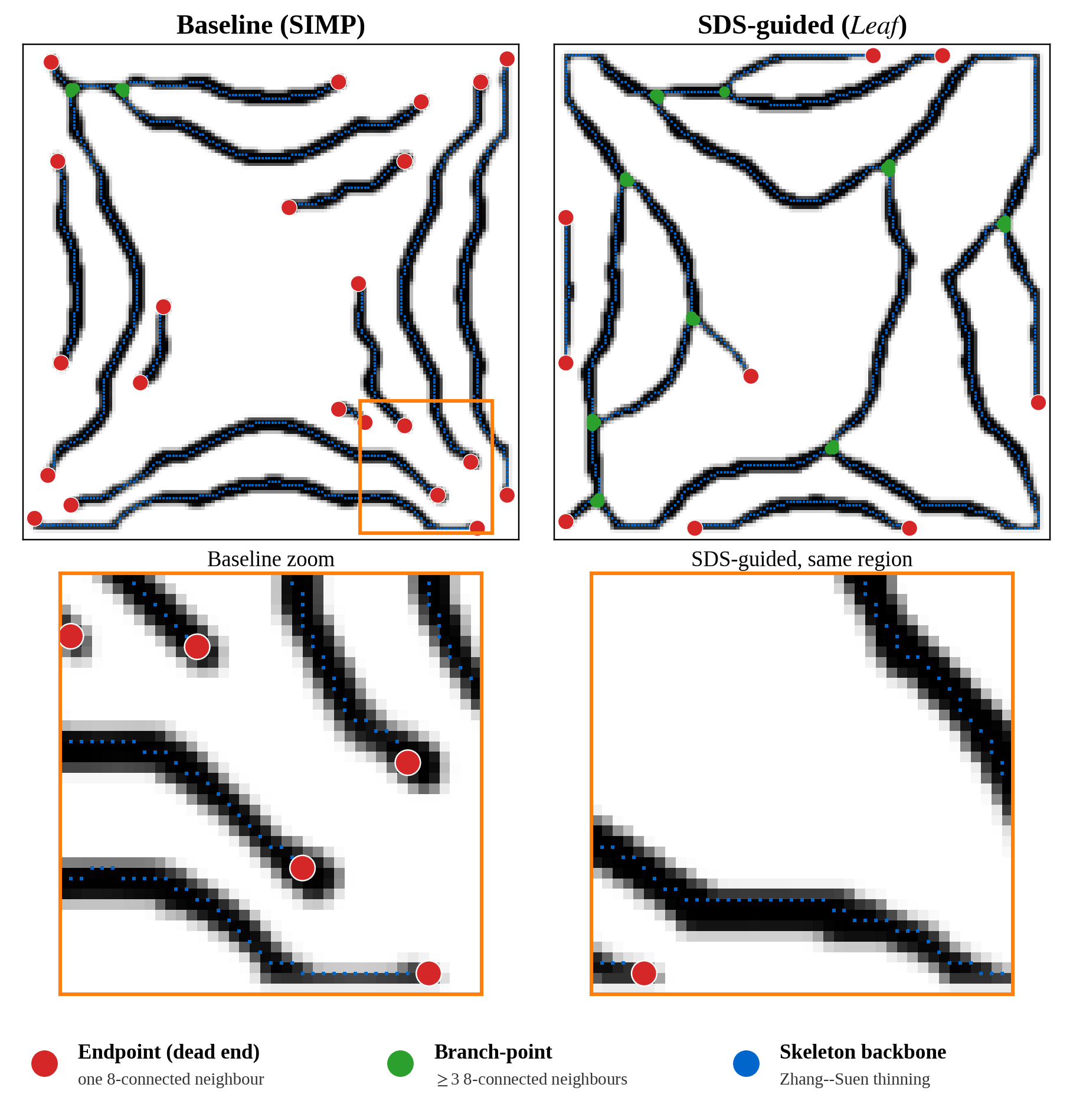}}
\caption{\rev{Qualitative illustration of dead-end suppression on $D_{\text{Hole}}$. Skeletons (Zhang--Suen thinning, blue) of the baseline SIMP design (left) and an SDS-guided design (\textit{Leaf}, right), with endpoints/dead ends (red, one 8-connected neighbour) and branch points (green, three or more). The magnified insets show that the baseline retains numerous free-ending ribs that never reach a support, whereas the SDS-guided network reroutes or removes them, simplifying rather than enriching the topology.}}
\label{fig:dead_end_example}
\end{figure}

In most domains tested, SDS-guided designs tend to eliminate non-load-bearing dead-end branches from the rib skeleton (free ends that extend from the skeleton but fail to reach a support).
\rev{This runs counter to a natural expectation. Because generative image models are tuned for visual diversity, one might anticipate that a diffusion prior would enrich the skeleton with additional branches. Coupled to the FEA sensitivity, however, SDS does the opposite (Fig.~\ref{fig:dead_end_example}). In the baseline design (left), many skeletal branches terminate without reaching a support, and the red endpoints in the magnified inset mark ribs that carry essentially no load, whereas the SDS-guided design (right) reroutes or removes these stubs, leaving a sparser network of fully connected load paths. The generative prior therefore simplifies the structure rather than diversifying it, and this pruning of dead ends is what improves physical performance.}

\begin{figure}[pos=htbp]
\centering
\subfloat[Baseline vs.\ best-SDS prompt skeleton endpoint counts across the six experimental domains. Per-bar labels give the percentage change from baseline.]{%
\includegraphics[width=0.48\textwidth]{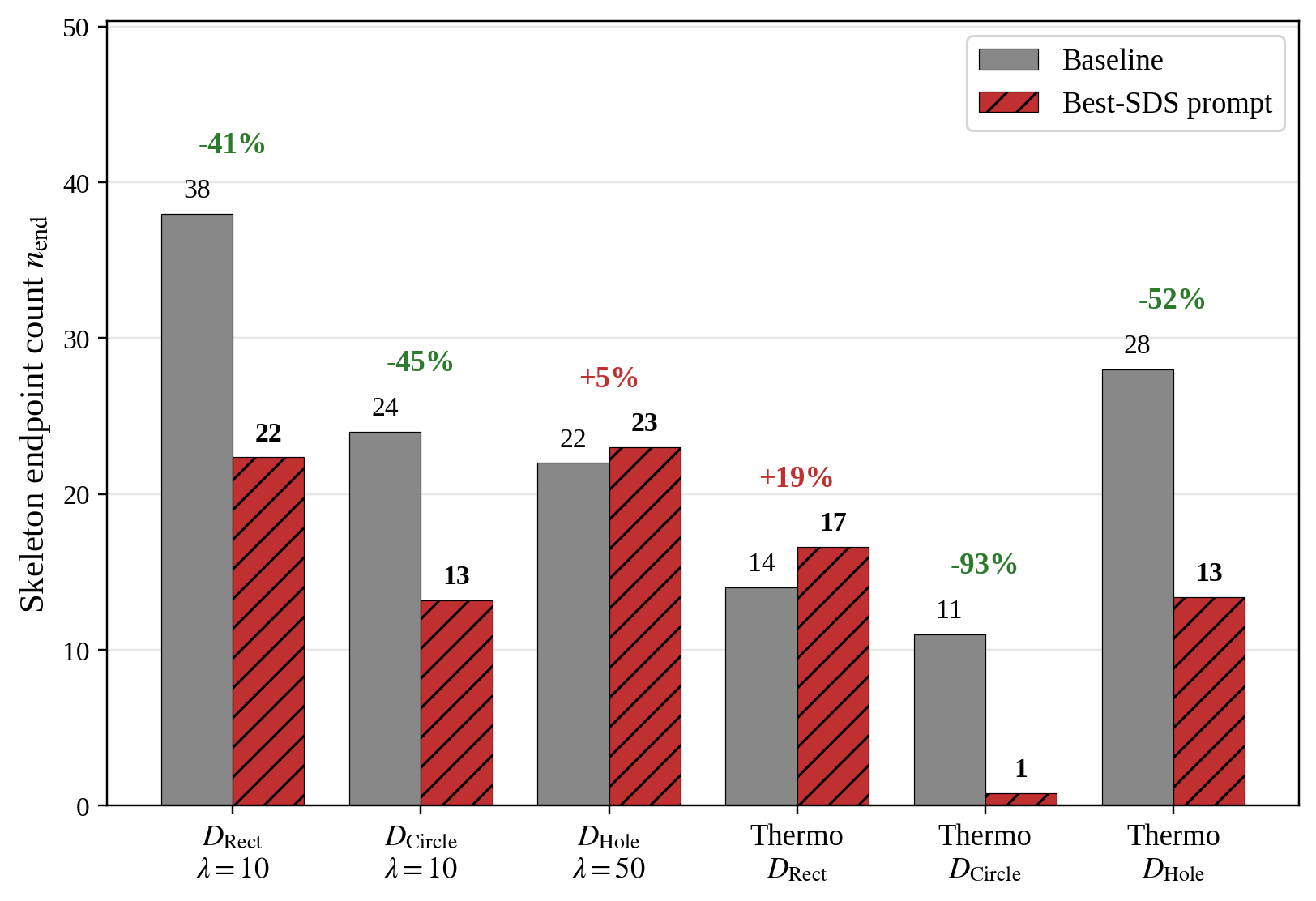}}\hfill
\subfloat[Prompt-level scatter of mean compliance improvement against mean skeleton endpoint count $\bar{n}_\text{end}$ across all six domains. Per-domain Pearson $r$ is annotated.]{%
\includegraphics[width=0.48\textwidth]{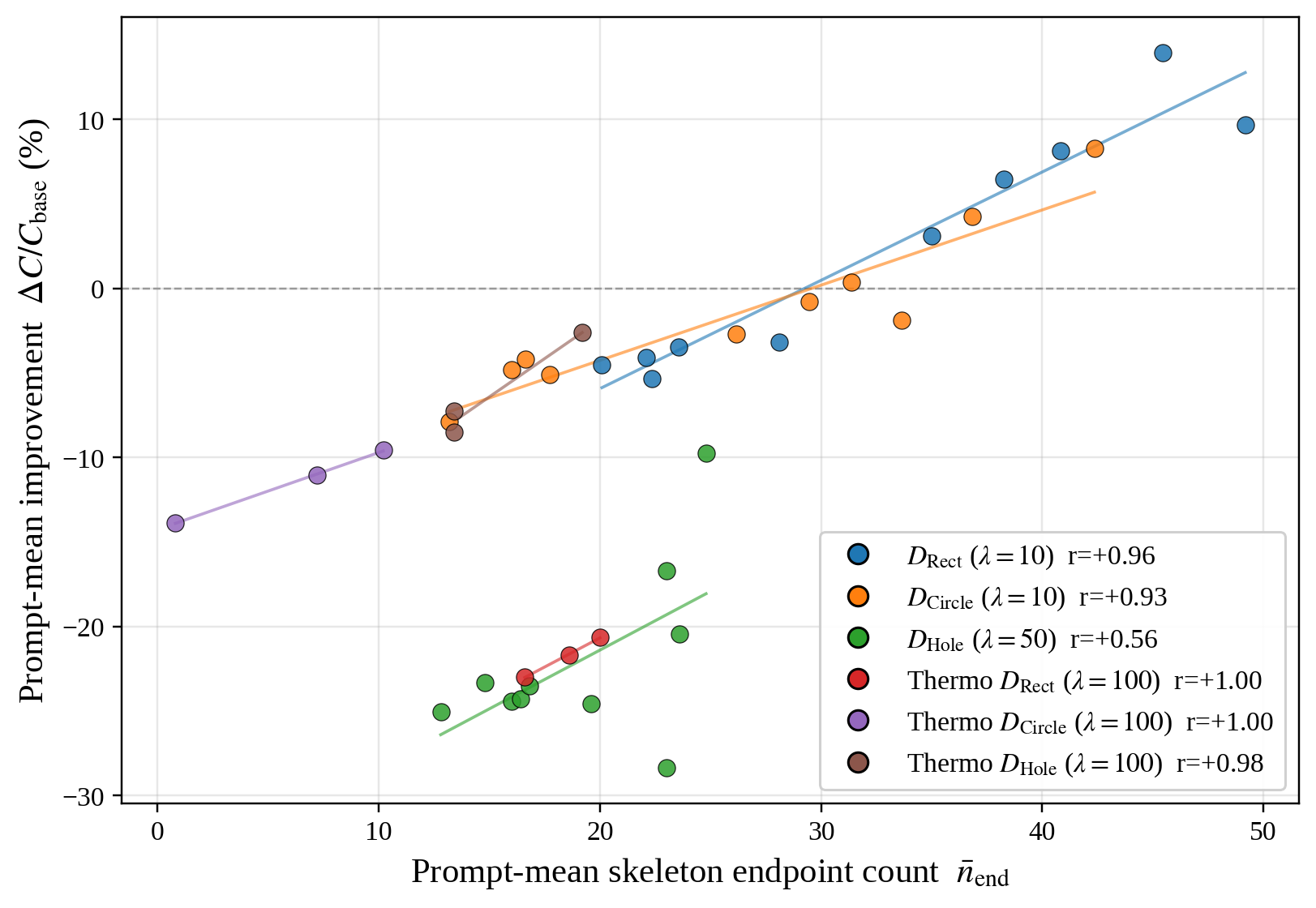}}
\caption{Cross-domain evidence that dead-end suppression is the structural signature of SDS-guided improvement. Each panel shows one representative design per domain, so its endpoint counts differ slightly from the five-seed means in Table~\ref{tab:morphology}.}
\label{fig:dead_end}
\end{figure}

Stable Diffusion, trained on billions of image-text pairs from LAION-5B \cite{schuhmann2022laion}, produces continuous linear networks in response to structural prompts prefixed with ``bold black lines on white background,'' and this bias propagates through the VAE encoder back to the density field.
\rev{A natural-image prior helps here for a concrete reason: the thin, branching line networks the model produces for such prompts are the morphologies (leaf venation, trabecular bone, spider webs) that efficient rib layouts approximate (Section~\ref{sec:introduction}). The visual bias of the prior and the structural objective are thus morphologically aligned, and the ``bold black lines on white background'' prefix explicitly steers the model toward this line-network sub-manifold of its training distribution.}
Morphological analysis across all six experimental domains (Fig.~\ref{fig:dead_end} and Table~\ref{tab:morphology}) shows that the prompt-level Pearson correlation between mean compliance and skeleton endpoint count $n_\text{end}$ is positive in every domain ($r = +0.56$ to $+0.99$, median $r=+0.97$).
In several domains, top-performing prompts reduce endpoint counts by $30$--$50\%$ while preserving or extending total skeleton length: \textit{Spiderweb} drops endpoints in $D_{\text{Rect}}$ from 38 to 19 and elongates the network from 1090 to 1299 skeleton pixels. \textit{Leaf} in $D_{\text{Hole}}$ reduces endpoints from 22 to 13. \textit{Spiderweb} in thermoelastic $D_{\text{Circle}}$ collapses endpoints from 11 to below 1.
By contrast, neither branch-point count $n_\text{branch}$ nor raw connectivity correlates consistently with compliance across domains.
\rev{The dot-prompt ablation provides controlled evidence consistent with this mechanism (Fig.~\ref{fig:prompt_ablation}; setup and full table in Appendix~\ref{app:ablation}). On $D_{\text{Rect}}$, changing a single prompt word (``lines''$\to$``dots'') so that the prior places isolated, non-load-bearing points inflates the endpoint count beyond any other prompt and yields the largest degradation below the baseline ($+7.3\%$, the only condition with no seed improving on the baseline). The semantically irrelevant \textit{sunset} prompt, with the next-highest endpoint count, is the only other condition to worsen the baseline ($+3.6\%$), so the two most endpoint-rich prompts are exactly the two that degrade compliance. Across the thirty individual runs (five seeds for each of the six prompts) the endpoint count and the compliance change track each other at $r=0.87$. Semantically empty or prefix-only prompts yield far smaller gains ($-3.6\%$ and $-4.5\%$) than the structural prompt ($-11.9\%$). The improvement tracks the endpoint count, which structural prompts reliably minimize, rather than the mere presence of generative coupling. Even the semantically irrelevant \textit{fruit} prompt helps ($-5.5\%$) when it happens to yield a low-endpoint morphology. A complementary within-design test isolates the endpoint effect without altering the surrounding morphology: on fixed optimized designs, surgically removing all dead-end twigs (skeleton-endpoint pruning followed by re-analysis) raises compliance by only $\sim$6\% on average across nine designs, whereas removing a single load-bearing rib raises it by $+61.5\%$ on average (up to $+182\%$), and grafting an artificial dead-end changes compliance by under $1\%$ (Table~\ref{app:tab:surgery}). Dead ends therefore carry comparatively little load, providing direct intervention-based support for the mechanism, more directly than the dot ablation, which co-varies the global morphology. The interventions and per-design values are detailed in Appendix~\ref{app:surgery}.}

\begin{table}[pos=htbp]
\centering
\caption{\rev{Within-design causal surgery: mean compliance change ($\Delta C$) over three designs per domain for each intervention. Dead-end removal costs far less than removing a single load-bearing rib, and grafting a dead-end is inert.}}
\label{app:tab:surgery}
\rev{\begin{tabular}{@{}lccc@{}}
\toprule
Domain & Prune dead-ends & Prune load-bearing rib & Graft dead-end \\
\midrule
$D_{\text{Rect}}$   & $+6.9\%$ & $+18.1\%$  & $-0.8\%$ \\
$D_{\text{Circle}}$ & $+7.2\%$ & $+24.3\%$  & $-0.4\%$ \\
$D_{\text{Hole}}$   & $+3.5\%$ & $+142.0\%$ & $-0.1\%$ \\
\midrule
All (9 designs)   & $+5.9\%$ & $+61.5\%$  & $\approx 0$ \\
\bottomrule
\end{tabular}}
\end{table}

\begin{figure}[pos=htbp]
\centering
\rev{\subfloat[Optimized rib layouts under the six prompt conditions]{\includegraphics[width=0.62\linewidth]{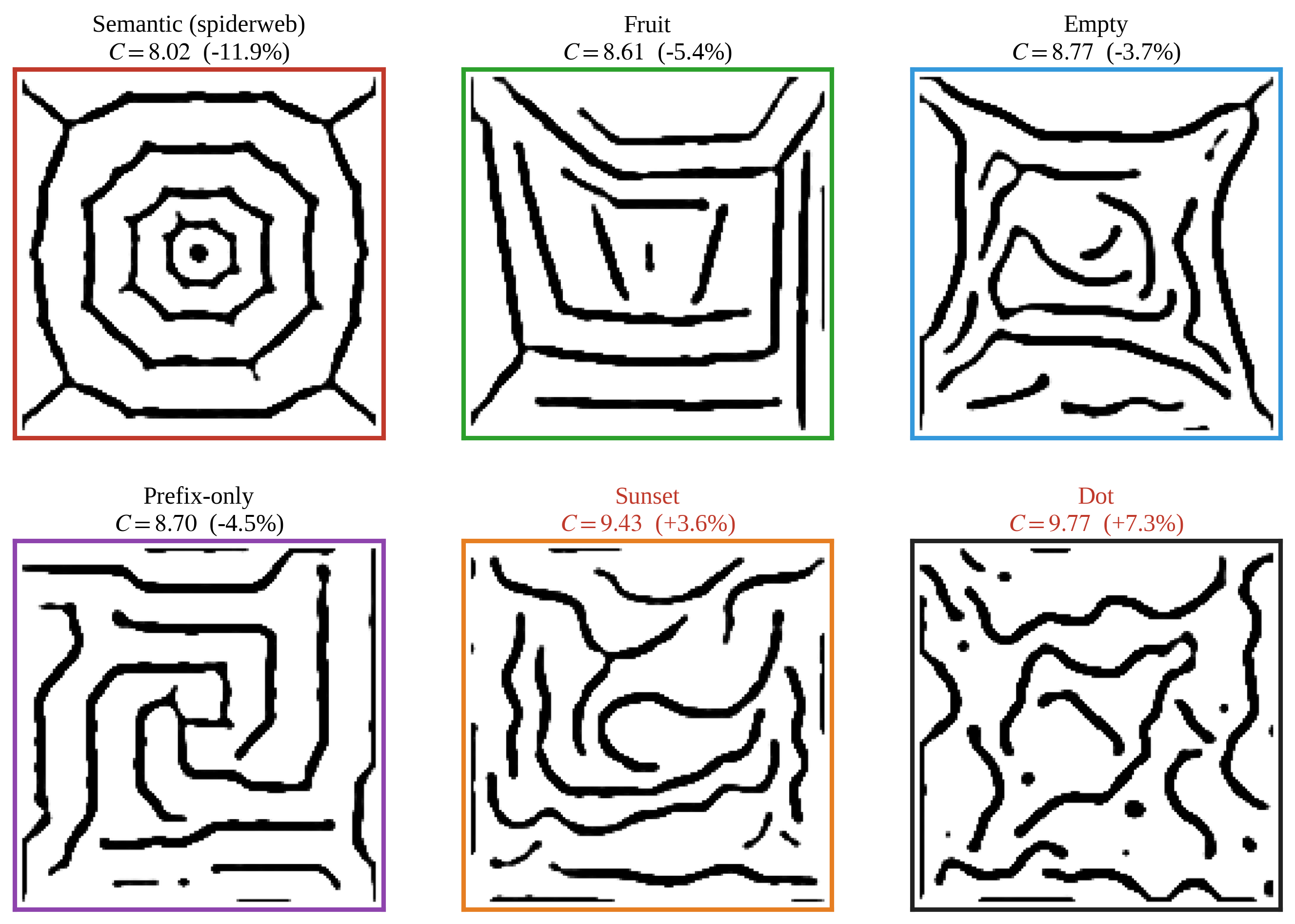}}\\[2pt]
\subfloat[Endpoint count vs.\ compliance change]{\includegraphics[width=0.42\linewidth]{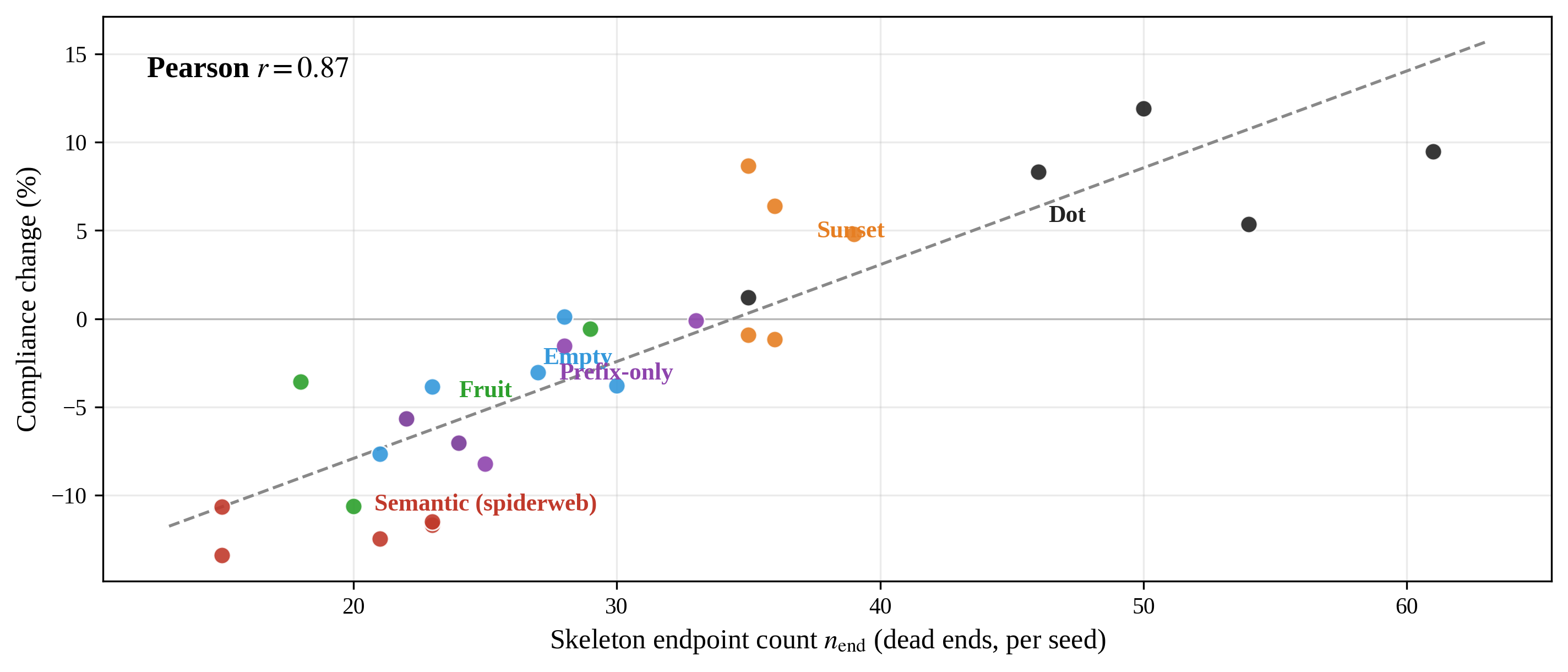}}}
\caption{\rev{Prompt comparison on $D_{\text{Rect}}$. (a) Optimized rib layouts under six prompts (one representative seed each; the labeled $C$ is the five-seed mean, with \textit{Spiderweb} the main-experiment reference), ordered from the best structural prompt (\textit{Spiderweb}) to the \textit{dot} prompt: as the prompt shifts from connected line networks toward isolated points, the layout accumulates fragmented, non-load-bearing stubs. (b) Per-seed compliance change versus per-seed skeleton endpoint count $n_\text{end}$ (five seeds for each of the six prompts, 30 runs); the two correlate at Pearson $r=0.87$, and the two highest-$n_\text{end}$ conditions (\textit{dot} and \textit{sunset}) are the ones that degrade the baseline, the \textit{dot} prompt most.}}
\label{fig:prompt_ablation}
\end{figure}
In most domains, therefore, SDS improves designs by eliminating non-load-bearing dead ends rather than by adding branches.
The physical rationale is direct: a dead-end branch, disconnected from any support pathway, carries near-zero stress and contributes negligibly to stiffness, yet consumes volume fraction that the bisection constraint must redistribute from load-carrying members elsewhere.
Each eliminated dead end therefore frees material budget for primary load paths without violating the volume target, explaining why endpoint count (not branch count or connectivity alone) tracks compliance across diverse physics settings.

\subsection{Physical explanation: escape from local optima}
\label{sec:escape}

\begin{figure}[pos=htbp]
\centering
\subfloat[$D_{\text{Rect}}$]{\raisebox{-.5\height}{\includegraphics[width=0.32\textwidth]{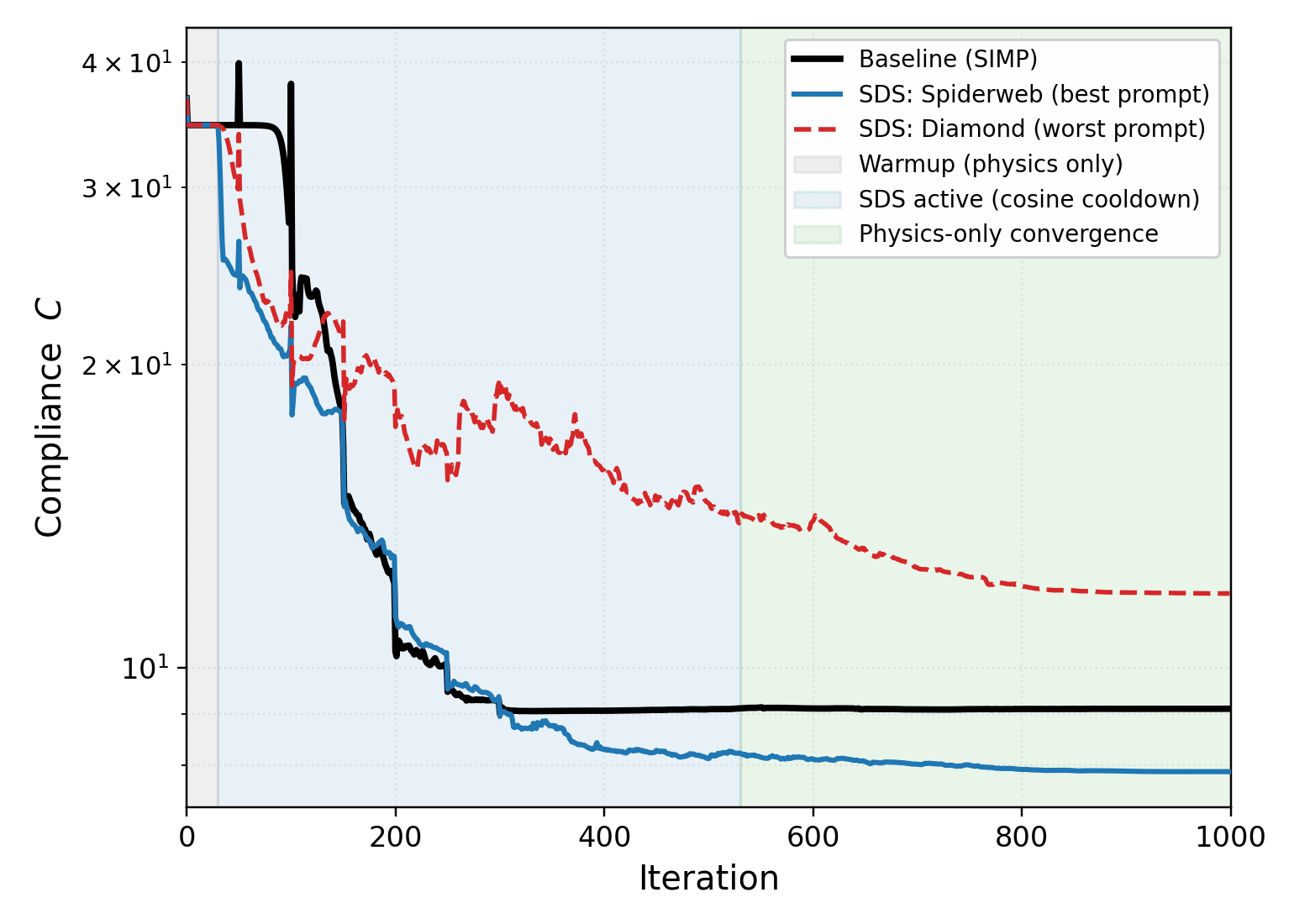}}}\hfill
\subfloat[$D_{\text{Circle}}$]{\raisebox{-.5\height}{\includegraphics[width=0.32\textwidth]{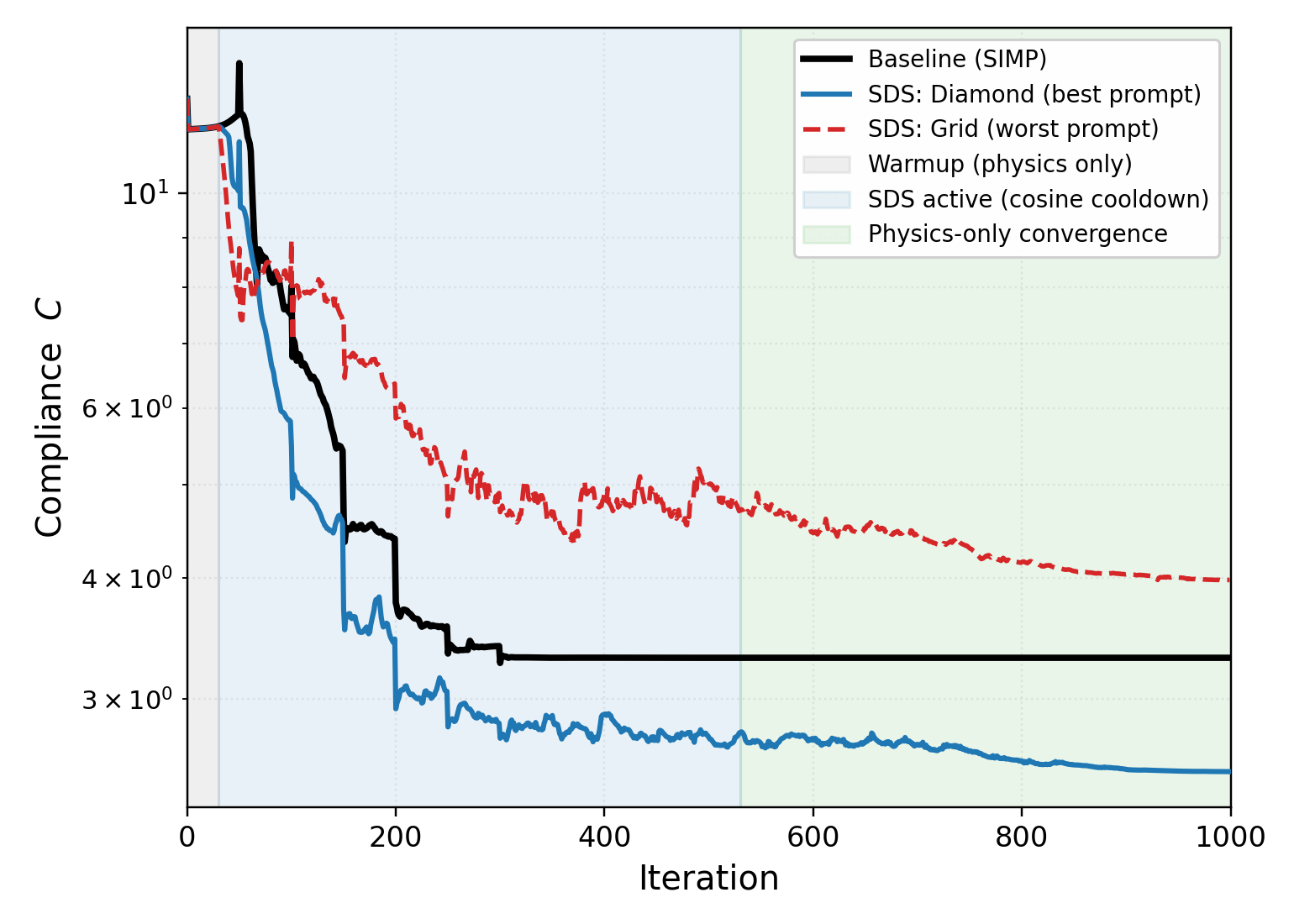}}}\hfill
\subfloat[$D_{\text{Hole}}$]{\raisebox{-.5\height}{\includegraphics[width=0.32\textwidth]{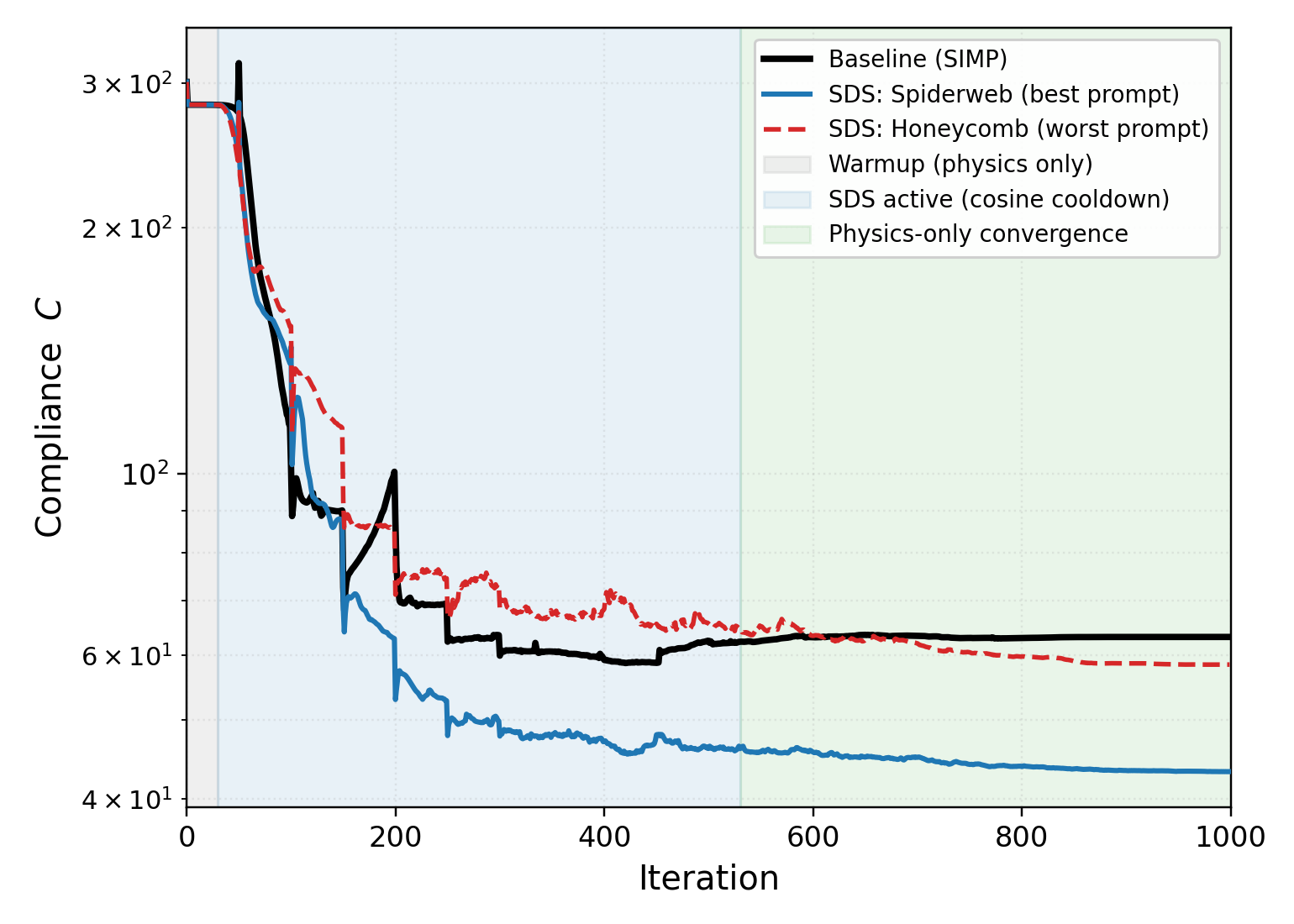}}}
\caption{Compliance convergence history on a log-scale $y$-axis for the three plate-bending domains. Each panel superimposes the baseline SIMP trajectory (black), the best-prompt best-seed run (blue), and a high-compliance worst-prompt run (red dashed). Background shading marks the three scheduling phases: warmup (gray, 0--30), SDS-active cosine cooldown (blue, 30--530), and physics-only convergence (green, 530--1{,}000).}
\label{fig:convergence}
\end{figure}

The convergence histories (Fig.~\ref{fig:convergence}) make the optimization-level consequence of dead-end suppression directly visible.
The SDS gradient deforms the effective energy landscape by adding a text-conditioned potential to the physics-based objective, enabling transitions to basins inaccessible under the physical landscape alone, analogous to how DreamFusion \cite{poole2022dreamfusion} discovers meaningful 3D structures from random NeRF initialization.
During the warmup phase all curves track each other while SIMP establishes initial load paths. Once SDS activates, the best-prompt curve separates downward from the baseline, locking in the gap by the end of the SDS-active phase.
The clearest example is $D_{\text{Circle}}$, where the best prompt reaches a mean $C=2.804$ ($-15.2\%$) compared to the baseline local optimum at $C=3.308$: a topology that SIMP from uniform initialization never reaches.

\begin{table}[pos=htbp]
\centering
\caption{Network morphology metrics of representative designs across all six experimental domains. Values are per-prompt means over $n=5$ seeds of (i) final compliance $C$, (ii) skeleton endpoint count $n_\text{end}$ (non-load-bearing terminal branches, i.e.\ dead ends), (iii) branch-point count $n_\text{branch}$, and (iv) total skeleton length in pixel units. $n_\text{end}$ correlates positively with $C$ within every domain tested (Pearson $r$ reported in the last column), whereas $n_\text{branch}$ shows no consistent sign across domains. Rows within each domain are ordered by final compliance (best first); the last row per domain is the worst-performing prompt.}
\label{tab:morphology}
\footnotesize
\begin{tabular}{llccccc}
\toprule
Domain & Prompt & $C$ & $n_\text{end}$ & $n_\text{branch}$ & $\ell_\text{skel}$ & $r(C, n_\text{end})$ \\
\midrule
\multirow{4}{*}{$D_{\text{Rect}}$ ($\lambda_\text{sds}^0=10$)}
 & Baseline            & $9.106$  & $38.0$ & $8.0$   & $1090$ & \multirow{4}{*}{$+0.96$} \\
 & \textit{Spiderweb}  & $8.022$  & $19.2$ & $40.6$  & $1299$ &  \\
 & \textit{Voronoi}    & $8.433$  & $14.8$ & $98.2$  & $1179$ &  \\
 & \textit{Grid}       & $9.643$  & $52.0$ & $17.2$  & $1040$ &  \\
\midrule
\multirow{4}{*}{$D_{\text{Circle}}$ ($\lambda_\text{sds}^0=10$)}
 & Baseline            & $3.308$  & $24.0$ & $0.0$   & $765$  & \multirow{4}{*}{$+0.93$} \\
 & \textit{Diamond}    & $2.804$  & $24.0$ & $31.2$  & $675$  &  \\
 & \textit{Spiderweb}  & $2.846$  & $14.4$ & $22.4$  & $833$  &  \\
 & \textit{Grid}       & $3.359$  & $32.2$ & $26.2$  & $727$  &  \\
\midrule
\multirow{4}{*}{$D_{\text{Hole}}$ ($\lambda_\text{sds}^0=50$)}
 & Baseline            & $63.113$ & $22.0$ & $6.0$   & $978$  & \multirow{4}{*}{$+0.56$} \\
 & \textit{Spiderweb}  & $45.22$  & $23.0$ & $18.8$  & $1100$ &  \\
 & \textit{Leaf}       & $47.30$  & $12.8$ & $23.8$  & $1048$ &  \\
 & \textit{Honeycomb}  & $56.97$  & $24.8$ & $42.8$  & $1006$ &  \\
\midrule
\multirow{4}{*}{Thermo $D_{\text{Rect}}$ ($\lambda_\text{sds}^0=100$)}
 & Baseline            & $613.6$  & $14.0$ & $48.0$  & $1056$ & \multirow{4}{*}{$+0.99$} \\
 & \textit{Spiderweb}  & $472.5$  & $16.6$ & $121.8$ & $1242$ &  \\
 & \textit{Voronoi}    & $480.5$  & $18.6$ & $130.6$ & $1204$ &  \\
 & \textit{Tree}       & $486.9$  & $20.0$ & $111.0$ & $1196$ &  \\
\midrule
\multirow{4}{*}{Thermo $D_{\text{Circle}}$ ($\lambda_\text{sds}^0=100$)}
 & Baseline            & $818.2$  & $11.0$ & $88.0$  & $1019$ & \multirow{4}{*}{$+0.99$} \\
 & \textit{Spiderweb}  & $704.6$  & $0.8$  & $24.0$  & $742$  &  \\
 & \textit{Voronoi}    & $727.7$  & $7.2$  & $79.6$  & $871$  &  \\
 & \textit{Tree}       & $739.9$  & $10.2$ & $72.2$  & $828$  &  \\
\midrule
\multirow{4}{*}{Thermo $D_{\text{Hole}}$ ($\lambda_\text{sds}^0=100$)}
 & Baseline            & $556.9$  & $28.0$ & $90.0$  & $1087$ & \multirow{4}{*}{$+0.98$} \\
 & \textit{Voronoi}    & $509.6$  & $13.4$ & $87.6$  & $1079$ &  \\
 & \textit{Spiderweb}  & $516.5$  & $13.4$ & $85.4$  & $1074$ &  \\
 & \textit{Tree}       & $542.4$  & $19.2$ & $92.8$  & $1067$ &  \\
\bottomrule
\end{tabular}
\end{table}

\subsection{Boundary condition: load-path alignment}
\label{sec:alignment}

Dead-end suppression operates most effectively when the prompt's geometric bias aligns with the domain's dominant load paths, the condition that governs which prompts succeed in which domains, as detailed in Section~\ref{sec:prompt_domain}.
When prompt and load path mismatch, the generative gradient opposes the physics gradient throughout the SDS-active phase, producing the $+6\%$ degradations seen for \textit{Diamond} and \textit{Grid} in $D_{\text{Rect}}$ (failure-mode analysis in Appendix~\ref{app:fails}).
The two mechanisms are therefore complementary: dead-end suppression is the common operation. Load-path alignment determines its effectiveness.

\subsection{Resolving the gray-density tendency}
\label{sec:gray_density}

\begin{figure}[pos=htbp]
\centering
\subfloat[No Heaviside]{\raisebox{-.5\height}{\includegraphics[width=0.32\textwidth]{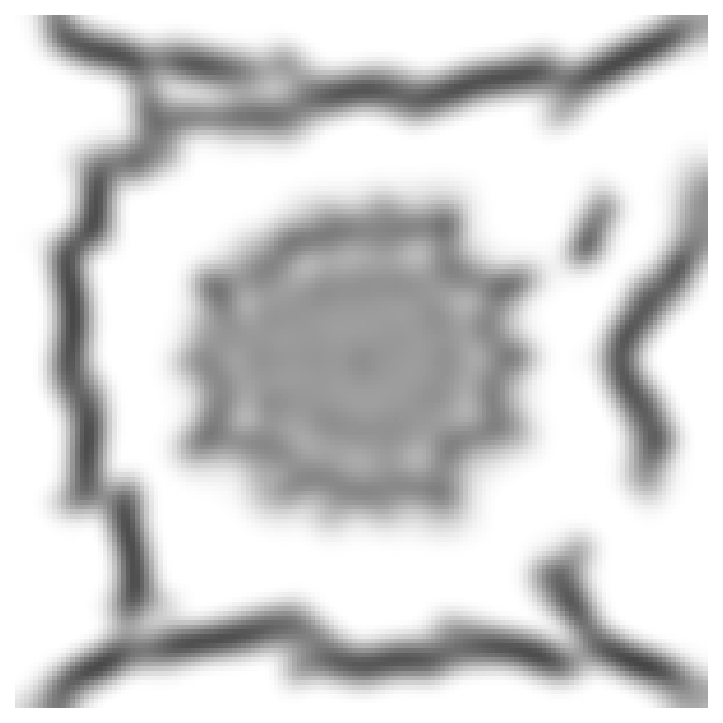}}}\hfill
\subfloat[Heaviside, $\beta_\text{max}=64$]{\raisebox{-.5\height}{\includegraphics[width=0.32\textwidth]{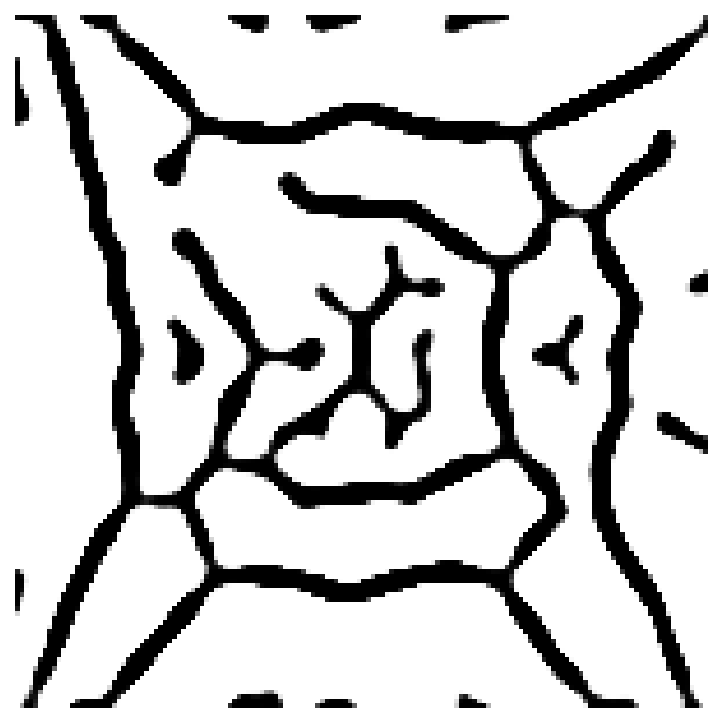}}}\hfill
\subfloat[Heaviside, $\beta_\text{max}=32$]{\raisebox{-.5\height}{\includegraphics[width=0.32\textwidth]{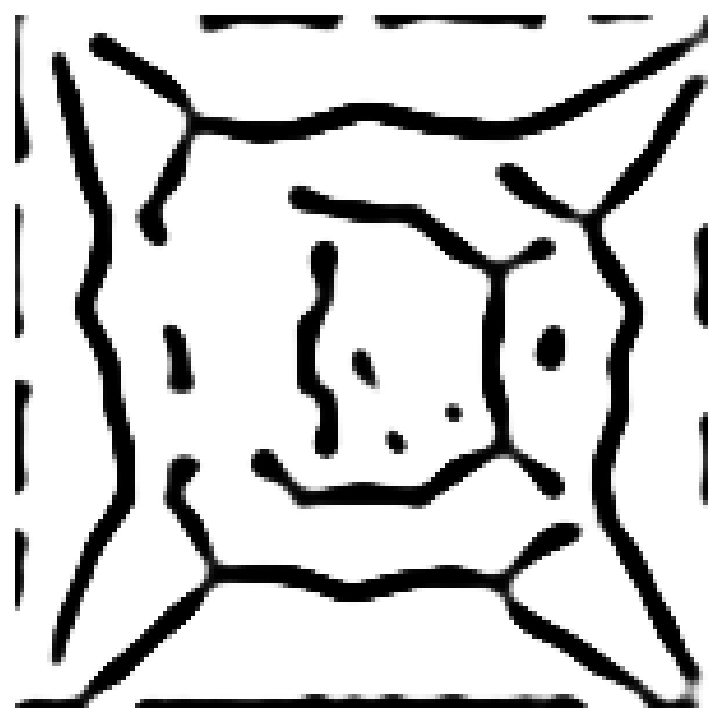}}}
\caption{Effect of Heaviside projection and $\beta$-continuation on binarization. Without Heaviside (a) the gray-density ratio reaches 42.6\%; with $\beta_\text{max}=64$ (b) it drops below 3\%.}
\label{fig:gray}
\end{figure}

SDS--SIMP coupling inherits a distinct manufacturability challenge: the score function of Stable Diffusion operates in continuous image pixel space, and the smoothness of its training distribution biases the density field toward intermediate values.
Without correction, the gray-density ratio of SDS-guided optimization reaches $42.6\%$, nearly double the $\sim$$20\%$ of pure SIMP (Fig.~\ref{fig:gray}).
This did not affect DreamFusion's NeRF outputs (which are inherently continuous radiance fields), but it renders SDS-guided density fields physically uninterpretable.

Heaviside projection with $\beta$-continuation \cite{guest2004achieving, wang2011projection} resolves the issue.
\rev{The projection and continuation schemes are themselves standard \cite{guest2004achieving, wang2011projection}. What is specific to this work is the identification that SDS's continuous pixel-space bias roughly doubles the gray-density ratio relative to pure SIMP, and the coupling of the $\beta$-schedule to the SDS schedule so that generative exploration and binarization do not interfere.}
Doubling $\beta$ every 50 iterations from $\beta_0=1$ to $\beta_\text{max}=64$ permits smooth global exploration early and enforces sharp binarization later, reducing the gray ratio to $2.5\%$ at negligible compliance cost (quantitative sweep over $\beta_\text{max}$ in Appendix~\ref{app:gray}).
The $\beta$-schedule synergizes with SDS scheduling: low-$\beta$ warmup coincides with early SDS activity to allow generative gradients to propagate across the entire field, while high-$\beta$ (64) phases overlap with SDS cooldown so the physics gradient finalizes the binarized design.

\subsection{Scheduling component ablation}
\label{sec:ablation}

To quantitatively verify the contribution of each component in the four-stage SDS scheduling strategy, a one-at-a-time ablation study was conducted under the $\textit{Spiderweb} \times D_{\text{Rect}} \times (\text{seed}=0)$ condition, selected because it exhibits the most consistent improvement ($-11.9\%$) across multiple seeds.
Table~\ref{tab:ablation} presents the complete results, including the thermoelastic $\lambda_\text{sds}^0$ sweep used to set the thermal SDS weight.

\begin{table}[pos=htbp]
\centering
\caption{SDS scheduling ablation (\textit{Spiderweb} prompt). The upper block perturbs scheduling hyperparameters in the mechanical $D_{\text{Rect}}$ setting ($\lambda_\text{sds}^0=10$ default, seed 0, default-SDS run $C=7.91$); the lower block sweeps $\lambda_\text{sds}^0$ in the thermoelastic $D_{\text{Rect}}$ setting ($n=5$ seeds, baseline $C=613.6$).}
\label{tab:ablation}
\begin{tabular}{@{}lcc@{}}
\toprule
Setting & Compliance & vs.\ Default (mech) / baseline (thermo) \\
\midrule
\multicolumn{3}{@{}l@{}}{\textbf{Mechanical $D_{\text{Rect}}$ ($\lambda_\text{sds}^0=10$ default, seed 0)}} \\
\midrule
\textbf{Default} & \textbf{7.91} & --- \\
\midrule
\textit{Weighting} & & \\
$\lambda_\text{sds} = 1$ & 9.60 & $+21.4\%$ \\
$\lambda_\text{sds} = 5$ & 8.38 & $+6.0\%$ \\
$\lambda_\text{sds} = 20$ & 8.45 & $+6.9\%$ \\
\midrule
\textit{Scheduling} & & \\
No $t$-annealing & 8.94 & $+13.0\%$ \\
warmup = 0 & 8.57 & $+8.3\%$ \\
warmup = 100 & 7.98 & $+0.9\%$ \\
\midrule
\textit{Guidance} & & \\
$\omega = 50$ & 8.28 & $+4.7\%$ \\
$\omega = 10$ & 8.57 & $+8.4\%$ \\
$\omega = 200$ & 8.67 & $+9.7\%$ \\
\midrule
\textit{Smoothing} & & \\
$\alpha = 0.5$ & 8.18 & $+3.4\%$ \\
$\alpha = 0.0$ (off) & 8.31 & $+5.0\%$ \\
\midrule
\multicolumn{3}{@{}l@{}}{\textbf{Thermoelastic $D_{\text{Rect}}$ ($\lambda_\text{sds}^0$ sweep, mean over 5 seeds)}} \\
\midrule
$\lambda_\text{sds}^0 = 10$ & 507.6 & $-17.3\%$ \\
$\lambda_\text{sds}^0 = 50$ & 486.6 & $-20.7\%$ \\
$\lambda_\text{sds}^0 = 100$ & \textbf{472.5} & $\mathbf{-23.0\%}$ \\
$\lambda_\text{sds}^0 = 200$ & 490.9 & $-20.0\%$ \\
\bottomrule
\end{tabular}
\end{table}

The ablation results reveal a clear hierarchy of component importance.
Gradient weighting ($\lambda_\text{sds}$) has the largest impact (+6.0--21.4\%): a U-shaped relationship is observed, in which $\lambda_\text{sds} = 1$ causes +21.4\% degradation due to insufficient SDS influence, while $\lambda_\text{sds} = 20$ causes +6.9\% degradation from SDS overwhelming the physics gradient.
The optimal $\lambda_\text{sds} = 10$ balances the two gradients, as the SDS gradient norm is experimentally approximately 0.1 times that of the physics gradient. This confirms that SDS is an auxiliary to physics-based optimization, not a replacement \cite{feng2023score, chung2023solving}.
Timestep annealing is the second most influential factor (+13.0\%): without annealing, random timesteps are sampled uniformly throughout optimization, leading to inefficient guidance in which local details are induced before global structure is established.
CFG scale ($\omega$) shows +4.7--9.7\% sensitivity, with both $\omega = 10$ (insufficient text guidance) and $\omega = 200$ (over-saturation causing gradient instability) degrading performance. $\omega = 100$ matches the original DreamFusion setting \cite{poole2022dreamfusion}.
EMA smoothing ($\alpha$) contributes +3.4--5.0\%, and warmup +0.9--8.3\%: immediate SDS activation (warmup = 0) biases the uniform initial density field in physically meaningless directions, while excessive warmup (100 iterations) delays SDS engagement to a point where $\beta$ is already elevated.

In summary, every single-component modification degrades performance, indicating that each scheduling component addresses a distinct failure mode rather than providing redundant regularization.
Gradient weight scheduling and timestep annealing serve as the primary coupling mechanisms between SDS and SIMP, EMA provides stability, and Heaviside projection is indispensable for binarization.
The thermoelastic $\lambda_\text{sds}^0$ sweep (lower block of Table~\ref{tab:ablation}) mirrors the mechanical U-shape but is shifted roughly tenfold toward larger values, quantifying the claim that the optimal SDS weight scales with the magnitude of the physics gradient under the raw weighted sum (Appendix~\ref{app:thermo_lambda}).

\section{Discussion}
\label{sec:discussion}

The central finding of this work is that engineer intent, expressed as a natural-language prompt, changes the outcome of a physics-constrained optimization by statistically significant margins.
The compliance reductions reported in Section~\ref{sec:results} are not stochastic by-products of perturbing the optimizer: in most experimental settings they trace to one operation, the suppression of non-load-bearing dead ends in the rib skeleton (Section~\ref{sec:mechanisms}).

The methodological contribution is the active gradient-level coupling of a pretrained visual prior to a finite-element sensitivity at every iteration.
Prior generative approaches to topology optimization either train a domain-specific model on thousands of labeled solutions \cite{maze2023diffusion, giannone2023aligning, bastek2024physics} or generate designs independently of physical analysis, so their outputs may be visually plausible yet physically suboptimal.
The framework introduced here keeps the diffusion model frozen and lets the FEA gradient, not a loss function trained on examples, control what survives the update.
This sets SDS-guided topology optimization apart from prior text- and data-driven approaches. Data-driven generative design needs labeled examples, and CLIP-based text guidance \cite{zhong2023topology} acts only through a global image--text similarity without pixel-level gradients. Semantic, preference-guided large-model optimization \cite{liang2025integrating} works at a complementary, higher level. Here, by contrast, the generative prior contributes pixel-level gradients directly to each density update.

This broadens how design intent can be specified.
Conventional parametric approaches encode intent through pre-chosen bases (rib orientations, truss templates, lattice cells), which implicitly restrict the design space to combinations the designer foresaw.
For instance, a rib parameterization over four orientation angles, three spacings, and two branching styles enumerates only $\sim$$24$ distinct topologies. A prompt such as ``leaf venation,'' by contrast, addresses the entire manifold of leaf-like patterns embedded in a pretrained diffusion model's billion-image prior, with no human enumeration required.
The FEA gradient continuously filters these prompt-induced candidates against physical optimality, so engineer intent becomes a direct optimization input rather than a pre-commitment that forecloses exploration.
Which regions of this expanded space actually pay off is governed by load-path alignment, as the cross-domain rank reversals make clear: \textit{Spiderweb} is excellent in $D_{\text{Rect}}/D_{\text{Circle}}/D_{\text{Hole}}$ yet drops to 9th in $D_{\text{Link}}$, and \textit{Voronoi} rises from 3rd to 1st between $D_{\text{Circle}}$ and $D_{\text{Link}}$. The framework surfaces this compatibility through physics feedback at every iteration, so the expansion is neither uniform nor automatic.
\rev{This exposes the central assumption behind using a natural-image prior for structural design: nothing guarantees that a visual prior aligns with a given load path. The results make the failure mode explicit. When the prompt-induced morphology conflicts with the dominant load paths, the generative gradient opposes the physics gradient and can degrade compliance ($+6\%$ for \textit{Diamond}/\textit{Grid} in $D_{\text{Rect}}$; see Sections~\ref{sec:prompt_domain} and \ref{sec:alignment}). The framework stays robust because the FEA sensitivity continuously constrains the prompt-induced update, and the scheduled decay of the SDS weight ensures that the final optimization stage is governed by the physics gradient alone. Cross-domain transfer of the visual prior is therefore beneficial only insofar as physics filters it, which is also why prompt screening remains necessary in practice.}

For the broader AI-for-science and knowledge-intensive design communities, the work extends SDS beyond its native vision domain into physics-constrained optimization, where the output must satisfy equilibrium and volume constraints rather than achieve visual plausibility.
Two transferable techniques emerge.
First, the optimal SDS weight $\lambda_\text{sds}^0$ scales with the magnitude of the physics gradient under the raw weighted sum, so identical prompt sets generalize across physics regimes once $\lambda_\text{sds}^0$ is retuned (empirically 10 for single-point loads, $\sim$50 for multi-point geometries with elevated baseline compliance, $\sim$100 for spatially diffuse thermal loads).
Second, $\beta$-continuation, developed here to reconcile SDS's continuous pixel-space bias with the binary nature of density fields, addresses a general problem in AI-guided optimization: reconciling continuous generative priors with discrete or binary solution spaces.
The same $\beta$-continuation should apply wherever a continuous generative prior must be reconciled with a binary design field, though this is not tested here.

The framework's benefit scales with the non-convexity of the physics landscape.
To probe this boundary, the identical pipeline was applied to 2D heat conduction topology optimization (thermal compliance $C = \mathbf{F}^\top \mathbf{T}$, conductivity interpolation $k_e = k_{\min} + \hat{\rho}_e^p (k_0 - k_{\min})$ with $k_0 = 1.0$, $k_{\min} = 10^{-3}$ \cite{andreassen2011efficient}) on a representative $D_{\text{Rect}}$ configuration: a center point source with $T = 0$ on all four edges, three prompts $\times$ five seeds.
SDS guidance either matched or worsened thermal compliance relative to the \textit{Adam} baseline across all prompts (Table~\ref{tab:heat}).
This negative result is consistent with the central finding of this study: heat conduction under SIMP penalization produces a fundamentally smoother objective landscape than structural compliance, and the optimal topology is typically a branching tree whose global structure is largely determined by the source--sink geometry, leaving few alternative topologies for SDS to explore.
The baseline SIMP optimizer already converges to near-optimal branching networks, and the SDS gradient introduces perturbations that disrupt rather than improve these solutions.
Pretrained generative priors are therefore most valuable for physics problems where gradient-based optimizers are known to trap severely, such as bending-dominated compliance, fluid--structure coupling, photonic crystal design, or metamaterial unit-cell optimization.

\begin{table}[pos=htbp]
\centering
\caption{Heat conduction results: SDS does not improve thermal compliance. Values are thermal compliance $C$ on the $D_{\text{Rect}}$ center-source configuration (mean $\pm$ std over 5 seeds); lower is better.}
\label{tab:heat}
\begin{tabular}{lcccc}
\toprule
 & Baseline (\textit{Adam}) & \textit{Spiderweb} & \textit{Voronoi} & \textit{Tree} \\
\midrule
$C$ & 3.547 & $3.718 \pm 0.44$ & $3.969 \pm 0.07$ & $3.973 \pm 0.15$ \\
\bottomrule
\end{tabular}
\end{table}

Two practical limitations bound the present study.
First, the framework is 2D: the density-to-image encoding feeds a 2D diffusion prior, so a 3D extension must first resolve the dimensionality mismatch between 3D density fields and 2D priors, with attendant discretization artifacts such as stair-stepping.
Second, prompt selection is manual, and the cross-domain rank reversals confirm that no single prompt is universally optimal.
The latter motivates a practical screening protocol for deployment: when retargeting to a new domain, one may try three to five prompts from distinct structural categories, one radial (\textit{Spiderweb}), one distributed-branching (\textit{Leaf} or \textit{Bone}), and one tiling (\textit{Grid} or \textit{Honeycomb}), with single-seed pilot runs, promoting only the shortlisted prompts to a full five-seed evaluation. The present study instead evaluates all prompts at five seeds to characterize the complete prompt--domain landscape.
Strong prompt--load-path alignment is typically visible within the first $\sim$200 iterations of the SDS-active phase. The compliance curve of a well-matched prompt separates downward from the baseline before $k\approx 230$, whereas poorly matched prompts stagnate or drift upward by the same point.
This early signal, combined with the U-shaped $\lambda_\text{sds}^0$ sweep (Table~\ref{tab:ablation}; Appendix~\ref{app:thermo_lambda}), reduces the search cost of adapting the framework to an unfamiliar physics regime to a handful of screening runs.

\section{Conclusions}
\label{sec:conclusion}

This paper presented a zero-shot rib design framework that merges a training-free generative prior with density-based topology optimization.
A frozen Stable Diffusion 2.1 is coupled to a Mindlin--Reissner plate (and plane-stress) SIMP solver through score distillation sampling, so that a natural-language prompt acts as an explicit knowledge representation of engineer design intent whose physical consequences are arbitrated by the FEA sensitivity at every iteration.
Four-stage scheduling (warmup, cosine cooldown, timestep annealing, EMA smoothing) stabilizes the coupling. Heaviside projection with $\beta$-continuation reduces the SDS-induced gray-density ratio from $42.6\%$ to below $3\%$, and an automated skeleton-based pipeline converts converged density fields into \rev{CAD-convertible candidate} STEP geometry in seconds.

Across 245 primary SDS runs spanning four geometric domains and two physics regimes, 38 of 49 prompt--domain combinations achieved statistically significant compliance reductions that outperform multi-start, perturbation, and image-guided baselines (Section~\ref{sec:results}). \rev{These 38 combinations remain significant under a Benjamini--Hochberg false-discovery-rate correction across all 49 tests ($q<0.05$), and carry a median Cohen's $|d|=1.47$.}
These gains trace, in most domains, to a physically interpretable signature (dead-end suppression in the rib skeleton) and a load-path alignment hypothesis explains which prompts succeed in which domains, while a negative result on heat conduction delineates the scope: pretrained generative priors help most where the physics landscape traps gradient-based optimizers severely.

Future work will extend the framework to 3D via multi-view SDS or native 3D diffusion priors \cite{jun2023shap, shi2023mvdream}, automate prior discovery through optimization in the continuous text-embedding space (textual inversion), and explore physics regimes with severe local-optima trapping such as fluid--structure interaction and metamaterial unit-cell design.
More broadly, this work uses a pretrained generative model as a design interface rather than a design generator: coupling a text prompt to a physics-constrained optimizer produces rib layouts that parametric methods and data-driven generative models do not reach.


\section*{Declaration of competing interest}
The authors declare that they have no known competing financial interests or personal relationships that could have appeared to influence the work reported in this paper.

\section*{Funding}
This work was supported by grants from the Ministry of Science and ICT (GTL24033-000, N10250154, and No. 2022-0-00986), the Ministry of Trade, Industry and Energy (RS-2025-02317327 and RS-2025-25444634), the Ministry of Oceans and Fisheries (PET0050), and Korea Hydro \& Nuclear Power Co., Ltd. (No. 8-Tech-07).

\section*{Data availability}
No training datasets were generated or used in this study.
The proposed framework is training-free and relies on a publicly available pretrained diffusion model (Stable Diffusion 2.1, \url{https://huggingface.co/stabilityai/stable-diffusion-2-1}).
All numerical results reported in this paper are computational outputs of the framework, fully determined by the code, input specifications (text prompts, domain geometries, boundary conditions), and random seeds documented in the main text and appendices.
The implementation code developed for this study will be made available by the corresponding author upon reasonable request for non-commercial research use.

\printcredits

\appendix
\numberwithin{equation}{section}
\numberwithin{figure}{section}
\numberwithin{table}{section}

\section{Finite element and SIMP modeling details}
\label{app:fea_simp}

\subsection{Mindlin--Reissner plate FEA}
\label{app:fea}

The Mindlin--Reissner plate theory \cite{mindlin1951influence, reissner1945effect} is adopted to analyze the bending behavior of rib-reinforced plates.
Since rib-reinforced plates behave locally as thick plates in the rib regions, the Mindlin--Reissner theory, which accounts for transverse shear deformation, is more appropriate than the Kirchhoff plate theory, which neglects shear deformation.
The displacement field is defined as follows:
\begin{equation}
u(x,y,z) = z \cdot \theta_y(x,y), \quad v(x,y,z) = -z \cdot \theta_x(x,y), \quad w = w(x,y)
\end{equation}
where $w$ is the transverse displacement, $\theta_x$ the rotation about the $x$-axis, and $\theta_y$ the rotation about the $y$-axis.
The bending strains and shear strains are given respectively as:
\begin{equation}
\boldsymbol{\kappa} = \begin{pmatrix} \partial\theta_y/\partial x \\ -\partial\theta_x/\partial y \\ -\partial\theta_x/\partial x + \partial\theta_y/\partial y \end{pmatrix}, \quad
\boldsymbol{\gamma} = \begin{pmatrix} \partial w/\partial x + \theta_y \\ \partial w/\partial y - \theta_x \end{pmatrix}
\end{equation}
In the Kirchhoff theory, $\boldsymbol{\gamma} = \mathbf{0}$ is enforced, whereas the Mindlin--Reissner theory permits $\boldsymbol{\gamma} \neq \mathbf{0}$, thereby accurately capturing the shear strain energy in thick plates.
By explicitly accounting for transverse shear deformation ($\theta_x \neq \partial w / \partial x$, $\theta_y \neq \partial w / \partial y$), the Mindlin--Reissner theory is well suited for structures with localized stiffness variations such as ribs.

To numerically implement this theory without shear locking, the MITC4 element is employed \cite{bathe1985four}.
This 4-node bilinear quadrilateral element has three degrees of freedom per node $(w, \theta_x, \theta_y)$, and the shape functions are $N_i(\xi, \eta) = \frac{1}{4}(1 + \xi_i\xi)(1 + \eta_i\eta)$.
The element stiffness matrix is decomposed into bending and shear contributions as given in Eq.~\eqref{eq:element_stiffness}, where $\mathbf{D}_b$ is the bending constitutive matrix ($3 \times 3$) and $\mathbf{D}_s$ is the shear constitutive matrix ($2 \times 2$, including the Reissner correction factor $\kappa_s = 5/6$).
A standard displacement-based 4-node Mindlin element suffers from shear locking in thin plates due to parasitic shear energy.
The MITC4 element overcomes this by using an assumed strain technique that evaluates shear strains at four tying points located at the midpoints of the element edges and then interpolates them into the interior, yielding accurate solutions even in the thin-plate limit ($h \to 0$).

\subsection{SIMP interpolation: rib stiffness modeling}
\label{app:simp}

In conventional SIMP, the elastic modulus is interpolated as $E_e = \rho_e^p E_0$, effectively reducing the stiffness of zero-density regions to nearly zero.
However, in the rib-reinforced plate model, zero-density regions still have physical substance, as in the base plate.
Therefore, a modified SIMP scheme is adopted that directly interpolates bending and shear stiffnesses based on the presence or absence of ribs.
Regions with attached ribs have an effective thickness $t_{\max}$ (base plate thickness + rib height), while regions without ribs have only the base plate thickness $t_{\min}$.
Since the flexural rigidity of a plate is proportional to the cube of the thickness ($D_b \propto t^3$) and the shear stiffness is linearly proportional to the thickness ($D_s \propto t$), the stiffness varies significantly depending on the presence or absence of ribs.
To reflect this physical difference, the stiffness matrices for the two limiting states are defined as follows:
\begin{equation}
\mathbf{D}_b^{\max} = \frac{E t^3_{\max}}{12(1-\nu^2)} \mathbf{C}_b, \qquad \mathbf{D}_b^{\min} = \frac{E t_{\min}^3}{12(1-\nu^2)} \mathbf{C}_b
\end{equation}
\begin{equation}
\mathbf{D}_s^{\max} = \frac{\kappa_s E t_{\max}}{2(1+\nu)} \mathbf{I}_2, \qquad \mathbf{D}_s^{\min} = \frac{\kappa_s E t_{\min}}{2(1+\nu)} \mathbf{I}_2
\end{equation}
where $\mathbf{C}_b$ is the normalized bending constitutive matrix, $\mathbf{I}_2$ is the $2 \times 2$ identity matrix, and $\kappa_s = 5/6$ is the Reissner shear correction factor.
For the physical density $\hat{\rho}_e$, the bending and shear constitutive matrices of each element are interpolated as follows:
\begin{equation}
\mathbf{D}_b^e(\hat{\rho}_e) = \hat{\rho}_e^{p}\mathbf{D}_b^{\max} + (1 - \hat{\rho}_e^{p}) \mathbf{D}_b^{\min}
\end{equation}
\begin{equation}
\mathbf{D}_s^e(\hat{\rho}_e) = \hat{\rho}_e^{p} \mathbf{D}_s^{\max} + (1 - \hat{\rho}_e^{p}) \mathbf{D}_s^{\min}
\end{equation}
where $p = 3$ is the penalty parameter. This interpolation possesses the following properties:
\begin{itemize}
\item $\hat{\rho}_e = 1$ (rib present): $\mathbf{D}_b^e = \mathbf{D}_b^{\max}$, corresponding to the maximum bending stiffness at $t_{\max}$.
\item $\hat{\rho}_e = 0$ (base plate only): $\mathbf{D}_b^e = \mathbf{D}_b^{\min}$, corresponding to the baseline bending stiffness at $t_{\min}$.
\item Since $t_{\min} > 0$, $\mathbf{D}_b^{\min} > 0$ is guaranteed, preventing the unphysical assumption that plate regions without ribs bear no load whatsoever.
\end{itemize}
Accordingly, the element stiffness matrix is expressed as follows:
\begin{equation}
\mathbf{K}_e(\hat{\rho}_e) = \int_{\Omega_e} \mathbf{B}_b^T
\mathbf{D}_b^e(\hat{\rho}_e) \mathbf{B}_b \,d\Omega + \int_{\Omega_e} \mathbf{B}_s^{\text{MITC},T} \mathbf{D}_s^e(\hat{\rho}_e) \mathbf{B}_s^{\text{MITC}} \,d\Omega
\end{equation}
Since both $\mathbf{D}_b^e$ and $\mathbf{D}_s^e$ follow the same interpolation form $\hat{\rho}_e^p(\cdot)^{\max} + (1 - \hat{\rho}_e^p)(\cdot)^{\min}$ with respect to $\hat{\rho}_e$, aggregating at the element stiffness matrix level yields $\mathbf{K}_e(\hat{\rho}_e) = \hat{\rho}_e^{p} \mathbf{K}_e^{\max} + (1 - \hat{\rho}_e^{p})\mathbf{K}_e^{\min}$.
Here, $\mathbf{K}_e^{\max}$ and $\mathbf{K}_e^{\min}$ are the element stiffness matrices corresponding to $t_{\max}$ and $t_{\min}$, respectively, and can be precomputed to reduce the stiffness matrix assembly cost at each iteration.

\subsection{Density filter}
\label{app:filter}

Checkerboard patterns and mesh dependency are well-known numerical pathologies in density-based topology optimization \cite{sigmund1998numerical}.
A checkerboard pattern, in which densities of 0 and 1 alternate between adjacent elements, achieves low compliance mathematically but is impossible to manufacture, constituting an artifact of finite element discretization.
To prevent this, a spatially weighted averaging filter is applied to the raw density $\boldsymbol{\rho}$ \cite{bourdin2001filters, bruns2001topology}:
\begin{equation}
\tilde{\rho}_e = \frac{\sum_{j \in N_e} w_{ej} \rho_j}{\sum_{j \in N_e} w_{ej}}, \quad w_{ej} = \max(0, r_\text{min} - \text{dist}(e, j))
\end{equation}
where $N_e$ is the set of neighboring elements within a radius $r_\text{min}$ of element $e$, and $w_{ej}$ is a linear weight inversely proportional to the distance.
In this study, $r_\text{min} = 5$ (in element size units) is used.
This filtering suppresses checkerboard patterns by smoothing abrupt density variations between adjacent elements, and consequently provides indirect control of the minimum feature size to approximately $r_\text{min}$.
The derivative of the filter is given by:
\begin{equation}
\frac{\partial \tilde{\rho}_e}{\partial \rho_j} = \frac{w_{ej}}{\sum_{k \in N_e} w_{ek}}
\end{equation}

\subsection{Maximum feature size control ($r_\text{max}$)}
\label{app:rmax}

While the density filter ($r_\text{min}$) controls the minimum member size, restricting the maximum member thickness is also important for manufacturability in rib reinforcement design.
Excessively thick ribs cause shrinkage defects during casting and reduce material efficiency.
In this study, the local volume constraint approach \cite{guest2009topology} is adopted to indirectly control the maximum member thickness to approximately $2 r_\text{max}$.
Specifically, a spatial averaging filter with the same structure as the $r_\text{min}$ filter is constructed with radius $r_\text{max}$, computing the average density in the neighborhood of each element $e$:
\begin{equation}
\bar{\rho}_e^{\text{max}} = \frac{\sum_{j \in N_e^{\text{max}}} w_{ej}^{\text{max}} \rho_j}{\sum_{j \in N_e^{\text{max}}} w_{ej}^{\text{max}}}, \quad w_{ej}^{\text{max}} = \max(0, r_\text{max} - \text{dist}(e, j))
\end{equation}
If this local average exceeds a threshold $\eta_\text{max}$, it indicates that material is overly concentrated in that region. This is suppressed via a quadratic penalty:
\begin{equation}
\mathcal{L}_\text{max} = \lambda_\text{max} \sum_{e} \left[\max(0, \bar{\rho}_e^{\text{max}} - \eta_\text{max})\right]^2
\end{equation}
where $\lambda_\text{max} = 10$ is the penalty weight and $\eta_\text{max} = 2 V^*$ is the default threshold.
The sensitivity of this penalty is efficiently computed through the chain rule of the averaging filter:
\begin{equation}
\frac{\partial \mathcal{L}_\text{max}}{\partial \rho_j} = \lambda_\text{max} \sum_{e} \frac{2 w_{ej}^{\text{max}}}{\sum_{k} w_{ek}^{\text{max}}} \max(0, \bar{\rho}_e^{\text{max}} - \eta_\text{max})
\end{equation}
This gradient disperses material away from thick concentration regions, guiding the rib network toward thinner, more distributed patterns.

Because the $r_\text{max}$ constraint caps how thick SDS-induced members may become, an overly aggressive setting can force strongly clustered patterns (e.g., thick radial branches) to thin out and converge to a less efficient compromise.
The value $r_\text{max}=10$ (with $r_\text{min}=5$) was adopted throughout this study to balance manufacturability against this loss of design freedom.

\subsection{Heaviside projection and $\beta$-continuation}
\label{app:heaviside}

Density filtering alone does not completely eliminate intermediate densities (gray elements). In fact, the filter tends to smooth density boundaries, widening the gray zone.
To address this, the smoothed Heaviside projection \cite{guest2004achieving} is applied to transform the filtered density $\tilde{\rho}_e$ into a physical density $\hat{\rho}_e$ that closely approximates a binarized distribution:
\begin{equation}
\hat{\rho}_e
= \frac{\tanh(\beta \eta) + \tanh\!\bigl(\beta(\tilde{\rho}_e - \eta)\bigr)}
       {\tanh(\beta \eta) + \tanh\!\bigl(\beta(1 - \eta)\bigr)}
\end{equation}
where $\eta = 0.5$ is the threshold and $\beta$ controls the projection sharpness: as $\beta \to \infty$, the function approaches the ideal Heaviside step function.

Setting $\beta$ to a large value from the outset causes convergence instability due to severe non-convexity \cite{wang2011projection}.
A continuation strategy is therefore applied in which $\beta$ starts at $\beta_0 = 1$ and doubles every $N_\beta = 50$ iterations until reaching $\beta_\text{max} = 64$:
\begin{equation}
\beta(k)
= \min\!\left(\beta_0 \cdot 2^{\lfloor k / N_\beta \rfloor},\;
              \beta_\text{max}\right)
\end{equation}
This $\beta$-continuation is particularly important in the present study because the SDS gradient inherently operates toward a continuous image space, tending to induce intermediate densities.
A natural two-phase division of labor is established: SDS forms global patterns during the low-$\beta$ stages, and the subsequent increase in $\beta$ sharply binarizes these patterns.
The effectiveness of this mechanism is quantitatively analyzed in Appendix~\ref{app:gray}.

\subsection{Compliance sensitivity (physics gradient)}
\label{app:sensitivity}

Exploiting the self-adjointness of the compliance $C = \mathbf{F}^\top \mathbf{U}$, the sensitivity with respect to the physical density $\hat{\rho}_e$ is derived as shown in Eq.~\eqref{app:eq:sensitivity}, where $\mathbf{u}_e$ and $\mathbf{K}_e$ denote the element-level displacement vector and stiffness matrix, respectively.
This sensitivity, further extended via the chain rule in Eq.~\eqref{app:eq:chainrule}, constitutes the physics gradient $\nabla_{\rho}C$ used in the combined optimization loop.
\begin{equation}
\label{app:eq:sensitivity}
\frac{\partial C}{\partial \hat{\rho}_e} = -\mathbf{u}_e^\top \frac{\partial \mathbf{K}_e}{\partial \hat{\rho}_e} \mathbf{u}_e = -p \hat{\rho}_e^{p-1}\mathbf{u}_e^\top \left( \mathbf{K}_e^{\max} - \mathbf{K}_e^{\min} \right) \mathbf{u}_e
\end{equation}
Since $\mathbf{K}_e^{\max} - \mathbf{K}_e^{\min}$ is a constant matrix independent of density, it can be precomputed, and no additional linear system solve is required for the sensitivity computation (self-adjointness of compliance).
Extending to the raw design variables $\rho_e$ via the chain rule:
\begin{equation}
\label{app:eq:chainrule}
\frac{\partial C}{\partial \rho_e} = \frac{\partial C}{\partial \hat{\rho}_e} \cdot \frac{\partial \hat{\rho}_e}{\partial \tilde{\rho}_e} \cdot \frac{\partial \tilde{\rho}_e}{\partial \rho_e}
\end{equation}
where the last two terms represent the chain rule contributions of the Heaviside projection (Appendix~\ref{app:heaviside}) and the density filter (Appendix~\ref{app:filter}), respectively.
This complete sensitivity is denoted by the physics gradient $\nabla_\rho C$ in this study and constitutes the key component combined with the SDS gradient.
The quantity $\mathbf{u}_e^\top (\mathbf{K}_e^{\max} - \mathbf{K}_e^{\min}) \mathbf{u}_e$ represents the strain energy associated with the stiffness gain from adding a rib to element $e$.
Elements with larger values of this quantity yield greater compliance reduction upon rib placement, and thus the sensitivity provides information on where rib placement would be most efficient.

\subsection{\rev{Finite-difference verification of the thermoelastic sensitivity}}
\label{app:fd_thermo}
\rev{For the thermoelastic objective the load $\mathbf{F}_\text{th}$ is assembled from the prescribed moment $\mathbf{M}_\text{th} = \alpha_T \Delta T [1,1,0]^\top$ without any dependence on the element stiffness $E(\hat{\rho})$. Consequently $\partial \mathbf{F}_\text{th}/\partial \hat{\rho} = \mathbf{0}$: the load carries no design-derivative term, and the self-adjoint sensitivity of Eq.~\eqref{app:eq:sensitivity} applies unchanged, with only the stiffness $\mathbf{K}(\hat{\rho})$ being density-dependent. To confirm this numerically, the analytic sensitivity was compared against a central finite difference $[\,C(\hat{\rho}_e + h) - C(\hat{\rho}_e - h)\,]/(2h)$ with $h = 10^{-4}$ at $30$ randomly selected elements of a mid-density field ($\hat{\rho}_e \sim \mathcal{U}[0.3, 0.7]$). The analytic and finite-difference gradients show excellent agreement (maximum relative error $3.2 \times 10^{-4}$, mean $7.0 \times 10^{-5}$, at the level of the central-difference truncation error; Pearson correlation $0.99999995$). In the adopted prescribed-thermal-moment formulation the assembled thermal load is density-independent, so the standard self-adjoint sensitivity is consistent with this formulation and no load-derivative correction is required.}

\section{SDS coupling and optimization details}
\label{app:sds_opt}

\subsection{Density field to image conversion}
\label{app:density_image}

Section~\ref{sec:sds} describes the conversion of $\hat{\boldsymbol{\rho}} \in \mathbb{R}^{n_\text{elx} \times n_\text{ely}}$ into the diffusion-model latent $\mathbf{z}_0 = \mathcal{E}(\mathbf{x}) \in \mathbb{R}^{4 \times 64 \times 64}$ via three-channel replication, bilinear resize to $512 \times 512$, and VAE encoding.
This pipeline is fully differentiable. The chain rule for backpropagation of the SDS gradient from $\mathbf{z}_0$ to $\hat{\boldsymbol{\rho}}$ is
\begin{equation}
\frac{\partial \mathbf{z}_0}{\partial \hat{\boldsymbol{\rho}}} = \frac{\partial \mathcal{E}}{\partial \mathbf{x}} \cdot \frac{\partial \mathbf{x}}{\partial \hat{\boldsymbol{\rho}}}
\end{equation}
The full SDS pipeline, including the forward diffusion, classifier-free guidance, and back-propagation path, is illustrated in Fig.~\ref{fig:sds_process}.

\subsection{EMA smoothing}
\label{app:ema}

The SDS gradient is inherently stochastic.
Since the timestep $t$ and noise $\epsilon$ are randomly sampled at each iteration, considerable variance exists in both the direction and magnitude of the gradient.
In particular, under the setting of this study where a single-sample gradient is used without mini-batch averaging, this variance can undermine optimization stability.
To mitigate this, EMA smoothing is applied as defined in Eq.~\eqref{eq:ema} (Section~\ref{sec:scheduling}).
The smoothing coefficient is set to $\alpha = 0.9$, with gradients exponentially averaged over approximately the most recent 10 iterations.
This provides an appropriate balance between smoothing the noise of individual iterations and reflecting the temporal evolution of the SDS gradient due to timestep annealing.
The ablation study (Section~\ref{sec:ablation}) shows that $\alpha = 0$ (no EMA) results in +5.0\% degradation and $\alpha = 0.5$ yields +3.4\% degradation.

\subsection{Volume constraint enforcement}
\label{app:volume}

A bisection projection is employed to satisfy the volume fraction constraint $V(\hat{\boldsymbol{\rho}}) = V^*$ exactly at each iteration.
After the \textit{Adam} optimizer updates the densities, a uniform shift $\Delta$ is applied to all design variables such that the projected volume matches the target:
\begin{equation}
\rho_e^{(k+1)} \leftarrow \rho_e^{(k+1)} + \Delta, \quad \text{where } \Delta \text{ solves} \quad \frac{1}{N_\text{active}} \sum_{e \in \Omega_\text{active}} \hat{\rho}_e(\boldsymbol{\rho}^{(k+1)} + \Delta) = V^*
\label{app:eq:bisection}
\end{equation}
While traditional topology optimization uses the Lagrangian multiplier method \cite{bendsoe1989optimal} or \textit{MMA} \cite{svanberg1987method} to handle volume constraints, this direct bisection approach is adopted, given the compatibility with the \textit{Adam} optimizer and the stochastic nature of the SDS gradient.
The scalar $\Delta$ is found via bisection on the monotone left-hand side of Eq.~\eqref{app:eq:bisection}.
Here, $\Omega_\text{active}$ denotes the set of active elements within the domain (excluding elements outside the circle in $D_{\text{Circle}}$).
The bisection exploits the monotonicity of the Heaviside projection and converges to machine precision within approximately 30 iterations.
This method fixes the volume fraction exactly at the target value, eliminating ambiguity caused by volume-fraction differences when comparing compliance.

\subsection{Optimizer: \textit{Adam} vs.\ \textit{OC}}
\label{app:optimizer}

\begin{figure}[pos=htbp]
\centering
\subfloat[$D_{\text{Rect}}$: Compliance]{\raisebox{-.5\height}{\includegraphics[width=0.32\textwidth]{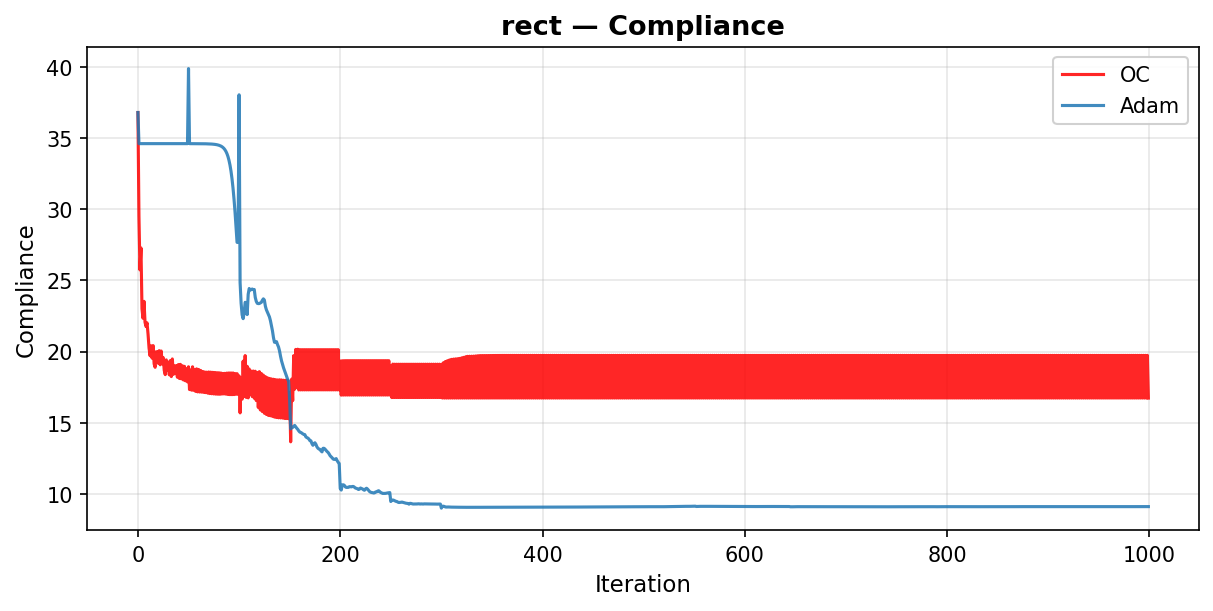}}}\hfill
\subfloat[$D_{\text{Circle}}$: Compliance]{\raisebox{-.5\height}{\includegraphics[width=0.32\textwidth]{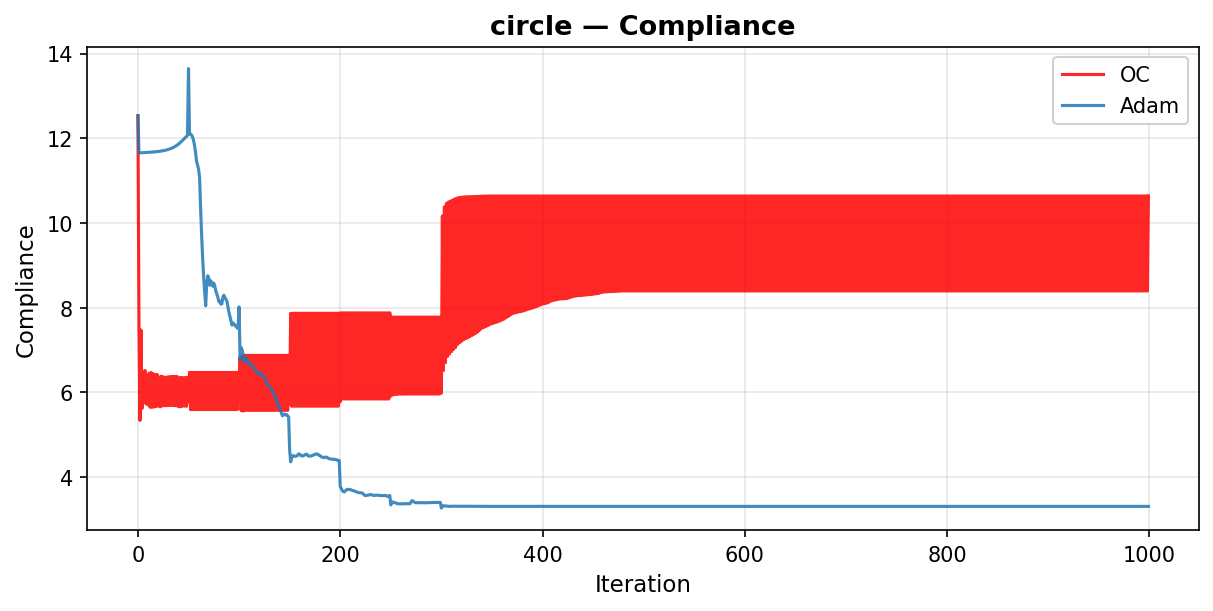}}}\hfill
\subfloat[$D_{\text{Hole}}$: Compliance]{\raisebox{-.5\height}{\includegraphics[width=0.32\textwidth]{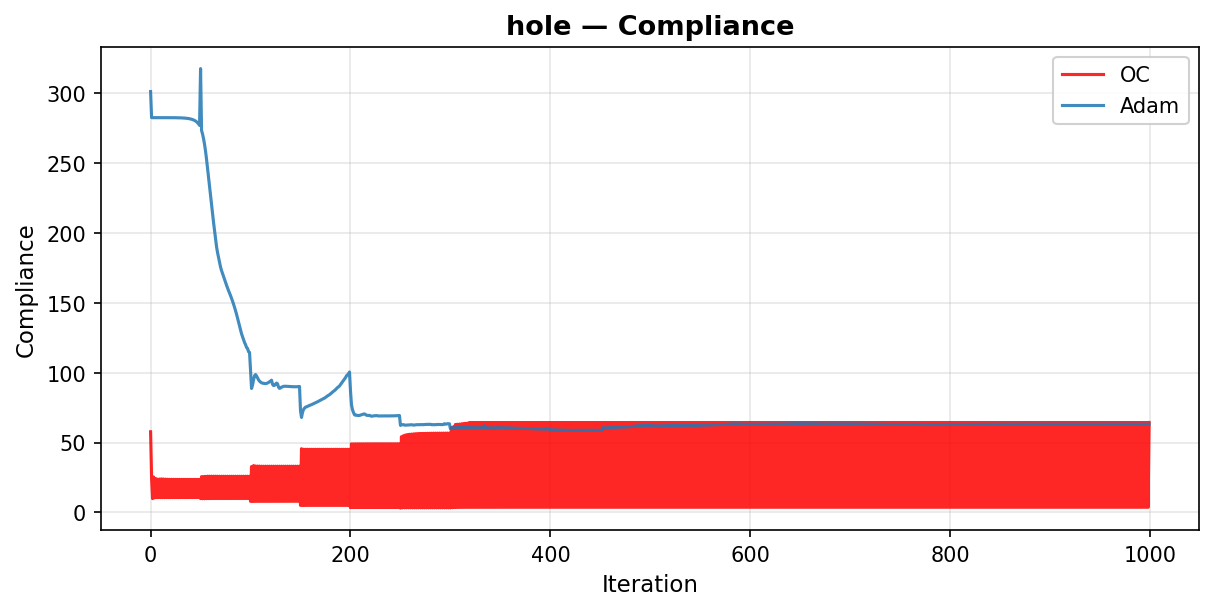}}}\\[1em]
\subfloat[$D_{\text{Rect}}$: Volume fraction]{\raisebox{-.5\height}{\includegraphics[width=0.32\textwidth]{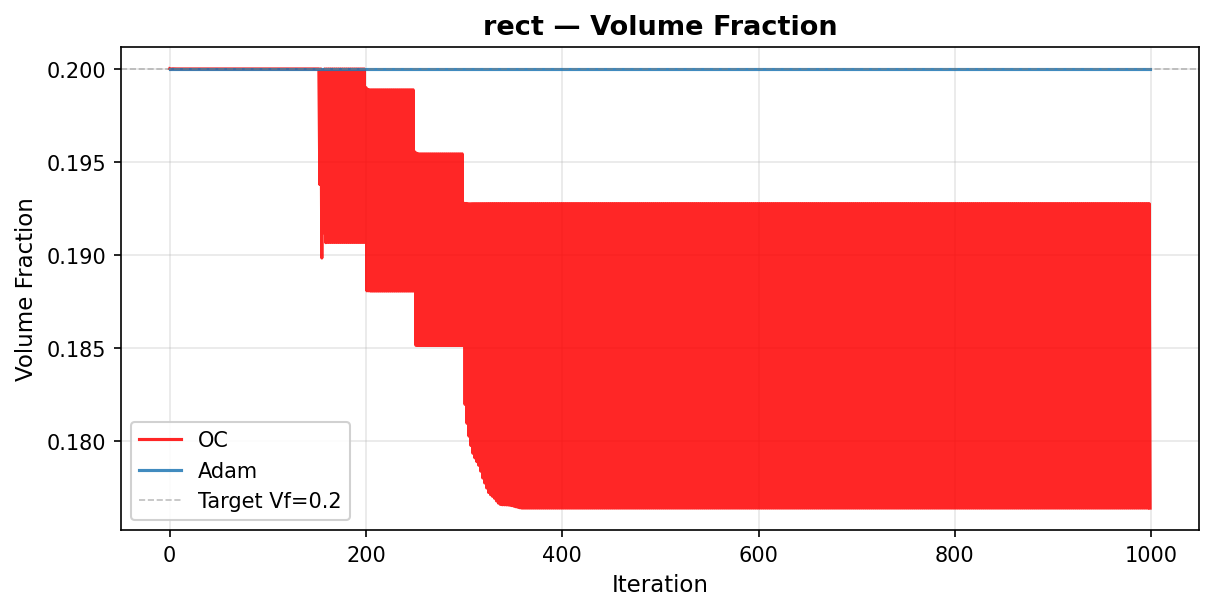}}}\hfill
\subfloat[$D_{\text{Circle}}$: Volume fraction]{\raisebox{-.5\height}{\includegraphics[width=0.32\textwidth]{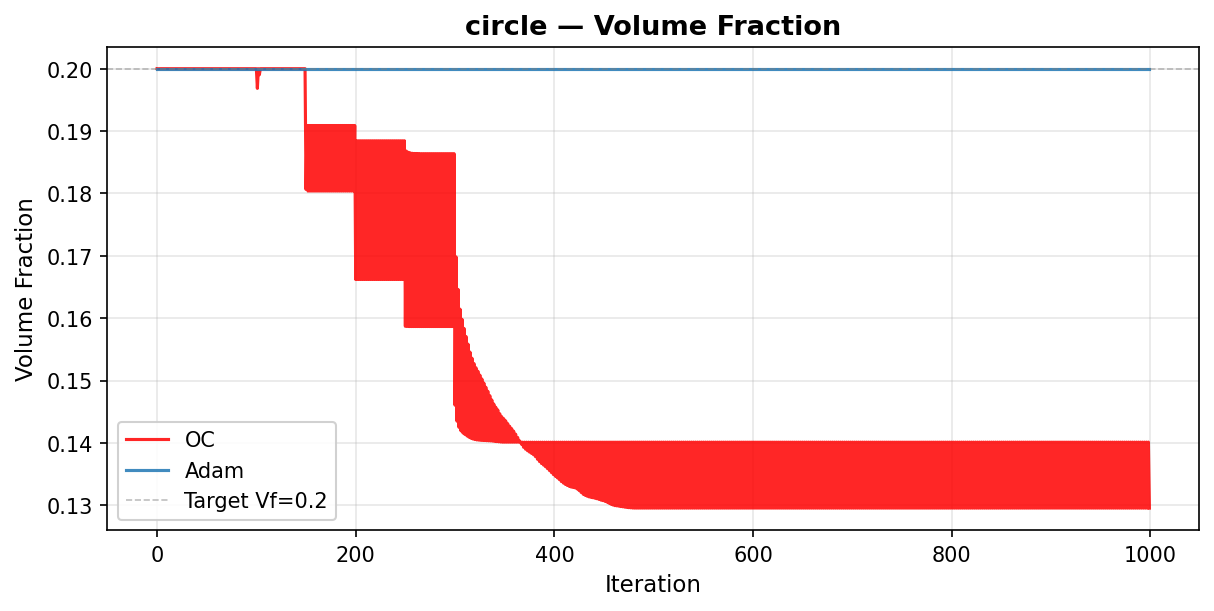}}}\hfill
\subfloat[$D_{\text{Hole}}$: Volume fraction]{\raisebox{-.5\height}{\includegraphics[width=0.32\textwidth]{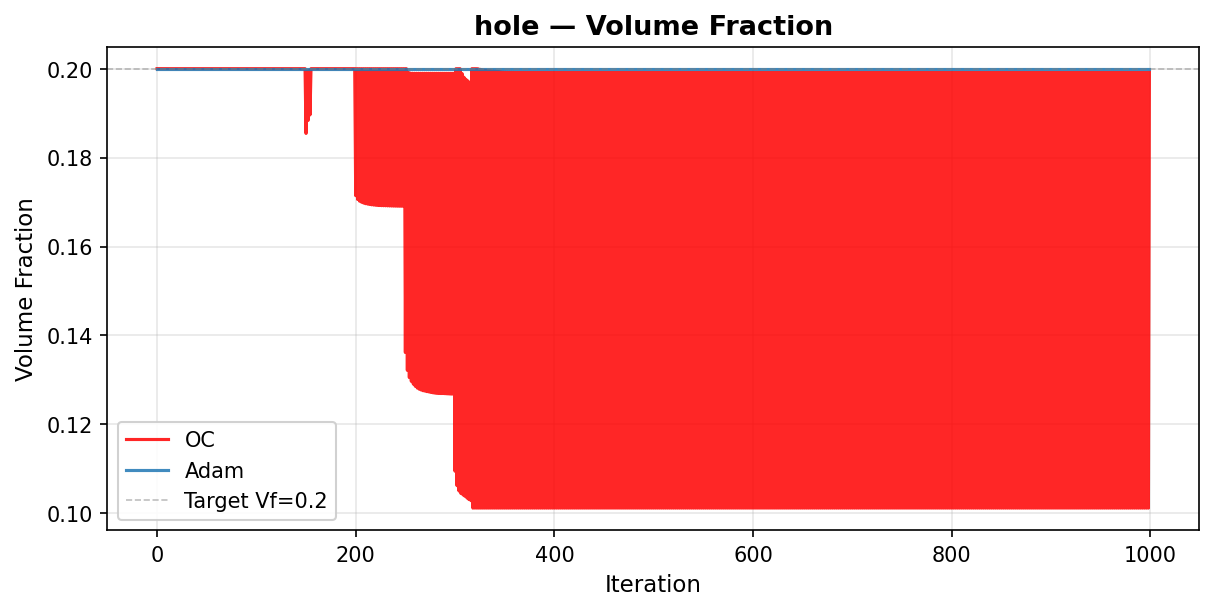}}}
\caption{Convergence comparison of \textit{OC} (oscillatory) vs.\ \textit{Adam} (stable) under Heaviside projection.}
\label{app:fig:oc_vs_adam}
\end{figure}

The \textit{Adam} optimizer \cite{kingma2014adam} is used in this study instead of \textit{OC}, which is widely employed in conventional topology optimization.
There are two reasons for this choice.
First, it offers good compatibility with the SDS gradient.
\textit{OC} is a heuristic update rule that exploits the special structure of compliance sensitivity (self-adjointness and non-negativity), making it difficult to integrate arbitrary gradients, such as the SDS gradient, in a natural way.
\textit{Adam} tracks the first and second moments of arbitrary gradients and applies an adaptive learning rate, making it well suited for combining physics and SDS gradients.
Second, it converges stably in conjunction with the Heaviside projection.
In experiments, \textit{OC} combined with Heaviside was observed to exhibit period-2 oscillation at high $\beta$ values without converging (see Fig.~\ref{app:fig:oc_vs_adam}).
This occurs because the bisection update of \textit{OC} interacts with the sharp nonlinearity of the Heaviside projection, causing oscillation between two stable attractors.
\textit{Adam} naturally damps this oscillation through momentum and learning rate decay ($0.04 \to 0.01$, cosine annealing).
The momentum coefficients ($\beta_1 = 0.5$, $\beta_2 = 0.9$) are set lower than the standard defaults ($0.9$, $0.999$) to shorten the gradient memory, allowing the optimizer to adapt quickly as the SDS weight and timestep change during the scheduling phases.
This behavior is illustrated in Fig.~\ref{app:fig:oc_vs_adam}.

\begin{figure}[pos=htbp]
\centering
\subfloat[$D_{\text{Rect}}$: \textit{OC} ($C=16.74$)]{\raisebox{-.5\height}{\includegraphics[width=0.32\textwidth]{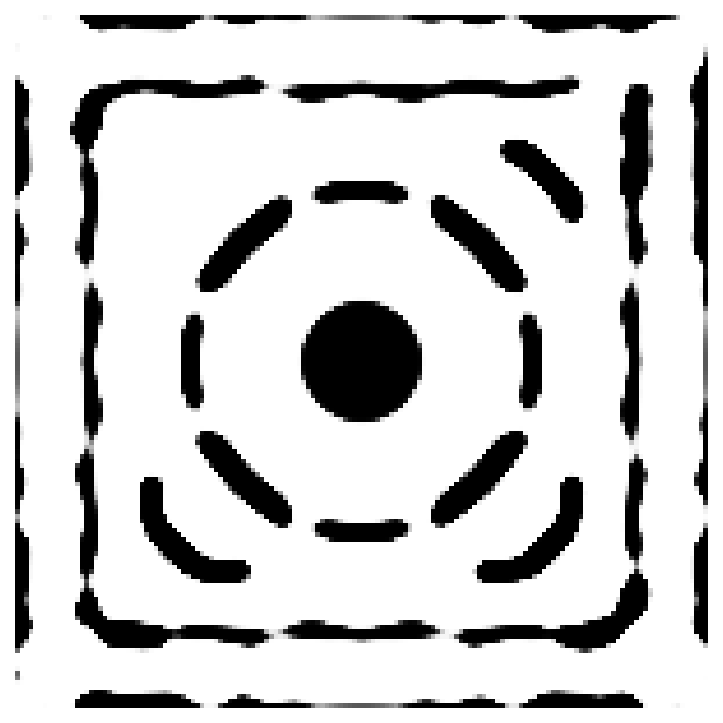}}}\hfill
\subfloat[$D_{\text{Circle}}$: \textit{OC} ($C=10.64$)]{\raisebox{-.5\height}{\includegraphics[width=0.32\textwidth]{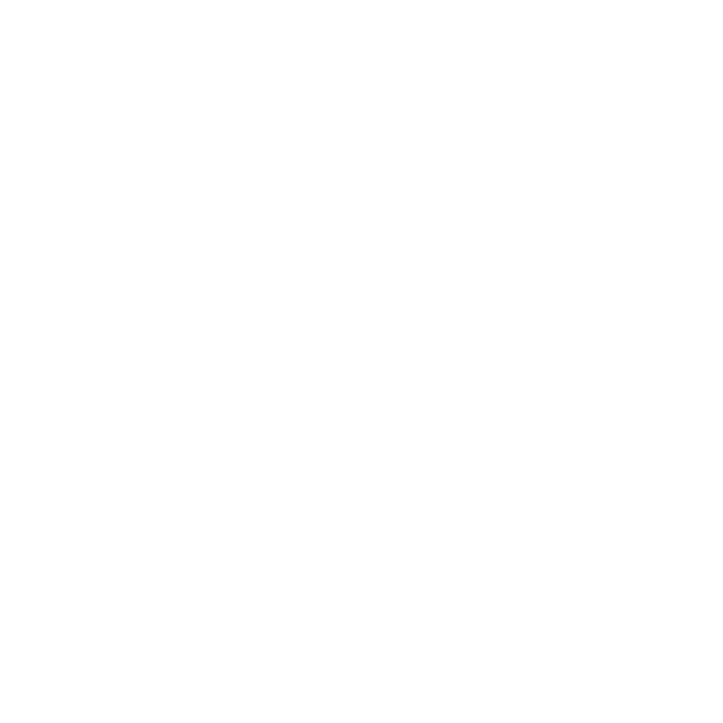}}}\hfill
\subfloat[$D_{\text{Hole}}$: \textit{OC} ($C=148.96$)]{\raisebox{-.5\height}{\includegraphics[width=0.32\textwidth]{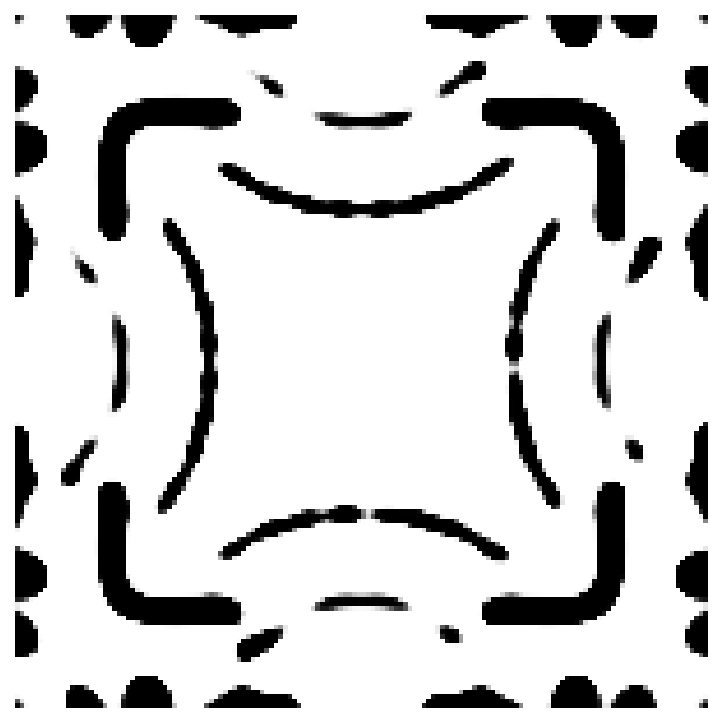}}}
\caption{\textit{OC} baseline density fields (no SDS). (a) $D_{\text{Rect}}$: disconnected rib pattern with compliance $1.84\times$ worse than \textit{Adam}. (b) $D_{\text{Circle}}$: catastrophic failure due to period-2 oscillation; volume fraction collapses to $V=0.13$, yielding a nearly empty domain. (c) $D_{\text{Hole}}$: severely fragmented density field with compliance $2.36\times$ worse than \textit{Adam}.}
\label{app:fig:oc_baseline}
\end{figure}

The same SIMP optimization (without SDS) was conducted with the conventional \textit{OC} optimizer across all three plate-bending domains. The quantitative comparison appears in Table~\ref{tab:oc_baseline} and the corresponding density fields in Fig.~\ref{app:fig:oc_baseline}.
The most severe case is $D_{\text{Circle}}$, where \textit{OC} compliance oscillates between $C = 10.639$ and $C = 8.393$ in the final iterations, and the volume fraction collapses to $V = 0.13$ (target $V^* = 0.20$), resulting in a nearly empty density field (Fig.~\ref{app:fig:oc_baseline}b).
In $D_{\text{Hole}}$, \textit{OC} compliance ($C = 148.959$) is $2.36\times$ worse than the \textit{Adam} baseline ($C = 63.113$), with severe fragmentation visible in the density field (Fig.~\ref{app:fig:oc_baseline}c).
In $D_{\text{Rect}}$, \textit{OC} produces a recognizable but disconnected structure with compliance $1.84\times$ worse than \textit{Adam} (Fig.~\ref{app:fig:oc_baseline}a).
The \textit{Adam} baseline therefore represents a strong conventional SIMP result, justifying the choice of \textit{Adam} over \textit{OC} for the Heaviside-projected SIMP framework used throughout this study.

\begin{table}[pos=htbp]
\centering
\caption{Baseline compliance comparison: \textit{Adam} vs.\ \textit{OC} optimizer (no SDS, $V^*=0.20$).}
\label{tab:oc_baseline}
\begin{tabular}{lccc}
\toprule
Domain & \textit{Adam} $C$ & \textit{OC} $C$ & \textit{OC} / \textit{Adam} \\
\midrule
$D_{\text{Rect}}$ & 9.106 & 16.736 & $1.84\times$ \\
$D_{\text{Circle}}$ & 3.308 & 10.639 & $3.22\times$ \\
$D_{\text{Hole}}$ & 63.113 & 148.959 & $2.36\times$ \\
\bottomrule
\end{tabular}
\end{table}

\section{Post-processing details: density field to CAD geometry}
\label{app:postprocessing}

\begin{figure}[pos=htbp]
\centering
\includegraphics[width=\textwidth]{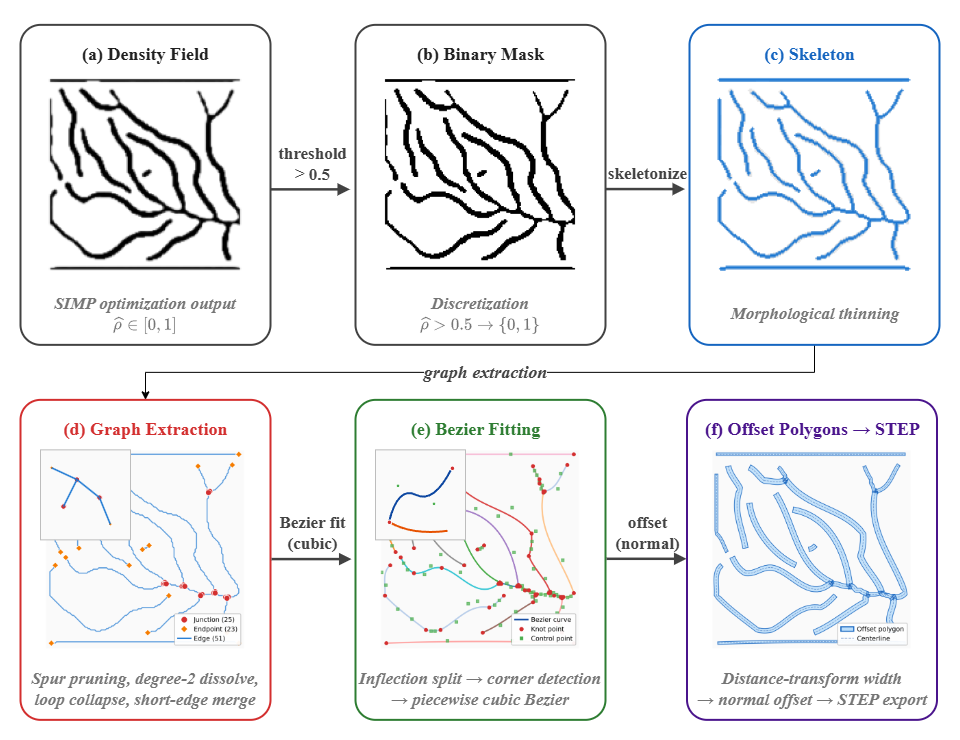}
\caption{Post-processing pipeline: density field $\rightarrow$ binarization $\rightarrow$ skeletonization $\rightarrow$ graph $\rightarrow$ B\'{e}zier fitting $\rightarrow$ CAD export.}
\label{app:fig:postprocessing}
\end{figure}

The optimized density field $\hat{\boldsymbol{\rho}}$ is a continuous 2D image that cannot be directly manufactured.
A six-stage pipeline is presented to automatically convert it into a \rev{CAD-convertible candidate} geometry (STEP format).
The complete pipeline is illustrated in Fig.~\ref{app:fig:postprocessing}.
This optimization-to-CAD/CAE (computer-aided engineering) integration shares a similar objective to prior work \cite{yoo2021integrating} that automated a 3D conceptual wheel design-evaluation pipeline in deep-learning-based generative design.
However, the present study adopts a skeleton-based approach tailored to density-field topology-optimization results.
The entire pipeline completes in 5 seconds on a single CPU core and requires no user parameters beyond the extrusion height.

\paragraph{Step 1: Binarization}
The continuous density field is converted into a binary mask $M$ using a threshold of $\rho_\text{th} = 0.5$:
\begin{equation*}
M(i,j) = \begin{cases} 1 & \text{if } \hat{\rho}(i,j) > 0.5 \\ 0 & \text{otherwise} \end{cases}
\end{equation*}
Since the Heaviside projection ($\beta = 64$) already produces a nearly binary distribution (gray density < 3\%, as shown in Appendix~\ref{app:gray}), additional morphological post-processing (closing, opening) is unnecessary.

\paragraph{Step 2: Skeletonization and Distance Transform}
Two operations are performed in parallel on the binary mask. First, Zhang--Suen thinning \cite{zhang1984fast} extracts the one-pixel-wide medial axis (skeleton) of the foreground region.
This skeleton captures the topological structure of the rib network, including branching patterns and connectivity.
Second, a Euclidean distance transform computes the distance from each foreground pixel to the nearest background pixel.
The distance value at skeleton pixels corresponds to the local half-thickness of the respective rib, which is used in \textit{Step 5} to assign physically meaningful rib widths.

\paragraph{Step 3: Graph Extraction}
The skeleton image is converted into a topological graph $G = (V, E)$.
Each skeleton pixel is classified by its degree in the 8-connected neighborhood:
pixels with degree $\geq 3$ are registered as branch point nodes, and pixels with degree $= 1$ as endpoint nodes.
Skeleton paths between nodes are traced via depth-first search to construct edges $E$, where each edge is represented as $(v_a, v_b, \mathbf{P})$ with $\mathbf{P} \in \mathbb{R}^{n_e \times 2}$ being the pixel coordinate sequence of the skeleton path.

\paragraph{Step 4: B\'{e}zier Curve Fitting}
Each edge path $\mathbf{P}$ is approximated by a piecewise cubic B\'{e}zier curve.
First, points where the local direction changes by more than $30^\circ$ within an adaptive window are detected as corners, thereby splitting the path into sub-segments.
For each segment, the two endpoints are fixed, and the internal control points $\mathbf{P}_1$, $\mathbf{P}_2$ are determined via least-squares fitting based on chord-length parameterization.
\begin{equation*}
\mathbf{B}(\xi) = (1-\xi)^3 \mathbf{P}_0 + 3(1-\xi)^2 \xi\, \mathbf{P}_1 + 3(1-\xi)\xi^2\, \mathbf{P}_2 + \xi^3 \mathbf{P}_3
\end{equation*}
All control points are converted from pixel coordinates to physical domain coordinates $(x, y)$.

\paragraph{Step 5: Offset Polygon Generation}
The B\'{e}zier centerline of each edge is expanded into a closed 2D polygon with physically meaningful thickness.
The distance transform values are sampled along the skeleton path of each edge, and their mean is taken as the half-width $w_{\text{half}}$ of the edge (in pixel units).
This is converted to domain coordinates using the average pixel-to-domain scale factor $s = (W/n_\text{elx} + H/n_\text{ely})/2$.
A minimum half-width of 0.3 is enforced to prevent degenerate polygons.
Each B\'{e}zier segment is evaluated at 20 equally spaced parameter values to obtain centerline points and tangent vectors.
At each point, the unit normal $\mathbf{n} = (-\dot{y}, \dot{x})/|\dot{\mathbf{B}}|$ is computed, and left/right offset curves are generated:
\begin{equation*}
\mathbf{C}_{\text{left}} = \mathbf{B}(\xi) + w_{\text{half}} \cdot \mathbf{n}(\xi), \quad \mathbf{C}_{\text{right}} = \mathbf{B}(\xi) - w_{\text{half}} \cdot \mathbf{n}(\xi)
\end{equation*}
The left curve and the reversed right curve are connected to form a closed polygon.

\paragraph{Step 6: CAD Export}
The 2D polygons are extruded into 3D solids to ensure CAD compatibility.
Each polygon is converted into a closed wire, a planar face is constructed, and the face is extruded by a specified height $h$ (default 5 mm) in the $z$-direction.
All solids are exported as an AP214 schema (ISO 10303-214) STEP data file.
Optionally, a base plate with a configurable thickness can be added beneath the rib structure.

\section{Additional results}
\label{app:additional}

\subsection{Alternative baselines: numerical summary}
\label{app:baselines}

Best-of-run compliance values underpinning the comparison in Section~\ref{sec:baselines} appear in Table~\ref{app:tab:baseline_summary}.

\begin{table}[pos=htbp]
\centering
\caption{Best-of-run compliance $C$ for the proposed SDS framework against three SIMP-only alternatives (lower is better). All methods share the same SIMP solver and therefore require no additional training. \rev{The multi-start and perturbation entries are the best of 50 runs per domain; a matched-budget distributional comparison is given in Fig.~\ref{app:fig:equalbudget}.} $^{\dagger}$Best SDS run on $D_{\text{Hole}}$ is at $\lambda_\text{sds}^0=50$ (see Appendix~\ref{app:hole_lambda}); the remaining domains use $\lambda_\text{sds}^0=10$.}
\label{app:tab:baseline_summary}
\begin{tabular}{lccc}
\toprule
Method & $D_{\text{Rect}}$ & $D_{\text{Circle}}$ & $D_{\text{Hole}}$ \\
\midrule
Baseline SIMP (\textit{Adam})    & 9.106  & 3.308 & 63.113 \\
Multi-start SIMP (best) & \rev{8.102}  & \rev{3.166} & \rev{47.434} \\
Perturbation SIMP (best)& \rev{8.339}  & \rev{3.117} & \rev{47.495} \\
Image-guided (best)     & 9.217  & 3.105 & 49.505 \\
\textbf{SDS (best)}     & \textbf{7.886} & \textbf{2.522} & \textbf{43.20}$^{\dagger}$ \\
\bottomrule
\end{tabular}
\end{table}

\begin{figure}[pos=htbp]
\centering
\rev{\subfloat[$D_{\text{Rect}}$]{\includegraphics[width=0.32\textwidth]{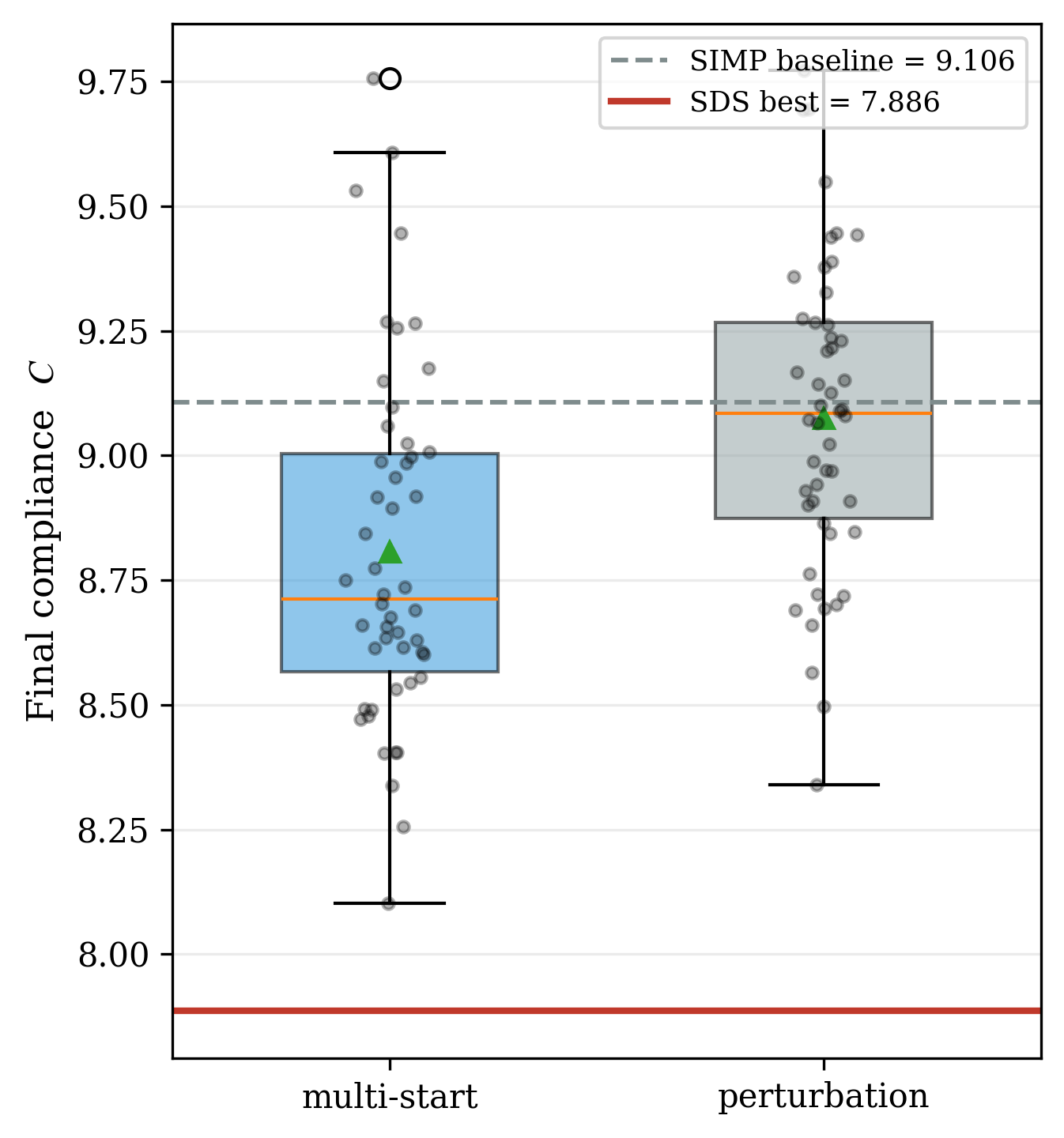}}\hfill
\subfloat[$D_{\text{Circle}}$]{\includegraphics[width=0.32\textwidth]{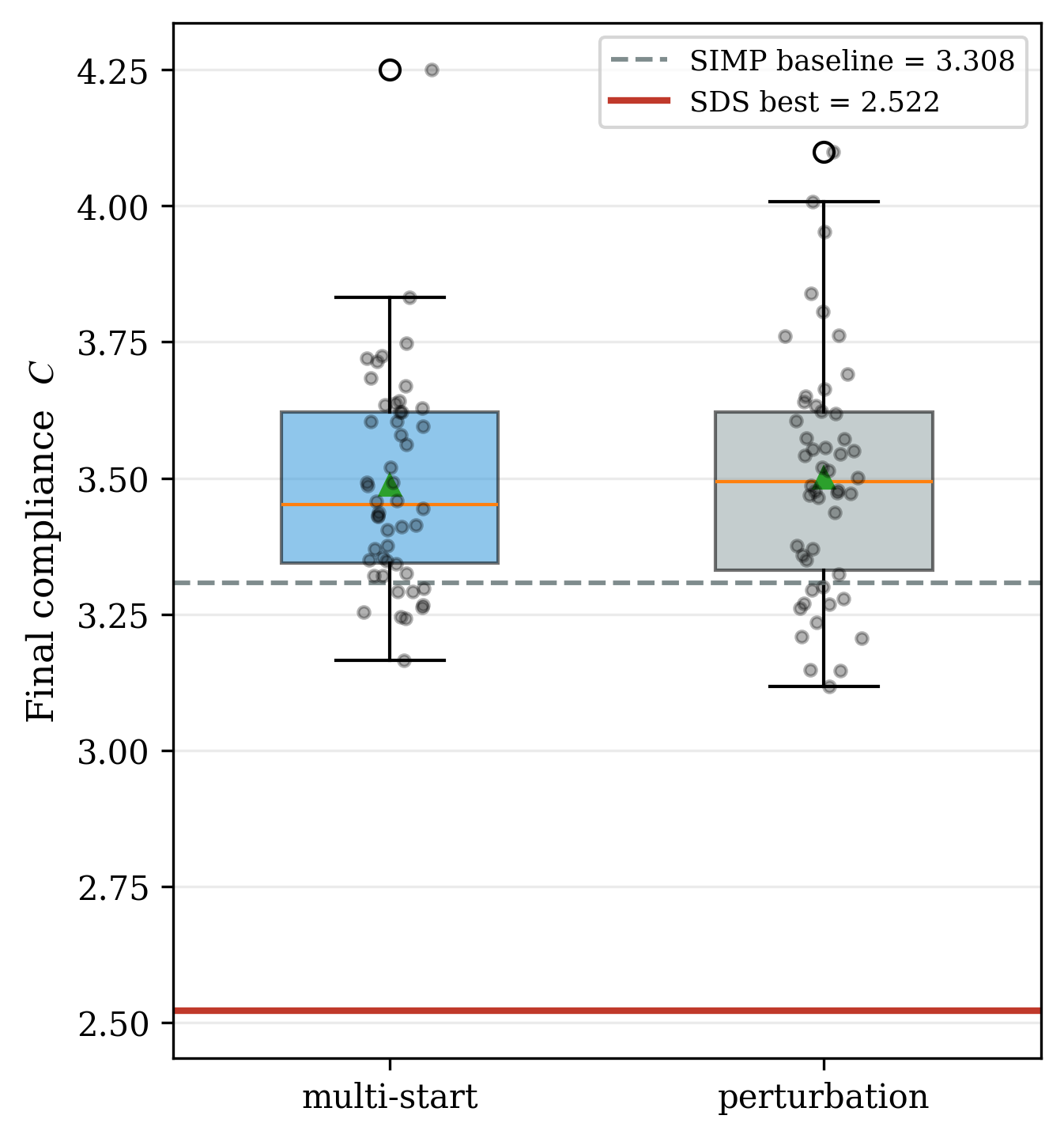}}\hfill
\subfloat[$D_{\text{Hole}}$]{\includegraphics[width=0.32\textwidth]{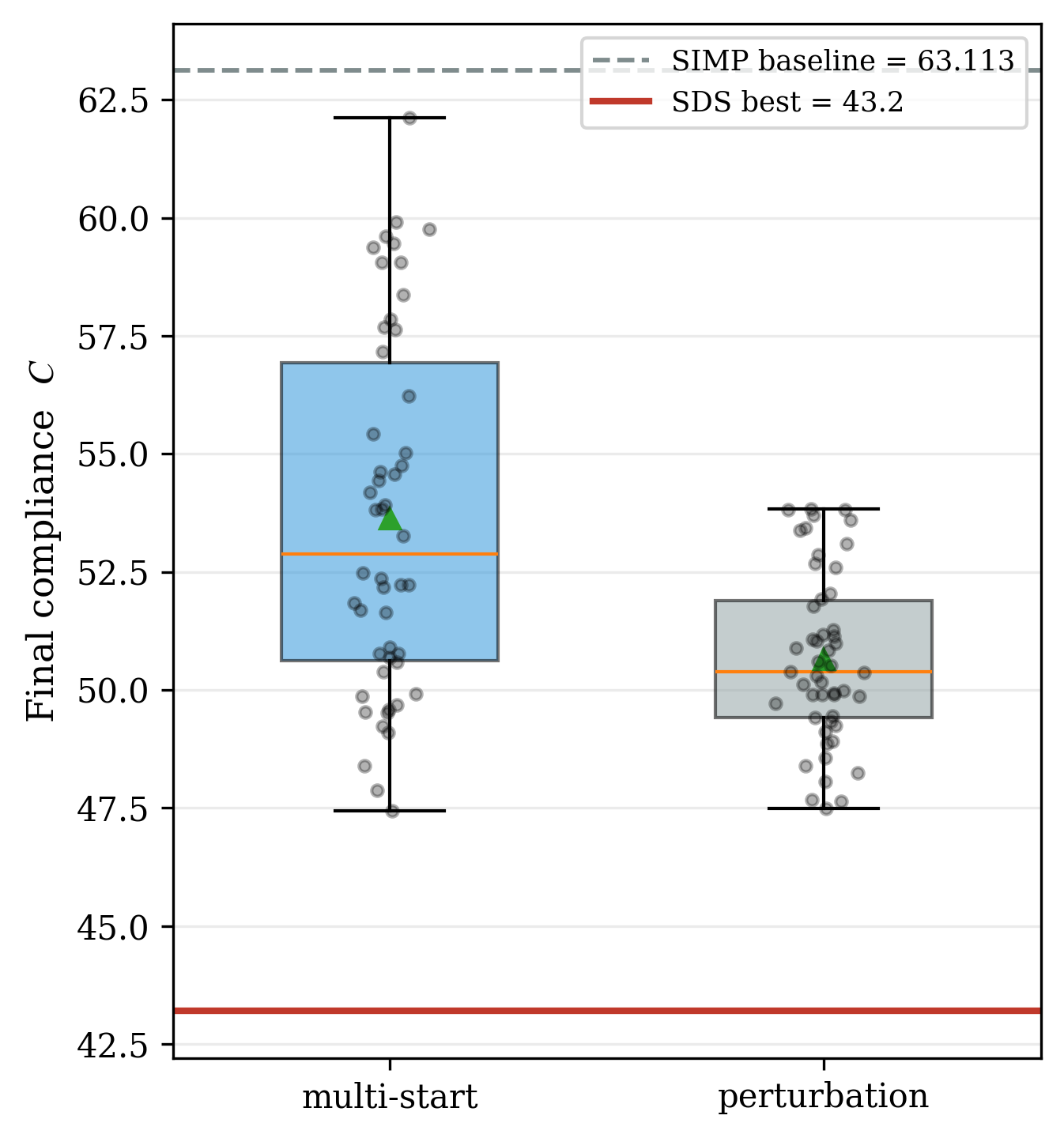}}}
\caption{\rev{Matched-number-of-runs comparison of the SIMP-only baselines against SDS across the three plate-bending domains. Box plots show the final compliance of 50 multi-start and 50 perturbation runs per domain (individual runs jittered; green triangle = mean); the dashed line is the SIMP baseline and the solid red line the best SDS compliance. In every domain the best SDS result lies below the entire baseline distribution, so neither multi-start nor perturbation reaches it under a matched budget.}}
\label{app:fig:equalbudget}
\end{figure}

\rev{A best-of-run comparison alone could be confounded by the number of restarts allocated to each method. To control for this, the two SIMP-only baselines were re-run under a matched budget of 50 random restarts (multi-start) and 50 periodic-perturbation runs ($\sigma=0.05$) per domain, then compared against the five SDS seeds by resampling equal-sized five-run subsets. Fig.~\ref{app:fig:equalbudget} shows the resulting distributions. In none of $5000$ resampled five-run subsets, for either baseline in any of the three domains, does the best baseline compliance reach the best SDS compliance. Indeed, even the best of all 50 runs per baseline falls short of the best SDS result in every domain, so no resampled five-run subset reaches it (0 of 5000). The baseline distributions lie at or above the SDS result, and in $D_{\text{Circle}}$ both baselines remain near the SIMP baseline compliance. The SDS improvement is therefore not reproducible by additional unstructured search under a matched number-of-runs budget. Wall-clock costs are reported separately because SDS and SIMP use different CPU/GPU resources.}

\begin{figure}[pos=htbp]
\centering
\subfloat[$D_{\text{Rect}}$]{\includegraphics[width=0.32\textwidth]{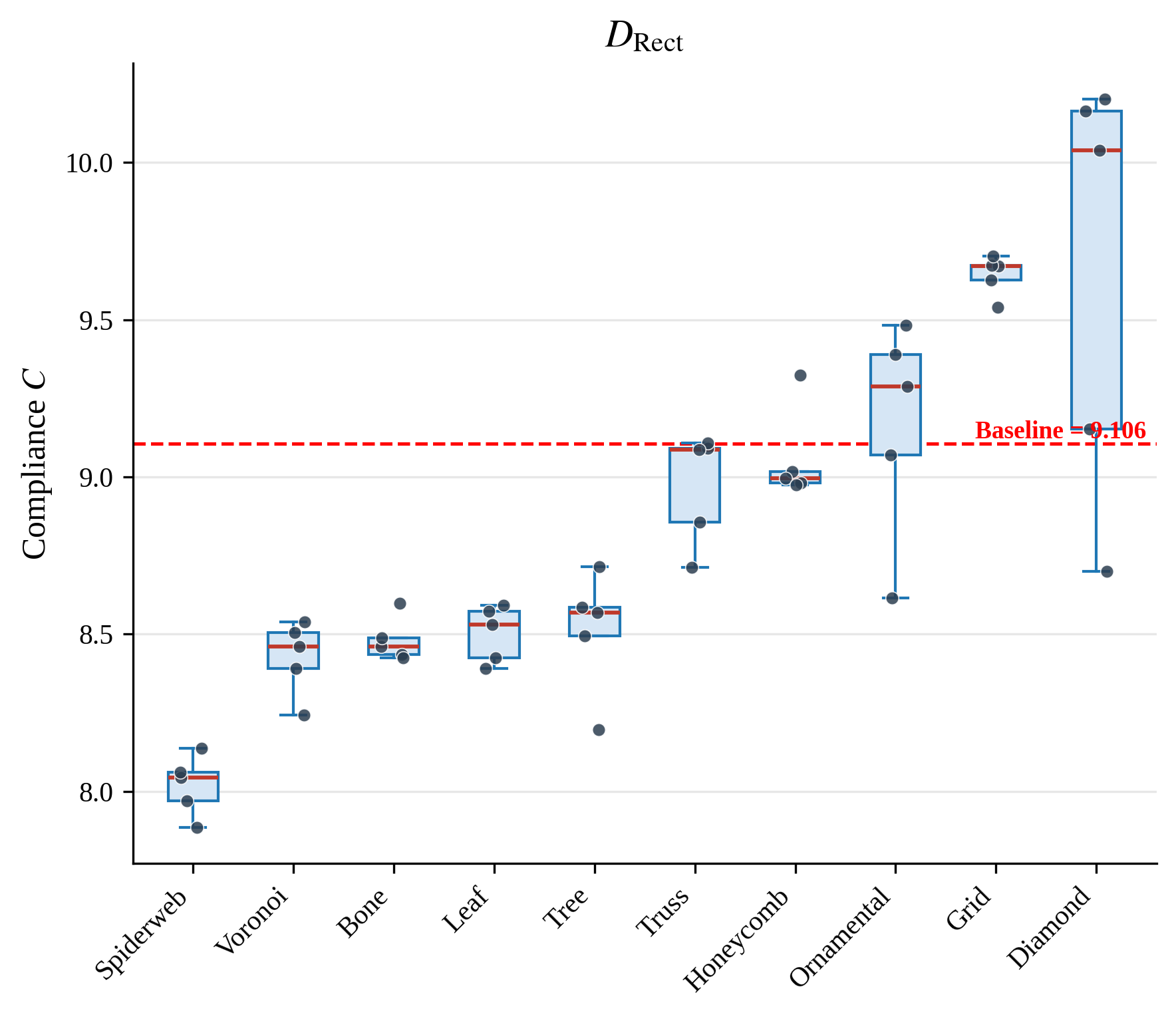}}\hfill
\subfloat[$D_{\text{Circle}}$]{\includegraphics[width=0.32\textwidth]{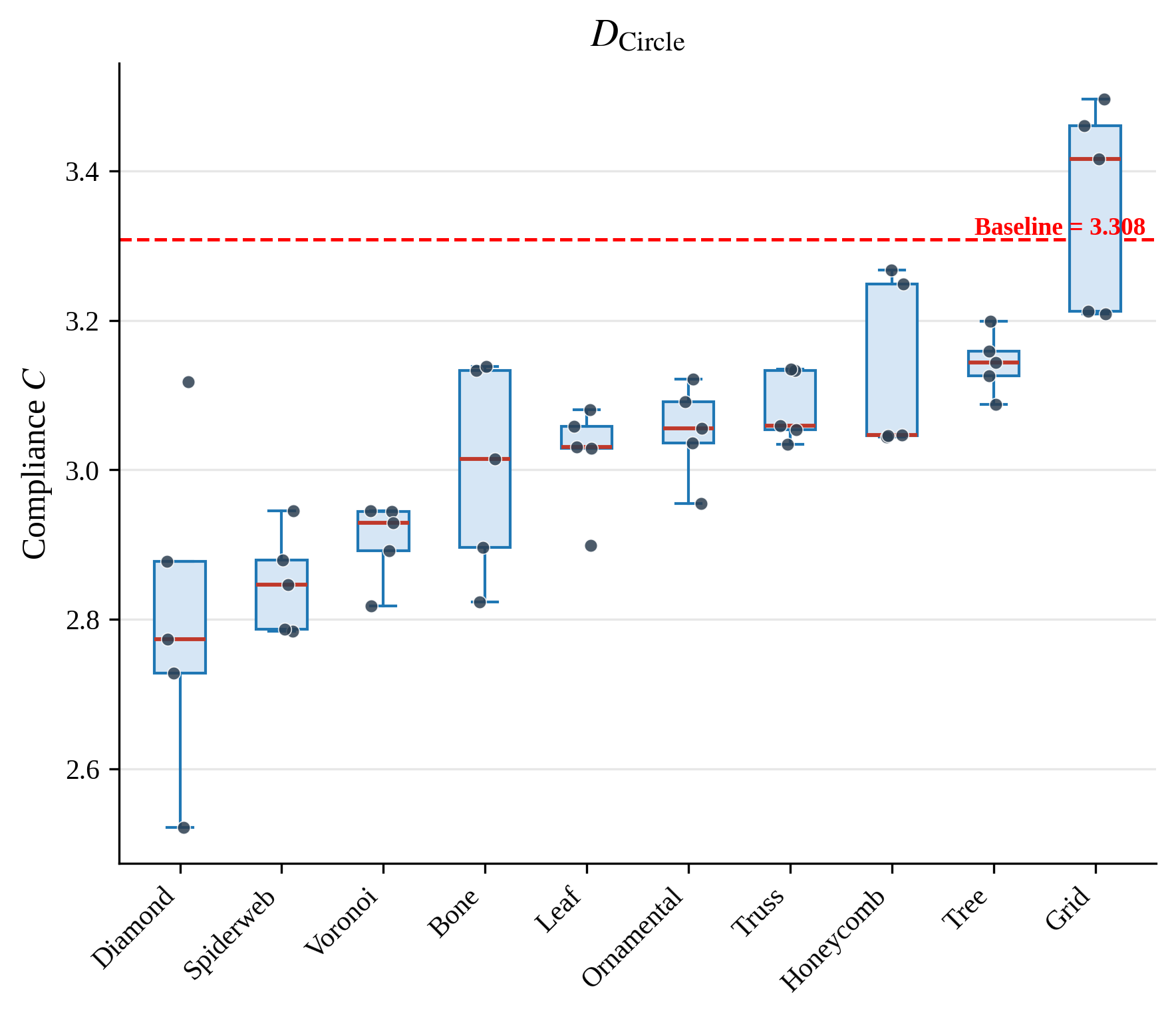}}\hfill
\subfloat[$D_{\text{Hole}}$]{\includegraphics[width=0.32\textwidth]{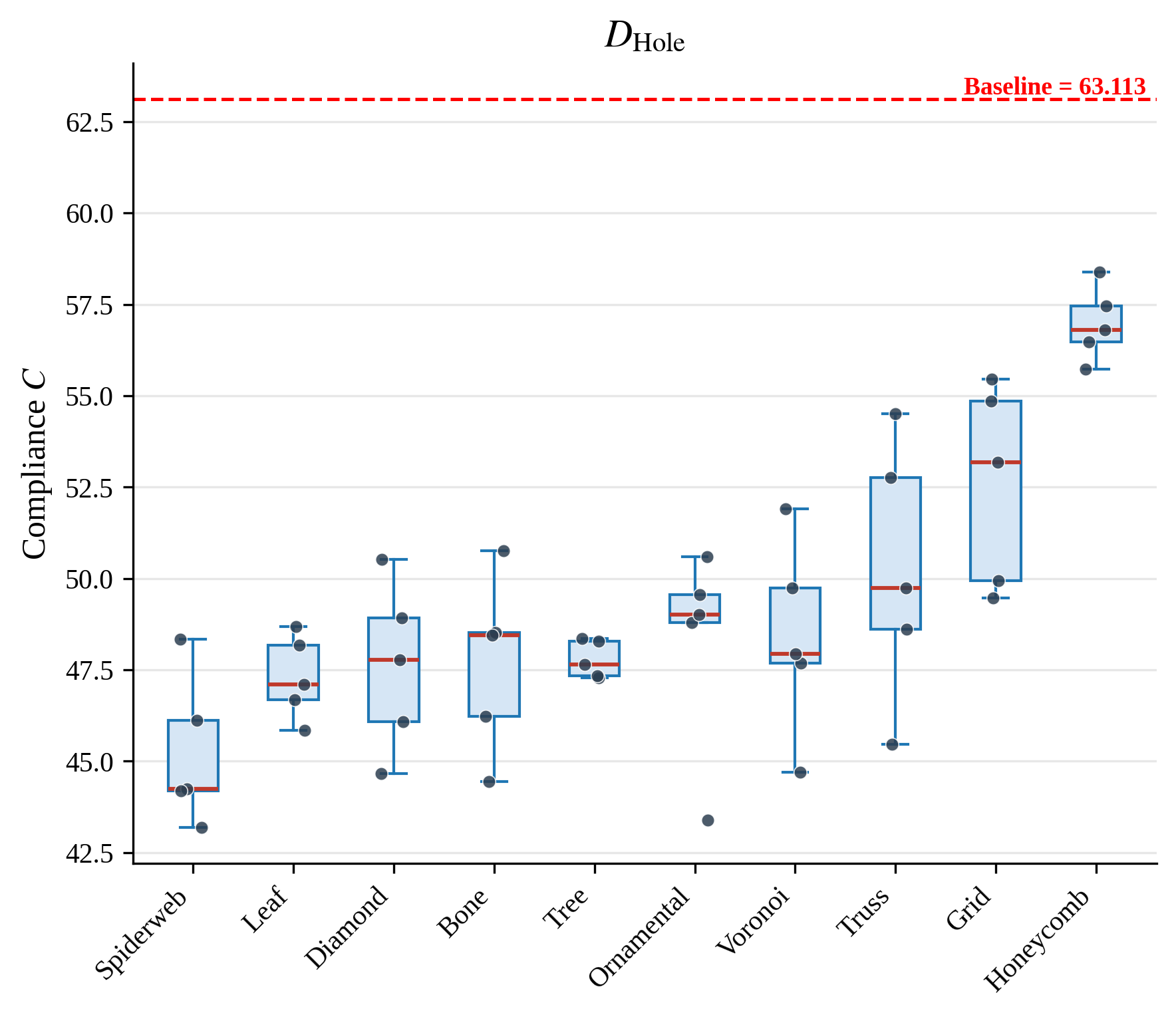}}
\caption{\rev{Seed-level compliance for the three mechanical plate-bending domains, all ten prompts, five seeds each (dots: individual seeds; box: interquartile range; red line: median). Prompts are ordered by mean compliance; the dashed line is the SIMP-only baseline.}}
\label{app:fig:seed_mech}
\end{figure}

\begin{figure}[pos=htbp]
\centering
\includegraphics[width=0.58\linewidth]{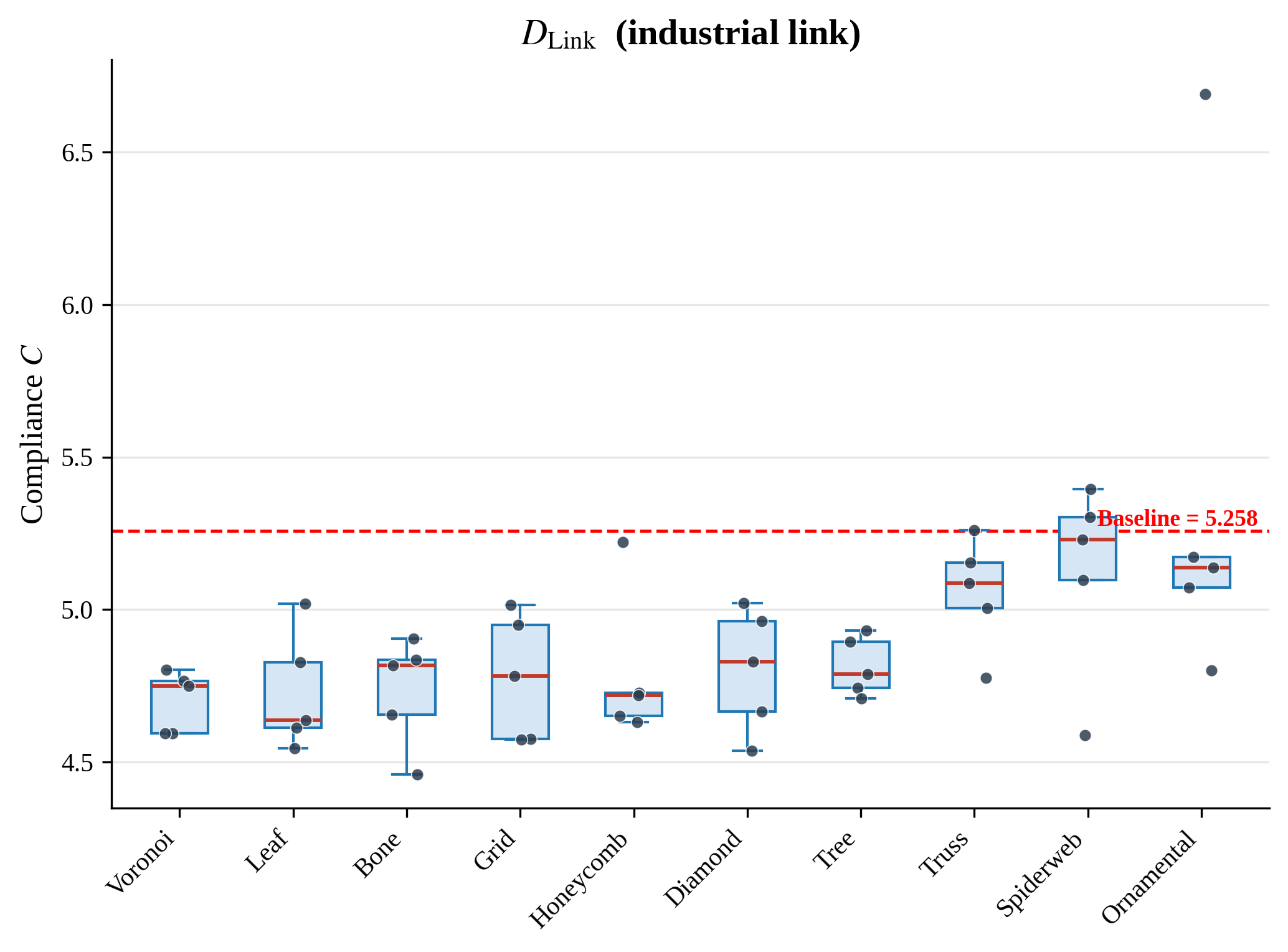}
\caption{\rev{Seed-level compliance for the industrial $D_{\text{Link}}$ domain (ten prompts, five seeds each; baseline $C=5.258$ dashed).}}
\label{app:fig:seed_link}
\end{figure}

\begin{figure}[pos=htbp]
\centering
\subfloat[$D_{\text{Rect}}$]{\includegraphics[width=0.32\textwidth]{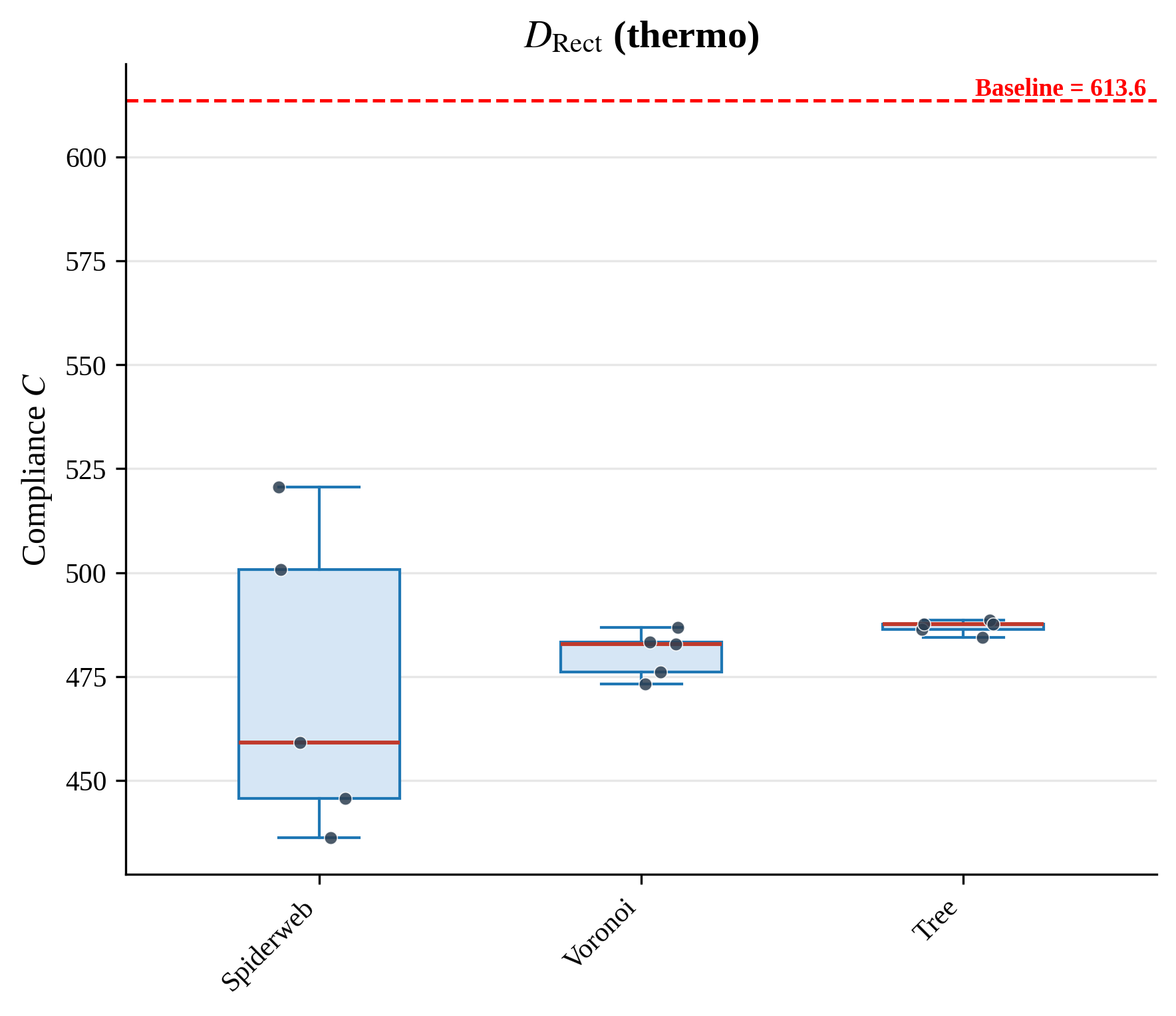}}\hfill
\subfloat[$D_{\text{Circle}}$]{\includegraphics[width=0.32\textwidth]{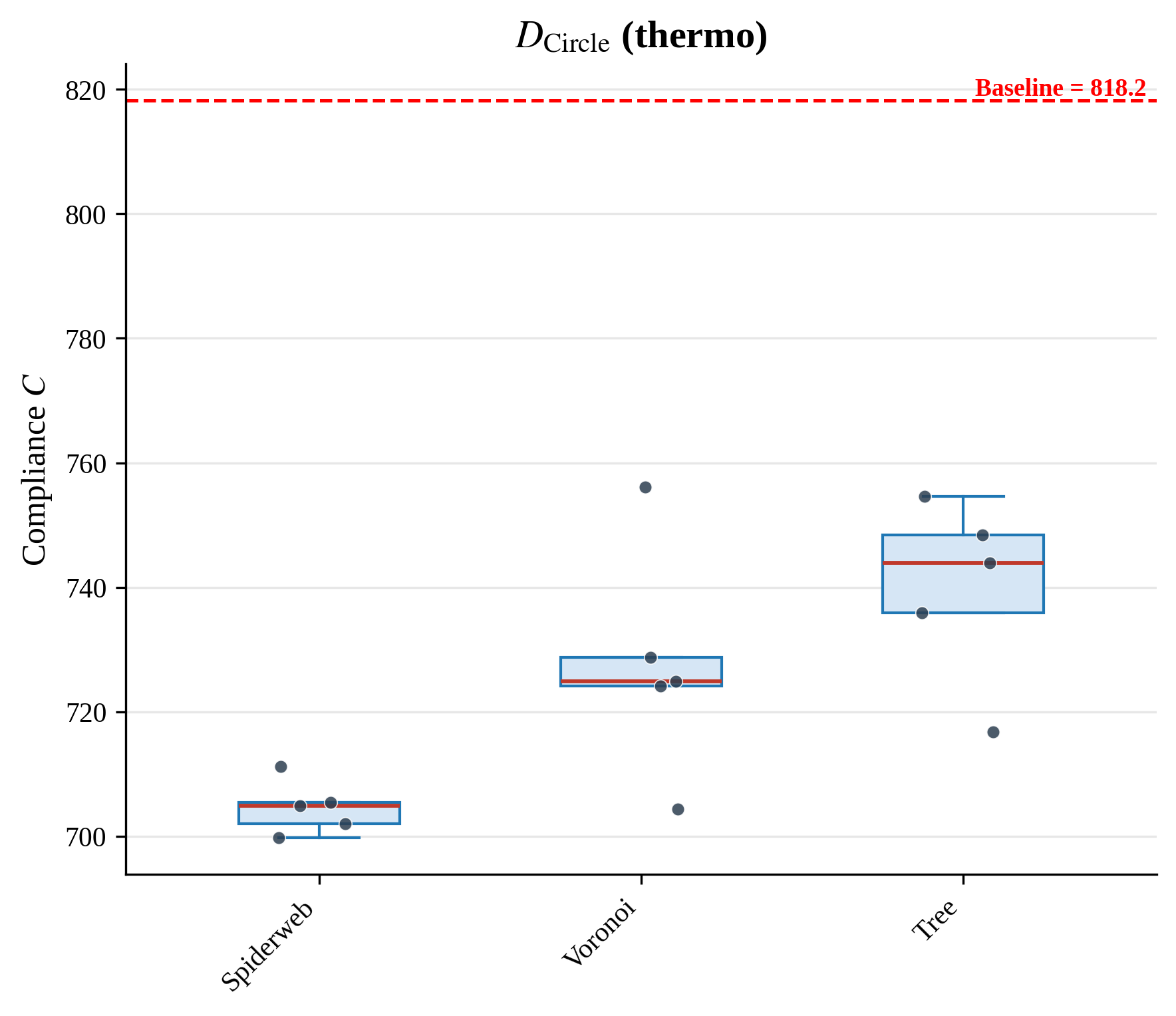}}\hfill
\subfloat[$D_{\text{Hole}}$]{\includegraphics[width=0.32\textwidth]{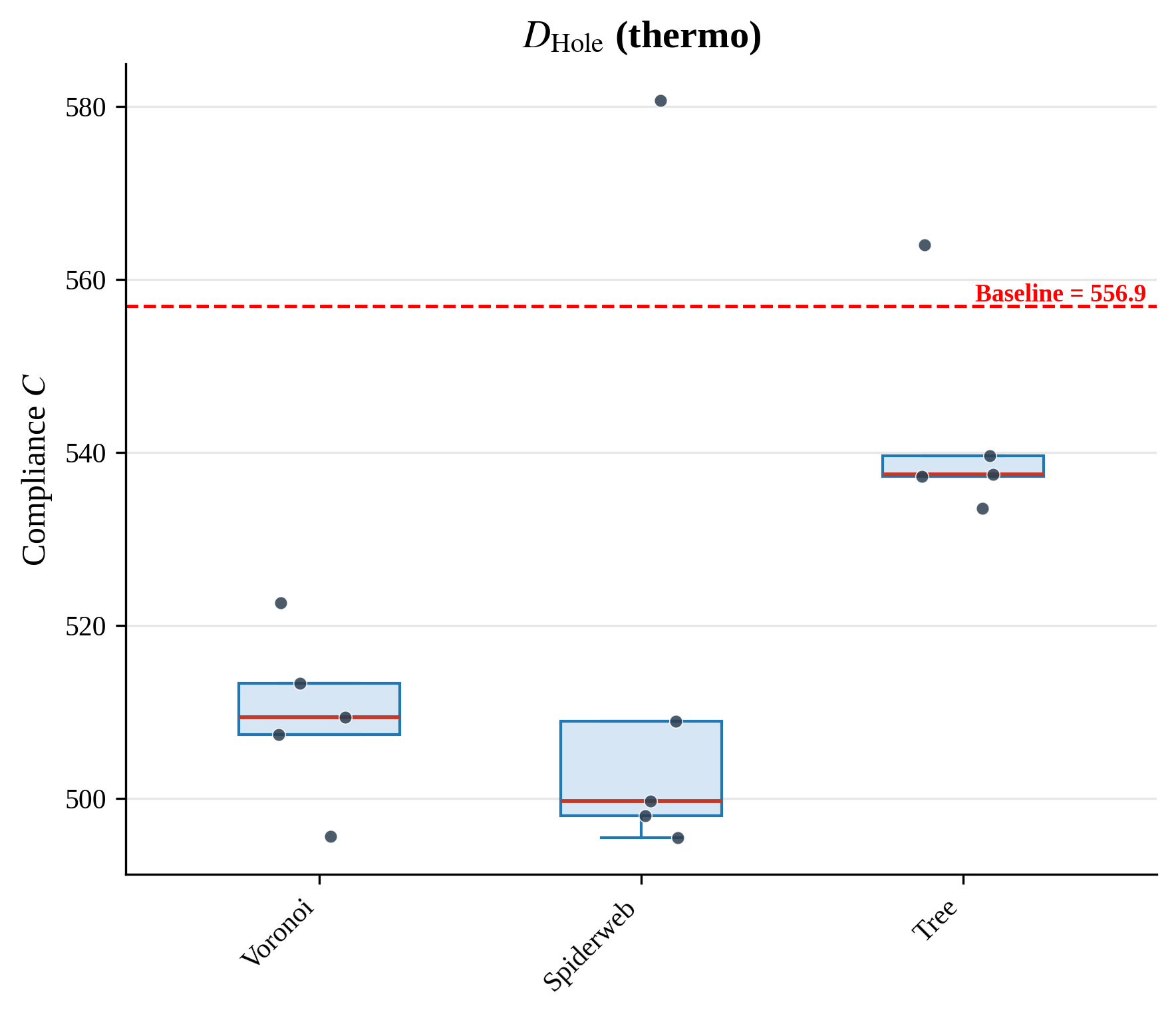}}
\caption{\rev{Seed-level thermoelastic compliance for the three plate-bending domains (three prompts each, five seeds); every prompt improves on the baseline (dashed).}}
\label{app:fig:seed_thermo}
\end{figure}

\rev{For full transparency of the per-seed results underlying the mean$\pm$std values in Table~\ref{tab:summary}, Figs.~\ref{app:fig:seed_mech}, \ref{app:fig:seed_link}, and \ref{app:fig:seed_thermo} show the compliance of every individual seed (five per prompt--domain combination) as strip-and-box plots, with the SIMP-only baseline marked. Prompt and hyperparameter selection was carried out on separate single-seed pilot runs before this five-seed statistical evaluation, keeping model selection distinct from final testing.}

\subsection{$D_{\text{Hole}}$ SDS weight ($\lambda_\text{sds}^0 = 50$)}
\label{app:hole_lambda}

$D_{\text{Hole}}$ uses $\lambda_\text{sds}^0=50$ rather than the $\lambda_\text{sds}^0=10$ adopted for $D_{\text{Rect}}$, $D_{\text{Circle}}$, and $D_{\text{Link}}$.
Because the baseline compliance ($C=63.113$) is roughly seven times that of $D_{\text{Rect}}$, the physics-gradient magnitude is correspondingly larger. Under the raw-weighted-sum update, $\lambda_\text{sds}^0$ must rise to keep the generative signal competitive.
The scaling rationale is identical to the thermoelastic case (Appendix~\ref{app:thermo_lambda}).
Full per-prompt results at $\lambda_\text{sds}^0=50$ appear in Table~\ref{tab:summary}. Domain-level interpretation is developed in Section~\ref{sec:main_results}.

\subsection{Gray density: quantitative ablation}
\label{app:gray}

This section reports the quantitative ablation of Heaviside projection and $\beta$-continuation. The mechanistic origin of the SDS-induced gray-density tendency is discussed in Section~\ref{sec:gray_density}.

Without Heaviside projection, SDS guidance yields a gray-density ratio of $42.6\%$ against $20\text{--}21\%$ for pure SIMP, nearly doubling the baseline.
$\beta$ doubles every 50 iterations following $\beta_k = \min(2^{\lfloor k/50 \rfloor}, 64)$.
Fig.~\ref{fig:gray} and Table~\ref{app:tab:heaviside} report the gray-density/compliance trade-off for $\beta_\text{max}\in\{32,64\}$ against the projection-off baseline. Higher $\beta_\text{max}$ drives the gray ratio below $3\%$ with no compliance penalty (Table~\ref{app:tab:heaviside}: $C=9.29$ at $\beta_\text{max}=64$ versus $9.93$ at $\beta_\text{max}=32$), which justifies the choice $\beta_\text{max}=64$.

\begin{table}[pos=htbp]
\centering
\caption{Effect of Heaviside projection and $\beta$-continuation on compliance and gray density.}
\label{app:tab:heaviside}
\begin{tabular}{lcccc}
\toprule
Experiment & Heaviside & $\beta_\text{max}$ & Compliance & Gray ratio (\%) \\
\midrule
A & OFF & --- & 18.66 & 42.6 \\
B & ON & 64 & 9.29 & 2.5 \\
C & ON & 32 & 9.93 & 5.1 \\
\bottomrule
\end{tabular}
\end{table}

\subsection{When SDS fails: over-constraining}
\label{app:fails}

Conversely, there are cases where SDS degrades performance.
Two representative failure modes are observed in $D_{\text{Rect}}$.
The \textit{Diamond} prompt ($+6.0\%$) over-constrains the density field into a rigid rhombus lattice that conflicts with the natural diagonal load paths from center to corners.
Similarly, the \textit{Grid} prompt ($+5.9\%$) forces an orthogonal lattice that wastes material in regions of low stress concentration.
In both cases, the SDS gradient consistently opposes the physics gradient: while the physics gradient seeks to concentrate material along load paths, the SDS gradient redistributes it to maintain the prompted pattern, leading to a compromise that is optimal for neither.

The cosine cooldown mitigates this conflict by gradually reducing SDS influence.
However, without cooldown, such conflicts persist until the final convergence, as confirmed in the scheduling ablation (Section~\ref{sec:ablation}): removing the timestep annealing alone causes $+13.0\%$ degradation even for the well-aligned \textit{Spiderweb} prompt.
For misaligned prompts, the degradation would be more severe.
These observations suggest a practical guideline: when the prompted pattern is uncertain in its compatibility with the load paths, a shorter cooldown or lower $\lambda_\text{sds}$ can serve as a conservative safeguard.

\subsection{\rev{Prompt ablation: isolating the semantic contribution}}
\label{app:ablation}
\rev{The contribution of the prompt's \emph{semantic content} is isolated by varying the prompt while holding the solver and the SDS coupling fixed. On $D_{\text{Rect}}$, the five control conditions share the same five random seeds ($\lambda_\text{sds}^0=10$), so they are compared under identical initialization and noise realizations (a matched-seed comparison); the \textit{Spiderweb} reference is the main-experiment value ($-11.9\%$, Table~\ref{tab:summary}). The analysis relies only on the relative ranking across the six conditions. The best structural prompt (\textit{Spiderweb}) is placed alongside five controls: an \textit{empty} prompt (unconditional SDS), a \textit{prefix-only} prompt (``bold black lines on white background,'' with the structural noun removed), two semantically irrelevant prompts (\textit{fruit}, \textit{sunset}), and a \textit{dot} prompt that changes a single word (``lines''$\to$``dots'') so that the prior forms isolated points rather than connected line networks. Table~\ref{app:tab:ablation} reports the final compliance, win rate, and mean skeleton endpoint count $n_\text{end}$. The corresponding rib morphologies and the $n_\text{end}$--compliance correlation appear in Fig.~\ref{fig:prompt_ablation} (Section~\ref{sec:dead_end}).}

\rev{Three findings emerge. First, the structural prompt ($-11.9\%$) far outperforms the empty and prefix-only controls ($-3.6\%$ and $-4.5\%$), and the improvement tracks the endpoint count that structural prompts minimize rather than the mere presence of a generative term. The semantically irrelevant \textit{fruit} prompt still helps ($-5.5\%$) by yielding a relatively low-endpoint morphology. Second, the \textit{dot} prompt degrades compliance the most ($+7.3\%$, 0/5 seeds winning), raising the endpoint count to the highest of any condition ($n_\text{end}=49$ versus $19$ for \textit{Spiderweb}). The semantically irrelevant \textit{sunset} prompt, with the next-highest endpoint count ($n_\text{end}=36$), is the only other condition to worsen the baseline ($+3.6\%$), so the two most endpoint-rich prompts are precisely the two that degrade compliance. Third, across the thirty individual runs (five seeds per prompt) the endpoint count and the compliance change are strongly correlated (Pearson $r=0.87$). This controlled variation is consistent with a causal role for the endpoint count in the dead-end suppression mechanism of Section~\ref{sec:dead_end}. The within-design surgery of Appendix~\ref{app:surgery}, which manipulates the endpoints without altering the surrounding morphology, provides the direct intervention-based support.}

\begin{table}[pos=htbp]
\centering
\caption{\rev{Prompt ablation on $D_{\text{Rect}}$ (mean $\pm$ std over five seeds; baseline $C=9.106$). Conditions are ordered by mean skeleton endpoint count $n_\text{end}$. ``Wins'' counts seeds improving over the baseline. The two conditions with the highest $n_\text{end}$ (\textit{dot}, \textit{sunset}) degrade compliance, the \textit{dot} prompt most.}}
\label{app:tab:ablation}
\begin{tabular}{lcccc}
\toprule
\rev{Prompt condition} & \rev{$C$} & \rev{$\Delta\%$} & \rev{Wins} & \rev{$n_\text{end}$} \\
\midrule
\rev{Semantic (\textit{Spiderweb})}   & \rev{$\mathbf{8.02\pm0.10}$} & \rev{$\mathbf{-11.9\%}$} & \rev{5/5} & \rev{19.2} \\
\rev{Random (\textit{fruit})}         & \rev{$8.61\pm0.31$} & \rev{$-5.5\%$} & \rev{5/5} & \rev{22.6} \\
\rev{Empty (unconditional)}           & \rev{$8.77\pm0.23$} & \rev{$-3.6\%$} & \rev{4/5} & \rev{25.8} \\
\rev{Prefix-only}                     & \rev{$8.70\pm0.29$} & \rev{$-4.5\%$} & \rev{5/5} & \rev{26.4} \\
\rev{Random (\textit{sunset})}        & \rev{$9.43\pm0.36$} & \rev{$+3.6\%$} & \rev{2/5} & \rev{36.2} \\
\rev{Dot (isolated points)}           & \rev{$9.77\pm0.34$} & \rev{$+7.3\%$} & \rev{0/5} & \rev{49.2} \\
\bottomrule
\end{tabular}
\end{table}

\subsection{$\lambda_\text{sds}^0$ sensitivity in thermoelastic loading}
\label{app:thermo_lambda}

To empirically locate the optimal SDS weight for thermoelastic loading, the $D_{\text{Rect}}$ domain was swept across $\lambda_\text{sds}^0 \in \{10, 50, 100, 200\}$ using the \textit{Spiderweb} prompt as a consistent probe (5 seeds per cell).
The value adopted for all three prompts in the thermoelastic experiments ($\lambda_\text{sds}^0=100$) was selected based on this single-prompt sweep.
The results appear in the lower block of Table~\ref{tab:ablation}: a clear U-shape emerges, in which small $\lambda_\text{sds}^0$ provides insufficient generative guidance against the larger and more diffuse thermal physics gradient, the improvement grows monotonically up to $\lambda_\text{sds}^0 \approx 100$ (best $-23.0\%$), and $\lambda_\text{sds}^0=200$ already shows mild saturation ($-20.0\%$), consistent with excessive SDS overriding the physics gradient.
This U-shape mirrors the $\lambda_\text{sds}$ ablation observed in the mechanical setting (upper block of the same table) but is shifted roughly tenfold toward larger values, quantifying the claim that the optimal SDS weight scales with the magnitude of the physics gradient under the raw weighted sum (Section~\ref{sec:thermoelastic}).

\subsection{\rev{Mesh refinement and the filter length scale}}
\label{app:mesh}

\rev{Section~\ref{sec:formulation} notes that fixing the filter radius in physical units reduces mesh-dependent features. To examine this beyond the single $150\times150$ resolution used in the main experiments, a mesh-refinement study was carried out on $D_{\text{Hole}}$ at three resolutions, $N\in\{100,150,200\}$. The essential requirement is that the filter radius be fixed in \emph{physical} units rather than in element counts: setting $r_\text{min}=N/30$ and $r_\text{max}=N/15$ holds the minimum feature size constant under refinement, whereas a fixed element-count radius would shrink the physical feature size and reintroduce mesh-dependent fine detail. $D_{\text{Hole}}$ is used here because its four-sided distributed loading provides a well-conditioned setting for assessing mesh convergence.}

\begin{figure}[pos=htbp]
\centering
\includegraphics[width=0.72\linewidth]{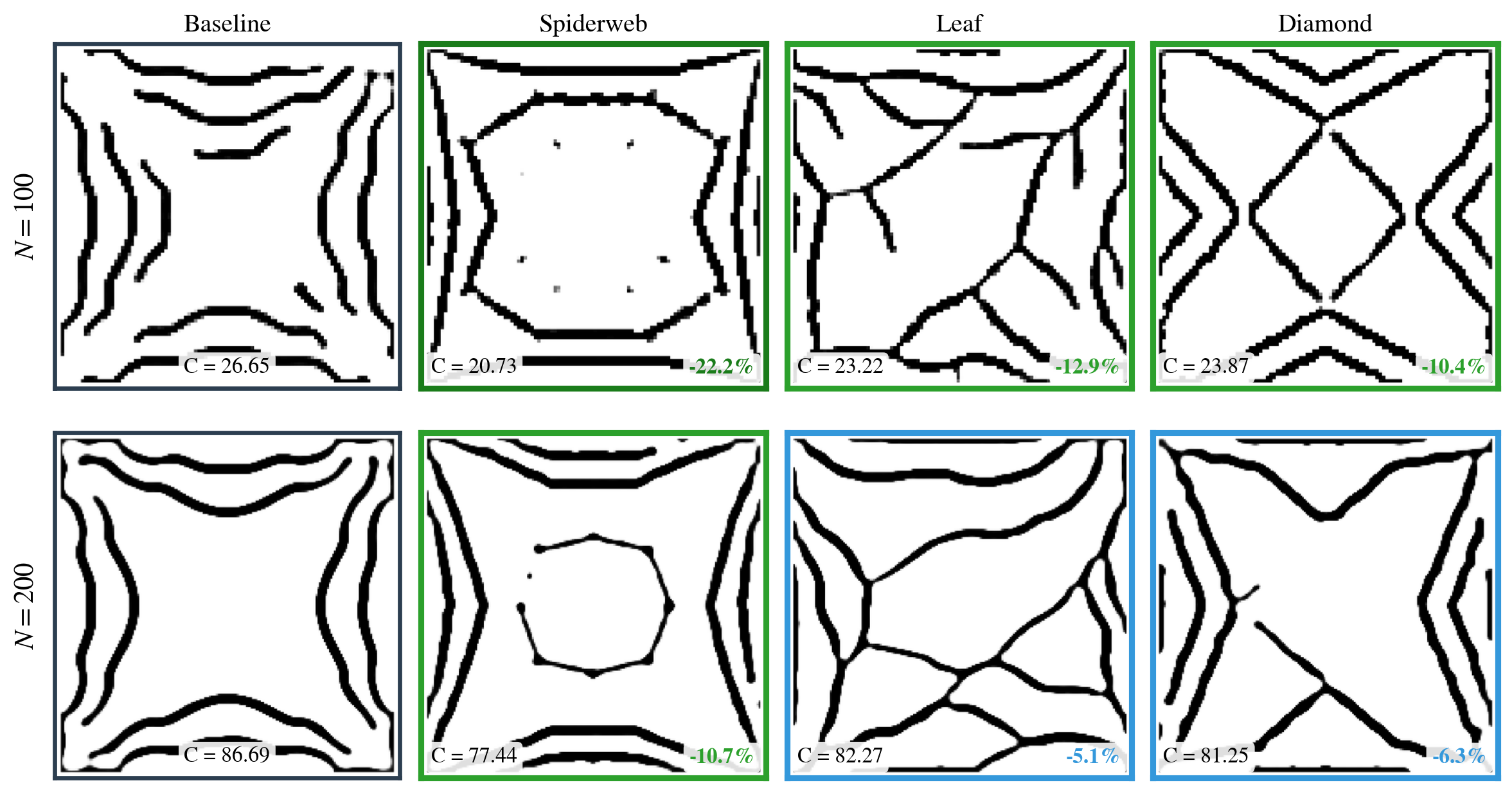}
\caption{\rev{Mesh refinement of $D_{\text{Hole}}$ with the filter radius fixed in physical units ($r_\text{min}=N/30$). Rows: $N=100$ and $N=200$ ($N=150$ appears in Fig.~\ref{fig:main_results} and Table~\ref{tab:summary}). Columns: the same-resolution SIMP baseline and three representative prompts (\textit{Spiderweb}, \textit{Leaf}, \textit{Diamond}; median seed). Cell-border color encodes the improvement over the baseline; per-panel compliance $C$ and $\Delta\%$ are overlaid.}}
\label{app:fig:mesh}
\end{figure}

\rev{Table~\ref{app:tab:mesh} and Fig.~\ref{app:fig:mesh} summarize the outcome, and two observations provide preliminary evidence of mesh-refinement robustness. First, the baseline SIMP topology is qualitatively consistent across resolutions (Fig.~\ref{app:fig:mesh}, first column): the same four curved ribs framing the central hole emerge at both resolutions, confirming that the physical-unit filter fixes the structural length scale. Second, the SDS improvement over the same-resolution baseline persists at every resolution (for the \textit{Spiderweb} probe, $-19.3\%$, $-28.4\%$, and $-10.0\%$ at $N=100,150,200$, mean over 5 seeds), so the improvement is not confined to a single discretization.}

\rev{At each resolution, $\lambda_\text{sds}^0$ was scaled with the baseline compliance following the convention adopted across domains (Section~\ref{sec:sds}). Because the density field is resized to $512\times512$ before entering the diffusion model, the SDS gradient magnitude is essentially resolution-independent, whereas the physics-gradient magnitude grows with the baseline compliance. Scaling $\lambda_\text{sds}^0$ accordingly keeps the two gradients in a comparable balance as the mesh is refined.}

\begin{table}[pos=htbp]
\centering
\caption{\rev{Mesh refinement of $D_{\text{Hole}}$ with the filter radius fixed in physical units ($r_\text{min}=N/30$, $r_\text{max}=N/15$). $C_\text{base}$ and $C_\text{SDS}$ (mean $\pm$ std over 5 seeds) are the SIMP-only and SDS-guided compliances; $\Delta\%$ is the improvement over the same-resolution baseline. The \textit{Spiderweb} prompt is used as a consistent probe.}}
\label{app:tab:mesh}
\begin{tabular}{@{}cccccc@{}}
\toprule
\rev{$N$} & \rev{$r_\text{min}$} & \rev{$C_\text{base}$} & \rev{$\lambda_\text{sds}^0$} & \rev{$C_\text{SDS}$} & \rev{$\Delta\%$} \\
\midrule
\rev{$100$} & \rev{$3.33$} & \rev{$26.65$} & \rev{$22.7$} & \rev{$21.50\pm1.45$} & \rev{$-19.3\%$} \\
\rev{$150$} & \rev{$5.00$} & \rev{$63.11$} & \rev{$50.0$} & \rev{$45.22\pm2.04$} & \rev{$-28.4\%$} \\
\rev{$200$} & \rev{$6.67$} & \rev{$86.69$} & \rev{$73.9$} & \rev{$78.01\pm3.67$} & \rev{$-10.0\%$} \\
\bottomrule
\end{tabular}
\end{table}

\subsection{\rev{3D re-analysis of the protruding-rib geometry}}
\label{app:reanalysis}

\rev{The optimization models a reinforced region as a locally stiffened Mindlin plate, whereas the exported geometry is a base plate with \emph{protruding} ribs whose bending stiffness derives from a neutral-axis offset. To verify that this 2D model preserves the design comparisons under a 3D protruding-rib solid representation reconstructed from the converged density field, each field was re-analyzed as a 3D solid: it was extruded into a base plate (thickness $1$) carrying ribs (height $5$) on the material elements and solved with 3D linear elasticity in FEniCSx under the same loading and support as the corresponding 2D problem for each domain.}

\begin{figure}[pos=htbp]
\centering
\includegraphics[width=0.55\linewidth]{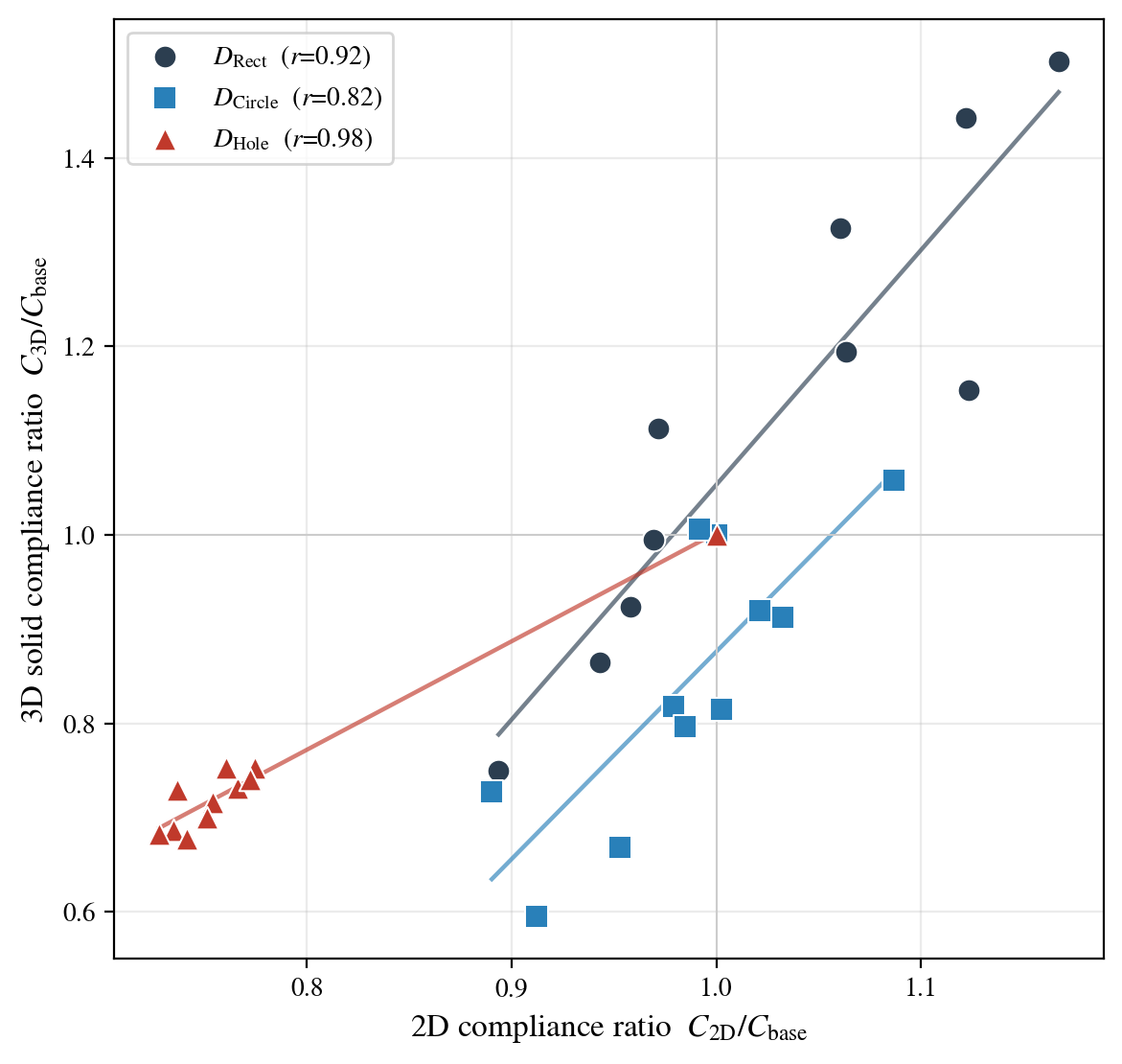}
\caption{\rev{Compliance of the 2D plate model (horizontal) versus the full 3D solid re-analysis of the protruding-rib geometry (vertical), both normalized by the SIMP-only baseline compliance, for the three plate-bending domains under matched loading and support. Points below and to the left of the baseline (ratio $=1$) improve over the baseline in both models; the positive correlation and per-domain trend lines show that the 2D design ranking is largely preserved in 3D.}}
\label{app:fig:reanalysis}
\end{figure}

\rev{Table~\ref{app:tab:reanalysis} and Fig.~\ref{app:fig:reanalysis} report the outcome across the three plate-bending domains. The compliance predicted by the 2D plate model correlates positively with that of the full 3D solid re-analysis in every domain (Pearson $r=0.92$, $0.82$, and $0.98$ for $D_{\text{Rect}}$, $D_{\text{Circle}}$, and $D_{\text{Hole}}$), and the top-ranked prompt coincides between the two models in $D_{\text{Rect}}$ (\textit{Spiderweb}). In $D_{\text{Circle}}$ and $D_{\text{Hole}}$ the best 2D prompt remains among the strongest designs in 3D. The relative ranking of prompts is thus largely preserved under the 3D solid representation, indicating that the 2D plate model retains substantial predictive value for comparative ranking, although this does not by itself eliminate the modeling discrepancy between locally thickened plates and protruding-rib solids. A full density--binary--STEP comparison including mass, volume fraction, and deformation fields would provide more comprehensive validation and remains future work.}

\begin{table}[pos=htbp]
\centering
\caption{\rev{Agreement between the 2D plate model and the full 3D solid re-analysis of the protruding-rib geometry, across the three plate-bending domains. $C_\text{base}$ is the SIMP-only baseline compliance (2D). Correlation is Pearson $r$ between the compliance values predicted by the two models over the prompt set (baseline included).}}
\label{app:tab:reanalysis}
\begin{tabular}{@{}lccc@{}}
\toprule
\rev{Domain} & \rev{$C_\text{base}$ (2D)} & \rev{$n$ prompts} & \rev{Pearson $r$} \\
\midrule
\rev{$D_{\text{Rect}}$}   & \rev{$9.106$}  & \rev{$10$} & \rev{$0.92$} \\
\rev{$D_{\text{Circle}}$} & \rev{$3.308$}  & \rev{$10$} & \rev{$0.82$} \\
\rev{$D_{\text{Hole}}$}   & \rev{$63.113$} & \rev{$10$} & \rev{$0.98$} \\
\bottomrule
\end{tabular}
\end{table}

\subsection{\rev{Within-design causal test of dead-end removal}}
\label{app:surgery}

\rev{The prompt ablation (Appendix~\ref{app:ablation}) shows that prompts inducing more skeleton endpoints yield worse compliance, and the \textit{dot} prompt establishes this by natural variation. A differentiable endpoint-count penalty is not available in the present implementation because the skeletonization that defines endpoints is non-differentiable, so a direct post-optimization intervention is used instead to isolate the mechanical role of endpoint branches. To rule out any confound from the prompt also reshaping the global morphology, on a fixed optimized design only the skeleton endpoints are altered and re-analyzed by FEA. Three interventions are applied to nine designs (three seeds in each of $D_{\text{Rect}}$, $D_{\text{Circle}}$, and $D_{\text{Hole}}$): (i) \emph{prune dead-ends}, in which every skeleton-endpoint twig, traced to its nearest branch point, is removed together with its local thickness; (ii) \emph{prune a load-bearing rib}, a control in which the single non-dead-end rib carrying the most strain energy is removed. And (iii) \emph{graft a dead-end}, in which an artificial stub is added at a branch point toward the void. Compliance is evaluated with the same plate FEA and boundary conditions used throughout.}

\begin{figure}[pos=htbp]
\centering
\includegraphics[width=0.8\linewidth]{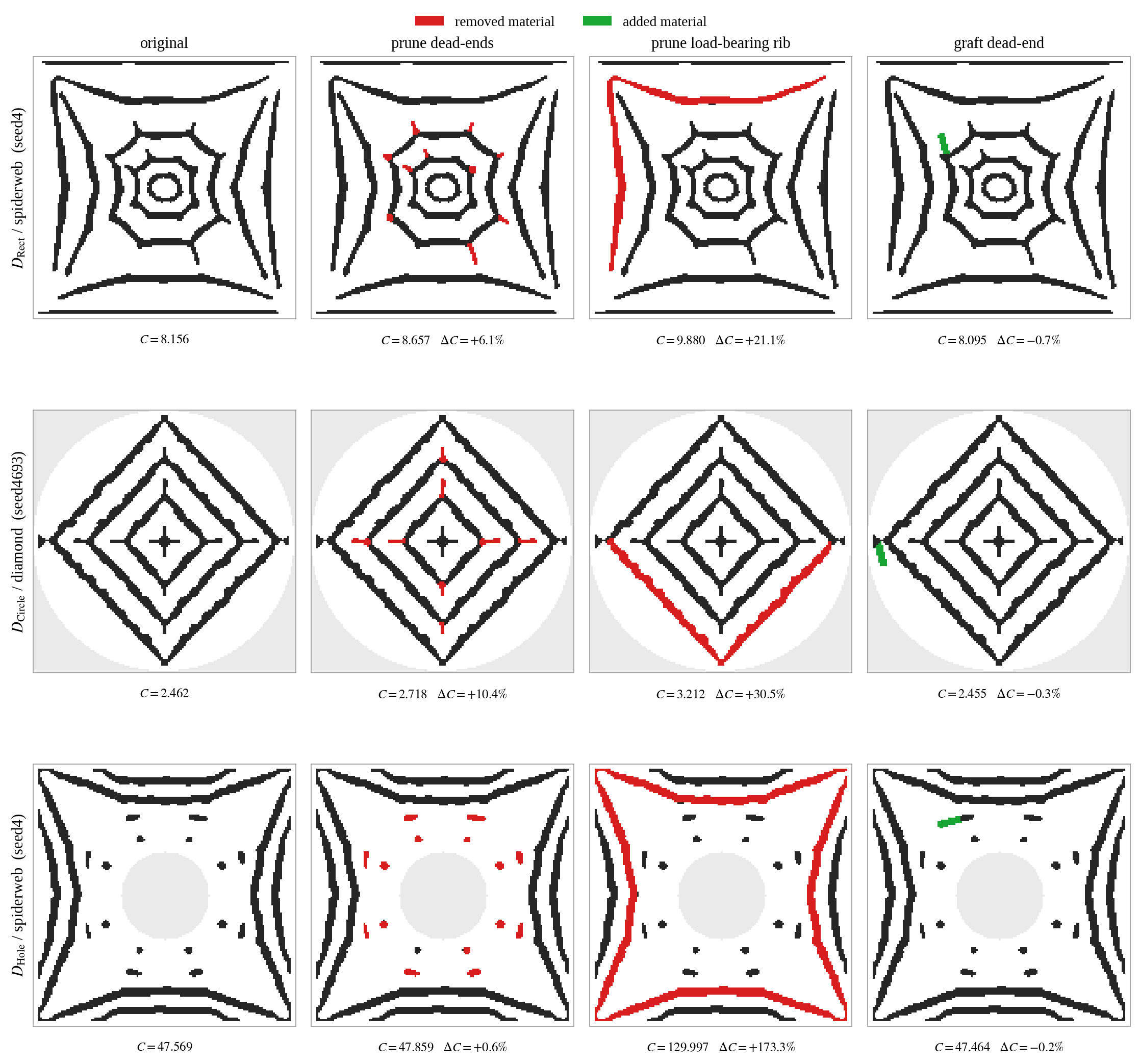}
\caption{\rev{Within-design causal test of dead-end (skeleton-endpoint) removal, one representative design per domain. Columns: original; dead-ends pruned (removed material in red); a single load-bearing rib pruned (red); an artificial dead-end grafted (added material in green). Each panel is labeled with its compliance $C$ and change $\Delta C$. Removing dead ends barely changes $C$ and grafting one is inert, whereas removing a load-bearing rib is catastrophic (up to $+182\%$ in $D_{\text{Hole}}$), establishing that dead ends carry comparatively little load.}}
\label{app:fig:surgery}
\end{figure}

\rev{Fig.~\ref{app:fig:surgery} illustrates the interventions, and the per-design means are collected in Table~\ref{app:tab:surgery} (Section~\ref{sec:dead_end}). The contrast is consistent across all nine designs. Pruning dead ends raises compliance by only $+5.9\%$ on average (range $+0.6$ to $+12.1\%$), whereas pruning a single load-bearing rib raises it by $+61.5\%$ on average (up to $+182\%$ in $D_{\text{Hole}}$). Grafting an artificial dead-end changes compliance by less than $1\%$ ($-0.9$ to $-0.1\%$), directly showing that a dead-end adds no meaningful load path. Consistently, the pruned dead-end regions carry only $0.07$--$3.7\%$ of the total strain energy, versus $8.3$--$49.5\%$ for the removed load-bearing rib. The small but non-zero cost of dead-end removal is the expected first-order effect ($\Delta C \approx p\,f$ when removing material of strain-energy fraction $f$ at penalization $p=3$). Because dead ends carry only a few percent of the strain energy, removing them costs roughly an order of magnitude less than removing a true rib. This intervention leaves the surrounding topology unchanged and thus confirms causally, without the morphological confound of the \textit{dot} prompt, that dead-end suppression underlies the SDS-guided improvement.}

\clearpage  
\bibliographystyle{elsarticle-num}
\bibliography{sn-bibliography}
\label{LastPage}

\end{document}